\documentclass[10pt,twocolumn,letterpaper]{article}
\usepackage{cvpr}
\usepackage{times}
\usepackage{epsfig}
\usepackage{graphicx}
\usepackage{amsmath}
\usepackage{amssymb}

\usepackage{times}
\usepackage{epsfig}
\usepackage{xspace}
\usepackage{color}
\usepackage{multirow}
\usepackage{graphics}
\usepackage{adjustbox}
\usepackage{subfigure}
\usepackage{amsmath}
\usepackage[noend]{algpseudocode}
\usepackage{algorithmicx,algorithm}
\usepackage{amsfonts,amssymb}
\usepackage{paralist} 
\usepackage{enumitem}
\usepackage{graphicx} 
\usepackage{epstopdf}
\usepackage{booktabs} 
\usepackage{setspace}

\usepackage[pagebackref=true,breaklinks=true,letterpaper=true,colorlinks,citecolor=blue,linkcolor=blue,bookmarks=false]{hyperref}

\long\def\ignorethis#1{}

\definecolor{Gray}{rgb}{0.35,0.35,0.35}
\definecolor{Blue}{rgb}{0,0.2,0.8}
\definecolor{Red}{rgb}{0.8,0.2,0}
\definecolor{Green}{rgb}{0.0,0.5,0.1}
\definecolor{Gray}{rgb}{0.4,0.4,0.4}

\newlength\paramargin
\newlength\figmargin
\newlength\secmargin

\usepackage{array}
\newcolumntype{L}[1]{>{\raggedright\let\newline\\\arraybackslash\hspace{0pt}}m{#1}}
\newcolumntype{C}[1]{>{\centering\let\newline\\\arraybackslash\hspace{0pt}}m{#1}}
\newcolumntype{R}[1]{>{\raggedleft\let\newline\\\arraybackslash\hspace{0pt}}m{#1}}

\def\ie{i.e.,~}

\def\etal{et~al.\xspace}

\graphicspath{{figure/eps/}}

\usepackage[breaklinks=true,bookmarks=false]{hyperref}
\cvprfinalcopy
\def\cvprPaperID{169} 

\ifcvprfinal\fi
\begin{document}
\title{Domain Adaptation for Image Dehazing}

\author{
	Yuanjie Shao$^1$, 
	Lerenhan Li$^1$, 
	Wenqi Ren$^2$, 
	Changxin Gao$^1$, 
	Nong Sang$^1$\thanks{Corresponding author.}\\
	$^1$National Key Laboratory of Science and Technology on Multispectral Information Processing,\\
	School of Artificial Intelligence and Automation, \\
	Huazhong University of Science and Technology, Wuhan, China\\
	$^2$Institute of Information Engineering, Chinese Academy of Sciences, Beijing, China\\
	{\tt\small \{shaoyuanjie, lrhli, cgao, nsang\}@hust.edu.cn, rwq.renwenqi@gmail.com}
}

\maketitle
\thispagestyle{empty}
\begin{abstract}
Image dehazing using learning-based methods has achieved state-of-the-art performance in recent years.
However, most existing methods train a dehazing model on synthetic hazy images, which are less able to generalize well to real hazy images due to domain shift.
To address this issue, we propose a domain adaptation paradigm,
which consists of an image translation module and two image dehazing modules. 
%
Specifically, we first apply a bidirectional translation network to bridge the gap between the synthetic and real domains by translating images from one domain to another.
And then, we use images before and after translation to train the proposed two image dehazing networks with a consistency constraint.
%
%
In this phase, we incorporate the real hazy image into the dehazing training via exploiting the properties of the clear image (e.g., dark channel prior and image gradient smoothing) to further improve the domain adaptivity.
By training image translation and dehazing network in an end-to-end manner, we can obtain better effects of both image translation and dehazing.
%
Experimental results on both synthetic and real-world images demonstrate that our model performs favorably against the state-of-the-art dehazing algorithms.
\end{abstract}

\section{Introduction}
Single image dehazing aims to recover the clean image from a hazy input,
which is essential for subsequent high-level tasks, such as object recognition and scene understanding.
Thus, it has received significant attention in the vision community over the past few years.
According to the physical scattering models~\cite{mccartney1976optics,narasimhan2002vision,li2017haze}, the hazing process is usually formulated as
\begin{equation}
\label{eqn:degradation}
    I(x) = J(x)t(x) + A(1 - t(x)).
\end{equation}
where $I(x)$ and $J(x)$ denote the hazy image and the clean image, $A$ is the global atmospheric light, and $t(x)$ is the transmission map. 
The transmission map can be represented as $t(x) = {e^{ - \beta d(x)}}$, where $d(x)$ and $\beta$ denote the scene depth and the atmosphere scattering parameter, respectively.
Given a hazy image $I(x)$, most dehazing algorithms try to estimate $t(x)$ and $A$.

However, estimating the transmission map from a hazy image is an ill-posed problem generally.
Early prior-based methods try to estimate the transmission map by exploiting the statistical properties of clear images, such as dark channel prior~\cite{He2011Single} and color-line prior~\cite{fattal2014dehazing}.
Unfortunately, these image priors are easily inconsistent with the practice, which may lead to inaccurate transmission approximations.
\begin{figure}[tbp]
\footnotesize
\centering
\renewcommand{\tabcolsep}{1pt} 
\renewcommand{\arraystretch}{1} 
\begin{center}
\begin{tabular}{ccc}
  \includegraphics[width=0.32\linewidth]{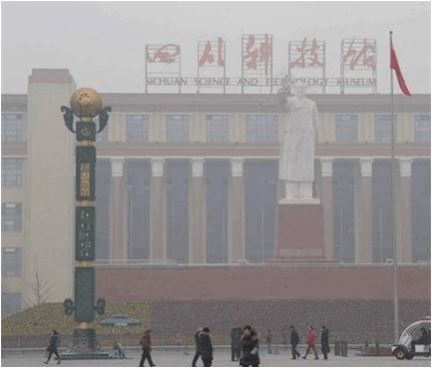} &
  \includegraphics[width=0.32\linewidth]{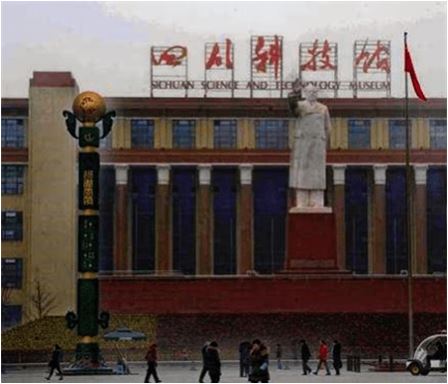} &
  \includegraphics[width=0.32\linewidth]{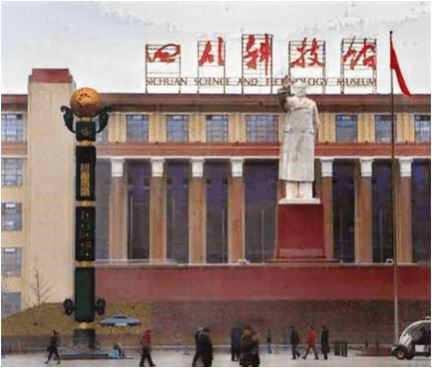} \\
  (a) Hazy image &
  (b) EPDN~\cite{qu2019enhanced} &
  (c) Ours\\
\end{tabular}
\end{center}
\vspace{-2mm}
\caption{Dehazed results on a real world hazy image. 
	Our method generates a cleaner image with fewer color distortion and brighter details.
	(a) Hazy image. 
	(b) The result of EPDN~\cite{qu2019enhanced}. 
	(c) Our result.
}
\vspace{-4mm}
\label{fig:intro}
\end{figure}
Thus, the quality of the restored image is undesirable.

To deal with this problem, convolutional neural networks (CNNs) have been employed
%
to estimate transmissions~\cite{Cai2016DehazeNet, ren2016single, Zhang_2018_CVPR} or predict clear images directly~\cite{li2017aod, Ren_2018_CVPR, Li_2018_CVPR, qu2019enhanced}.
These methods are valid and superior to the prior-based algorithms with significant performance improvements.
However, deep learning-based approaches need to rely on a large amount of real hazy images and their haze-free counterparts for training.
In general, it is impractical to acquire large quantities of ground-truth images in the real world. 
Therefore, most dehazing models resort to training on synthetic hazy dataset.
However, due to the domain shift problem, the models learned from synthetic data often fail to generalize well to real data.

To address this issue, we propose a domain adaptation framework for single image dehazing.
The proposed framework includes two parts, namely an image translation module and two domain-related dehazing modules (one for synthetic domain and another for real domain).
To reduce the discrepancy between domains, our method first employs the bidirectional image translation network to translate images from one domain to another.
Since image haze is a kind of noise and nonuniform highly depending on the scene depth, we incorporate the depth information into the translation network to guide the translation of synthetic to real hazy images.
Then, the domain-related dehazing network takes images of this domain, including the original and translated images, as inputs to perform image dehazing.
Moreover, we use a consistency loss to ensure that the two dehazing networks generate consistent results. 
In this training phase, to further improve the generalization of the network in the real domain, we incorporate the real hazy images into the training.
We hope that the dehazing results of the real hazy image can have some properties of the clear images, such as dark channel prior and image gradient smoothing.
We train the image translation network and dehazing networks in an end-to-end manner so that they can improve each other.
As shown in Figure~\ref{fig:intro}, our model produces a cleaner image when compared with recent dehazing work of EPDN~\cite{qu2019enhanced}.

We summarize the contributions of our work as follows:
\vspace{-3mm}
\begin{compactitem}
\item We propose an end-to-end domain adaptation framework for image dehazing, which effectively bridges the gap between the synthetic and real-world hazy images.
\item We show that incorporating real hazy images into the training process can improve the dehazing performance. 
\item We conduct extensive experiments on both synthetic datasets and real-world hazy images, which demonstrate that the proposed method performs favorably against the state-of-the-art dehazing approaches.
\end{compactitem}
\section{Related Work}
In this section, we briefly discuss the single image dehazing approaches and domain adaptation methods, which are related to our work.
\subsection{Single Image Dehazing}
%
\paragraph{Prior-based methods.}
Prior-based methods estimate the transmission maps and atmospheric light intensity based on the statistics of clear images.
Representative works in this regard include~\cite{tan2008visibility,He2011Single,zhu2015fast,fattal2014dehazing,berman2016non}.
Specifically, Tan~\cite{tan2008visibility} proposes a contrast maximization method for image dehazing since it observes that clear images tend to have higher contrast than their hazy counterparts. 
He~\etal~\cite{He2011Single} make use of dark channel prior (DCP) to estimate the transmission map, which is based on the assumption that pixels in haze-free patches are close to zero in at least one color channel.
Follow-up works have improved the efficiency and performance of the DCP method~\cite{tarel2009fast, meng2013efficient, li2015nighttime, nishino2012bayesian, Yang_2018_ECCV}.
Besides, the attenuation prior is adopted in~\cite{zhu2015fast} for recovering depth information of the hazy images. 
Fattal~\cite{fattal2014dehazing} uses the color-line assumption to recover the scene transmission, which asserts that pixels of small image patches exhibit
a one-dimensional distribution.
Similarly, Berman~\etal~\cite{berman2016non} assume that several hundreds of distinct colors can well approximate colors of a clear image, and then perform image dehazing based on this prior.
Though these methods have been shown effective for image dehazing, their performances are inherently limited since the assumed priors are not suited for all real-word images.
\vspace{-4mm}
\paragraph{Learning-based Methods.}
With the advance in deep convolutional neural networks (CNNs) and the availability of large-scale synthetic datasets, data-driven approaches for image dehazing have received significant attention in recent years.
Many methods~\cite{Cai2016DehazeNet, ren2016single, Zhang_2018_CVPR, li2017aod} directly utilize deep CNNs to estimate the transmissions and atmospheric light, and then restore the clean image according to the degradation model~\eqref{eqn:degradation}.
Cai~\etal~\cite{Cai2016DehazeNet} propose an end-to-end dehazing model, DehazeNet, to estimate the transmission map from hazy images.
Ren~\etal~\cite{ren2016single} utilize a coarse-to-fine strategy to learn the mapping between hazy inputs and transmission maps.
Zhang and Patel~\cite{Zhang_2018_CVPR} propose a densely connected pyramid network to estimate the transmission maps.
%
%
Li~\etal~\cite{li2017aod} propose an AOD-Net to estimate the parameter of the reformulated physical scattering model, which integrates the transmissions and atmospheric light.
Also, some end-to-end methods~\cite{Ren_2018_CVPR, Li_2018_CVPR, qu2019enhanced} have been proposed to recover the clean image directly instead of estimating the transmission map and atmospheric light. 
Ren~\etal~\cite{Ren_2018_CVPR} adopt a gated fusion network to recover the clean image from a hazy input directly.
Qu~\etal~\cite{qu2019enhanced} transform the problem of image dehazing to an image-to-image translation problem, and propose an enhanced pix2pix dehazing network.

However, due to the domain gap between synthetic and real data, the CNN-based models trained on synthetic images tend to have a significant performance drop when applied to the real domain. 
To this end,  Li~\etal~\cite{li2019semi} propose a semi-supervised dehazing model, which is trained on both synthetic and real haze image, and thus enjoys domain adaptivity between synthetic and real-world hazy images. 
However, only applying the real haze image for training does not really solve the problem of domain shift. 
Different from the above methods, our model first applies the image translation network to translate images from one domain to another, and then performs image dehazing on both synthetic and real domain using the translated images and their original images (synthetic or real).
The proposed approach can effectively solve the domain shift problem.
\begin{figure*}[t]
\begin{center}
    \includegraphics[width=0.96\linewidth]{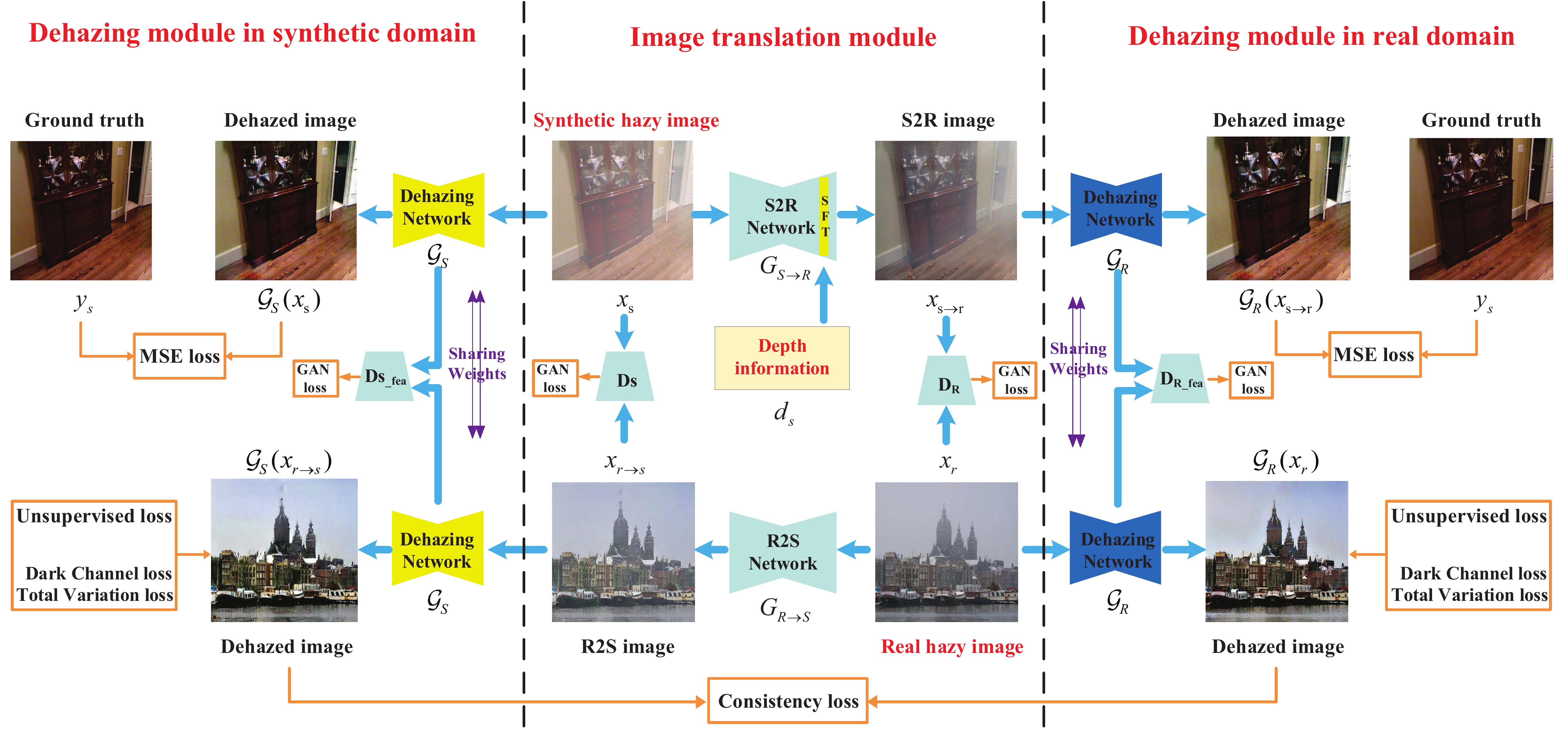}
\end{center}
\vspace{-2mm}
\caption{Architecture of the proposed domain adaptation framework for image dehazing.
        The framework consists of two parts, an image translation module and two image dehazing modules.
        The image translation module translates images from one domain to another to reduce the domain discrepancy.
        The image dehazing modules perform image dehazing on both synthetic and real domain.}
\label{fig:network}
\end{figure*}

\subsection{Domain Adaptation}
%
Domain adaptation aims to reduce the discrepancy between different domains~\cite{atapour2018real,chen2018domain,long2013transfer}. 
Existing work either to perform feature-level or pixel-level adaptation.
Feature-level adaptation methods aim at aligning the feature distributions between the source and target domains through minimizing the maximum mean discrepancy~\cite{long2015learning}, or applying adversarial learning strategies~\cite{tzeng2017adversarial,tsai2018learning} on the feature space.
Another line of research focuses on pixel-level adaptation~\cite{bousmalis2017unsupervised, shrivastava2017learning, dundar2018domain}. 
These approaches deal with the domain shift problem by applying image-to-image translation~\cite{bousmalis2017unsupervised, shrivastava2017learning} learning, or style transfer~\cite{dundar2018domain} methods to increase the data in the target domain.

Most recently, many methods perform feature-level and pixel-level adaptation jointly in many visual tasks, e.g.,  image classification~\cite{hoffman2017cycada}, semantic segmentation~\cite{chen2019learning}, and depth prediction~\cite{zheng2018t2net}. 
These methods~\cite{chen2019learning,zheng2018t2net} translate images from one domain to another with pixel-level adaptation via image-to-image translation networks, e.g., the CycleGAN~\cite{zhu2017unpaired}.
The translated images are then inputted to the task network with feature-level alignment.
In this work, we take advantage of CycleGAN to adapt the real hazy images to our dehazing model trained on synthetic data. 
Moreover, since the depth information is closely related to the formulation of image haze, we incorporate the depth information into the translating network to better guide the real hazy image translation. 

\section{Proposed Method}
This section presents the details of our domain adaptation framework.
First, we provide an overview of our method. 
Then we describe the details of the image translation module and image dehazing module.
Finally, we give the loss functions that are applied to train the proposed networks.

\subsection{Method Overview}
Given a synthetic dataset ${X_S} = \{ {x_s},{y_s}\} _{s = 1}^{{N_l}}$ and a real hazy image set ${X_R} = \{ {x_r}\} _{r = 1}^{{N_u}}$, where ${N_l}$ and ${N_u}$ denote the number of the synthetic and real hazy images, respectively.
We aim to learn a single image dehazing model which can accurately predict the clear image from real hazy image. 
Due to the domain shift, the dehazing model trained only on the synthetic data can not generalize well to the real hazy image. 

To deal with this problem, we present a domain adaptation framework, which consists of two main parts: the image translation network ${G_{S \to R}}$ and ${G_{R \to S}}$, and two dehazing networks ${{\cal G}_S}$ and ${{\cal G}_R}$. 
The image translation network translates images from one domain to another to bridge the gap between them.
Then the dehazing networks perform image dehazing using both translated images and source images (e.g., synthetic or real).

As illustrated in Figure~\ref{fig:network}, the proposed model takes a real hazy image ${x_{\rm{r}}}$ and a synthetic image ${x_{\rm{s}}}$ along with its corresponding depth images ${d_s}$ as input. 
We first obtain the corresponding translated images ${x_{{\rm{s}} \to {\rm{r}}}} = {G_{S \to R}}({x_{\rm{s}}},{d_s})$ and ${x_{r \to s}} = {G_{R \to S}}({x_r})$ using two image translators. 
And then, we pass ${x_s}$ and ${x_{r \to s}}$ to ${{\cal G}_S}$, ${x_r}$ and ${x_{r \to s}}$ to ${{\cal G}_R}$ to perform image dehazing.

\begin{table*}[t]
\footnotesize
\caption{Configurations of image translation module.
        ``Conv'' denotes the convolution layer, ``Res'' denotes the residual block, ``Upconv'' denotes the up-sample layer by transposed convolution operator and ``Tanh'' denotes the non-linear Tanh layer.}
\vspace{2mm}
\label{tab:tran_s2r}
\centering
\begin{tabular}{l|c|c|c|c|c|c|c|c|c}
\toprule
Layer        & Conv1 & Conv2 & Conv3   & Res4-12   & Upconv13  & Upconv14   & SFT layer  & Conv15 & Tanh  \\ \midrule
In\_channels  & 3     & 64    & 128      & 256      & 256      	& 128         & 64       & 64     & 3    \\
Out\_channels & 64    & 128   & 256      & 256      & 128       & 64          & 64       & 3      & 3    \\
Kernel size   & 7     & 3     & 3        & -        & 3        	& 3           & -        & 7      & -     \\
Stride        & 1     & 2     & 2        & -        & 2        	& 2           & -        & 1      & -     \\
Pad           & -     & 1     & 1        & -        & 1        	& 1           & -        & -      & -     \\ \bottomrule
\end{tabular}
\end{table*}

\subsection{Image Translation Module}
The image translation module includes two translators: synthetic to real network ${G_{S \to R}}$ and real to synthetic network ${G_{R \to S}}$.
The ${G_{S \to R}}$ network takes $({X_{\rm{s}}},{D_{\rm{s}}})$ as inputs, and generates translated images ${G_{S \to R}}({X_{\rm{s}}},{D_{\rm{s}}})$ with similar style to the real hazy images.
Another translator ${G_{R \to S}}$ performs image translation inversely.
Since the depth information is highly correlated to the hazing formulation, we incorporate it into the generator ${G_{S \to R}}$ to produce images with similar haze distribution in real cases.
%
%
%
\begin{figure}[t]
	\centering
    \includegraphics[width=\linewidth]{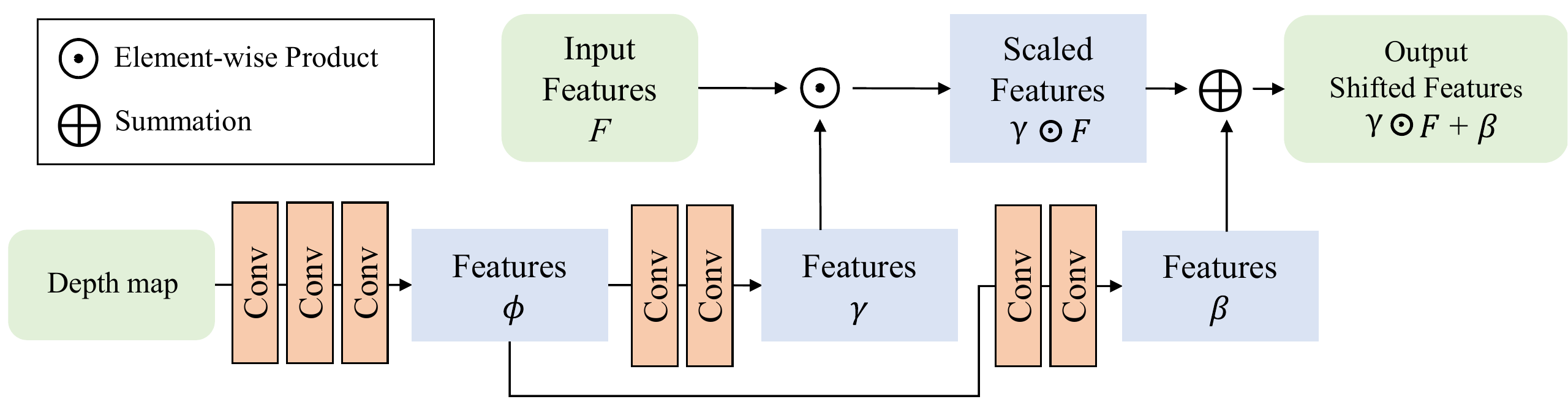}
	\caption{Structure of the SFT layer.
        In the translator ${G_{S \to R}}$, we consider the depth map as the guidance to assist the image translation.
        }
\label{fig:SFT_layer}
\end{figure}

\begin{figure}[t]
	\footnotesize
	\centering
	\renewcommand{\tabcolsep}{1pt} 
	\renewcommand{\arraystretch}{1} 
	\begin{center}
		\begin{tabular}{ccc}
			\includegraphics[width=0.32\linewidth]{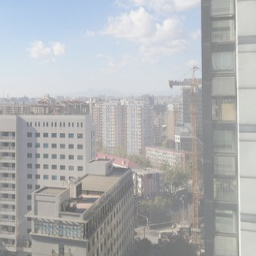} &
			\includegraphics[width=0.32\linewidth]{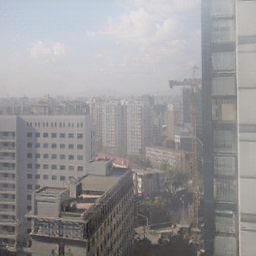} &
			\includegraphics[width=0.32\linewidth]{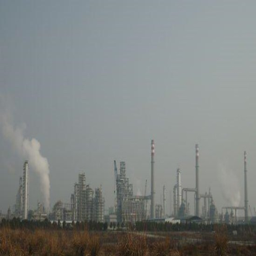} \\

			\includegraphics[width=0.32\linewidth]{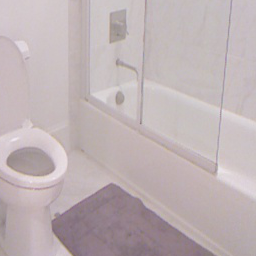} &
			\includegraphics[width=0.32\linewidth]{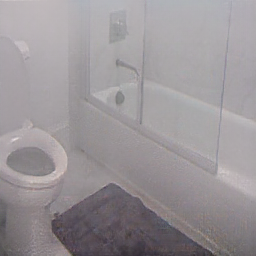} &
			\includegraphics[width=0.32\linewidth]{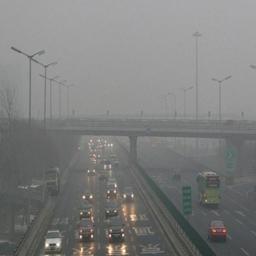} \\

			(a) Synthetic hazy image &
			(b) Translated image &
			(c) Real hazy image\\
		\end{tabular}
	\end{center}
	\vspace{-2mm}
	\caption{Translated results on two synthetic hazy images. 
	}
	\vspace{-4mm}
	\label{fig:translation}
\end{figure}

We adopt the spatial feature transform (SFT) layer~\cite{wang2018recovering, li2020dyna} to incorporate the depth information into the translation network, which can fuse features from depth map and synthetic image effectively. 
As shown in Fig.~\ref{fig:SFT_layer}, the SFT layer first applies three convolution layers to extract conditional maps $\phi$ from the depth map.
The conditional maps are then fed to the other two convolution layers to predict the modulation parameters, $\gamma$ and $\beta$, respectively.
Finally, we can obtain the output shifted features by:
\begin{equation}
	\label{eqn:SFT_layer}
	\mathrm{SFT}(F|\gamma ,\beta) = \gamma \odot F + \beta.
\end{equation}
where $\odot$ is the element-wise multiplication.
In the translator ${G_{S \to R}}$, we treat the depth map as the guidance and use the SFT layer to transform the features of the penultimate convolution layer.
As shown in Fig.~\ref{fig:translation}, the synthetic images are relative closer to the real-world hazy image after the translation.
We show the detailed configurations of the translator ${G_{S \to R}}$ in Table~\ref{tab:tran_s2r}.
We also employ the architectures, provided by CycleGAN~\cite{zhu2017unpaired}, for the generator ${G_{R \to S}}$ and discriminators ($D_R^{img}$ and $D_S^{img}$).

\subsection{Dehazing Module}
Our method includes two dehazing modules ${{\cal G}_S}$ and ${{\cal G}_R}$, which perform image dehazing on synthetic and real domains, respectively.
${{\cal G}_S}$ takes the synthetic image ${x_s}$ and the translated image ${x_{{\rm{r}} \to {\rm{s}}}}$ as inputs to perform image dehazing.
And ${{\cal G}_R}$ is trained on ${x_r}$ and ${x_{{\rm{s}} \to {\rm{r}}}}$.
For these two image dehazing networks, we both utilize a standard encoder-decoder architecture with skip connections and side outputs as~\cite{zheng2018t2net}.
The dehazing network in each domain shares the same network architecture but with different learned parameters.

\subsection{Training Losses}
In the domain adaptation framework, we adopt the following losses to train the network.
\vspace{-5mm}
\paragraph{Image translation Losses.}
The aim of our translate module is to learn the translators ${G_{S \to R}}$ and ${G_{R \to S}}$ to reduce the discrepancy between the synthetic domain ${X_S}$ and the real domain ${X_R}$. 
For translators ${G_{S \to R}}$, we expect the ${G_{S \to R}}({x_s},{d_s})$ to be indistinguishable from the real hazy image ${x_r}$. 
Thus, we employ an image-level discriminators $D_R^{img}$ and a feature-level discriminators $D_R^{feat}$, to perform a minmax game via an adversarial learning manner.
The $D_R^{img}$ aims at aligning the distributions between the real image ${x_r}$ and the translated image ${G_{S \to R}}({x_s},{d_s})$.
The discriminator $D_R^{feat}$ helps align the distributions between the feature map of ${x_r}$ and ${G_{S \to R}}({x_s},{d_s})$.
The adversarial losses are defined as:
\begin{equation}
\label{eqn:DR_img}
      \begin{array}{l}
L_{Gan}^{img}({X_R},({X_S},{D_S}),D_R^{img},{G_{S \to R}})\\[1.8mm]
  \quad \, = {\mathbb{E}_{{x_s} \sim {X_S},{d_s} \sim {D_S}}}[D_R^{img}({G_{S \to R}}({x_s},{d_s}))]\\[1.8mm]
  \quad \, + {\mathbb{E}_{{x_r} \sim {X_R}}}[D_R^{img}({x_r}) - 1].
\end{array}
\end{equation}
\begin{equation}
\label{eqn:DR_fea}
    \begin{array}{l}
L_{Gan}^{feat}({X_R},({X_S},{D_S}),D_R^{feat},{G_{S \to R}},{{\cal G}_{R}})\\[1.8mm]
 \quad \, = {\mathbb{E}_{{x_s} \sim {X_S},{d_s} \sim {D_S}}}[D_R^{feat}({{\cal G}_{R}}({G_{S \to R}}({x_s},{d_s})))]\\[1.8mm]
 \quad \, + {\mathbb{E}_{{x_r} \sim {X_R}}}[D_R^{feat}({{\cal G}_{R}}({x_r})) - 1].
\end{array}
\end{equation}

Similar to ${G_{S \to R}}$, the translator ${G_{R \to S}}$ has another image-level adversarial loss  and feature-level adversarial loss, which are denoted as  $L_{Gan}^{img}({X_S},{X_R},D_S^{img},{G_{R \to S}})$, $L_{Gan}^{feat}({X_S},{X_R},D_S^{feat},{G_{R \to S}},{{\cal G}_{S}})$, respectively.

In addition, we utilize the cycle-consistency loss~\cite{zhu2017unpaired} to regularize the training of translation network. 
Specifically, when passing an image ${x_s}$ to ${G_{S \to R}}$ and ${G_{R \to S}}$ sequentially, we expect the output should be the same image, and vice versa for ${x_r}$.
Namely, ${G_{R \to S}}({G_{S \to R}}({x_{\rm{s}}},{d_s})) \approx {x_{\rm{s}}}$ and ${G_{S \to R}}({G_{R \to S}}({x_r}),{d_r}) \approx {x_r}$.  
The cycle consistency loss can be expressed as:
\begin{equation}
\label{eqn:L_c}
    \begin{array}{l}
{L_c}  = {\mathbb{E}_{{x_s} \sim {X_S},{d_s} \sim {D_S}}}[||{G_{R \to S}}({G_{S \to R}}({x_s},{d_s}) - {x_s}|{|_1}]\\ [1.8mm]
\quad \, {\rm{    }} + {\mathbb{E}_{{x_r} \sim {X_R},{d_r} \sim {D_R}}}[||{G_{S \to R}}({G_{R \to S}}({x_r}),{d_r}) - {x_r}|{|_1}].
\end{array}
\end{equation}
Finally, to encourage the generators to preserve content information between the input and output, we also utilize an identity mapping loss~\cite{zhu2017unpaired}, which is denoted as:
\begin{equation}
\label{eqn:L_i}
    \begin{array}{l}
{L_{idt}} = {\mathbb{E}_{{x_s} \sim {X_S}}}[||{G_{R \to S}}({x_s}) - {x_s}|{|_1}]\\[1.8mm]
 \quad \quad  + {\mathbb{E}_{{x_r} \sim {X_R},{d_r} \sim {D_R}}}[||{G_{S \to R}}({x_r},{d_r}) - {x_r}|{|_1}].
\end{array}
\end{equation}

The full loss function for the the translating module is as follow:
\begin{equation}
\label{eqn:Loss_trans}
    \begin{array}{l}
{L_{tran}} = L_{Gan}^{img}({X_R},({X_S},{D_S}),D_R^{img},{G_{S \to R}})\\[1.8mm]
 \quad  \quad  \ \ \, + L_{Gan}^{feat}({X_R},({X_S},{D_S}),D_R^{feat},{G_{S \to R}},{{\cal G}_{R}})\\[1.8mm]
 \quad  \quad  \ \ \, + L_{Gan}^{img}({X_S},{X_R},D_S^{img},{G_{R \to S}})\\[1.8mm]
 \quad  \quad  \ \ \, + L_{Gan}^{feat}({X_S},{X_R},D_S^{feat},{G_{R \to S}},{{\cal G}_{S}})\\[1.8mm]
 \quad  \quad  \ \ \, + {\lambda _1}{L_c} + {\lambda _2}{L_{idt}}.
\end{array}
\end{equation}

\vspace{-5mm}
\paragraph{Image dehazing Losses.}
We can now transfer the synthetic images ${X_S}$ and the corresponding depth images ${D_S}$ to the generator ${G_{S \to R}}$, and obtain a new dataset ${X_{S \to R}} = {G_{S \to R}}({X_{S}},{D_{S}})$, which has a similar style with real hazy images. 
And then, we train a image dehazing network ${{\cal G}_R}$ on ${X_{S \to R}}$  and ${X_R}$ in a semi-supervised manner. 
For supervised branch, we apply the mean squared loss to ensure the predicted images ${{J_{S \to R}}}$ is close to clean images ${{Y_S}}$, which can be defined as:
\begin{equation}
\label{eqn:MSE_loss}
{L_{rm}} = \left\| {{J_{S \to R}} - {Y_S}} \right\|_2^2.
\end{equation}

In the unsupervised branch, we introduce the total variation and dark channel losses, which regularize the dehazing network to produce images with similar statistical characteristics of the clear images.
The total variation loss is an $\ell_1$-regularization gradient prior on the predicted images ${{J_R}}$: 
%
\begin{equation}
\label{eqn:TV_loss}
{L_{rt}} = {\left\| {{\partial _h}{J_R}} \right\|_1} + {\left\| {{\partial _v}{J_R}} \right\|_1}.
\end{equation}
where ${{\partial _h}}$ denotes the horizontal gradient operators, and ${{\partial _v}}$ represents the vertical gradient operators.

Furthermore, the recent work~\cite{He2011Single} has proposed the concept of the dark channel, which can be expressed as:
\begin{equation}
\label{eqn:DC}
D(I) = \mathop {\min }\limits_{y \in N(x)} \left[ {\mathop {\min }\limits_{c \in \{ r,g,b\} } {I^c}(y)} \right].
\end{equation}
where $x$ and $y$ are pixel coordinates of image $I$, $I^c$ represents $c$-th color channel of $I$, and $N(x)$ denotes the local neighborhood centered at $x$. 
He~\etal~\cite{He2011Single} have also shown that most intensity of the dark channel image are zero or close to zero .
Therefore, we apply the following dark channel (DC) loss to ensure that the dark channel of the predicted images are in consistence with that of clean image:
\begin{equation}
\label{eqn:DC_loss}
{L_{rd}} = {\left\| {D({J_R})} \right\|_1}.
\end{equation}
In addition, we also train a complementary image dehazing network ${{\cal G}_S}$ on ${X_S}$ and ${X_{R \to S}}$.
Similarly, we apply the same supervised loss and unsupervised loss to train the dehazing network ${{\cal G}_S}$, which are as follows:
\begin{equation}
\label{eqn:MSE_loss1}
{L_{sm}} = \left\| {{J_S} - {Y_S}} \right\|_2^2,
\end{equation}
\begin{equation}
\label{eqn:TV_loss1}
{L_{st}} = {\left\| {{\partial _h}{J_{R \to S}}} \right\|_1} + {\left\| {{\partial _v}{J_{R \to S}}} \right\|_1},
\end{equation}
\begin{equation}
\label{eqn:DC_loss1}
{L_{sd}} = {\left\| {D({J_{R \to S}})} \right\|_1}.
\end{equation}

Finally, considering that the outputs of the two dehazing networks should be consistency for real hazy images, \ie${{\cal G}_R}({X_R}) \approx {{\cal G}_S}({G_{R \to S}}({X_R}))$, we introduce following consistency loss:
\begin{equation}
\label{eqn:consis_loss}
{L_c} = {\left\| {{{\cal G}_R}({X_R}) - {{\cal G}_S}({G_{R \to S}}({X_R}))} \right\|_1}.
\end{equation}
\paragraph{Overall Loss Function.}
The overall loss function are defined as follow:
\begin{equation}
\label{eqn:whole_loss}
\begin{array}{l}
L = {L_{tran}} + \lambda_m ({L_{rm}} + {L_{sm}}) + \lambda_d ({L_{rd}} + {L_{sd}})\\[1.8mm]
\quad + \lambda_t ({L_{rt}} + {L_{st}}) + \lambda_c {L_c}.
\end{array}
\end{equation}
where $\lambda_m$, $\lambda_d$, $\lambda_t$ and $\lambda_c$ are trade-off weights.
\begin{figure*}[htbp]
	\scriptsize
	\centering
	\renewcommand{\tabcolsep}{1pt} 
	\renewcommand{\arraystretch}{1} 
	\begin{center}
		\begin{tabular}{ccccccccc}
			\includegraphics[width=0.107\linewidth]{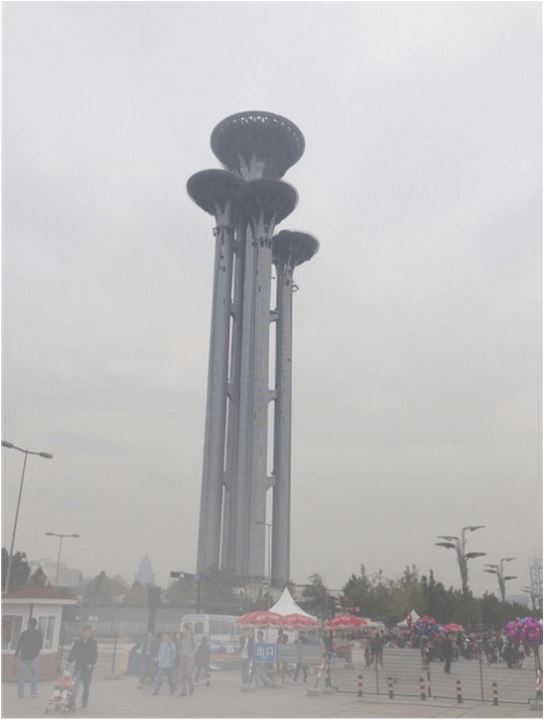} &
			\includegraphics[width=0.107\linewidth]{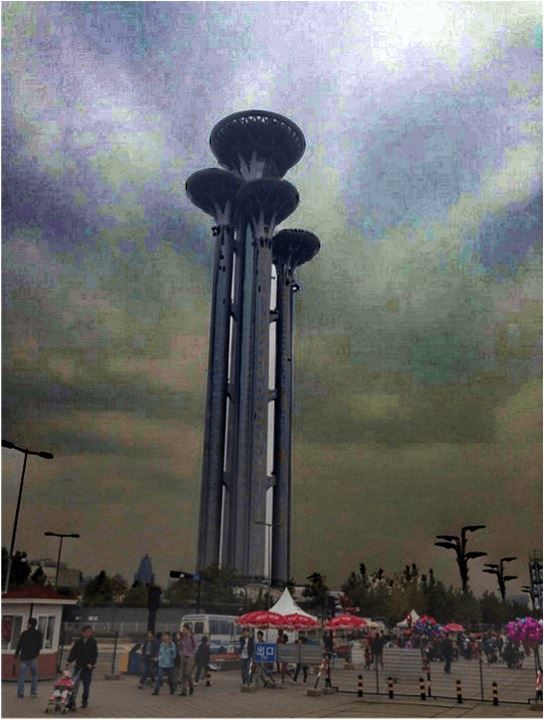} &
			\includegraphics[width=0.107\linewidth]{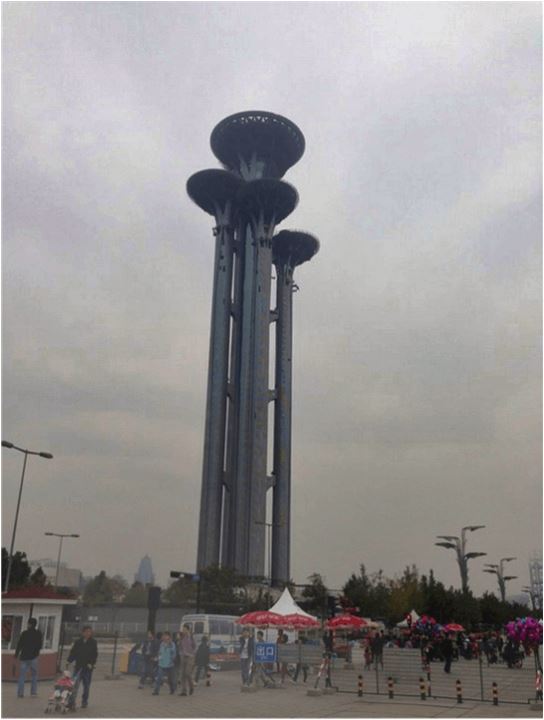} &
			\includegraphics[width=0.107\linewidth]{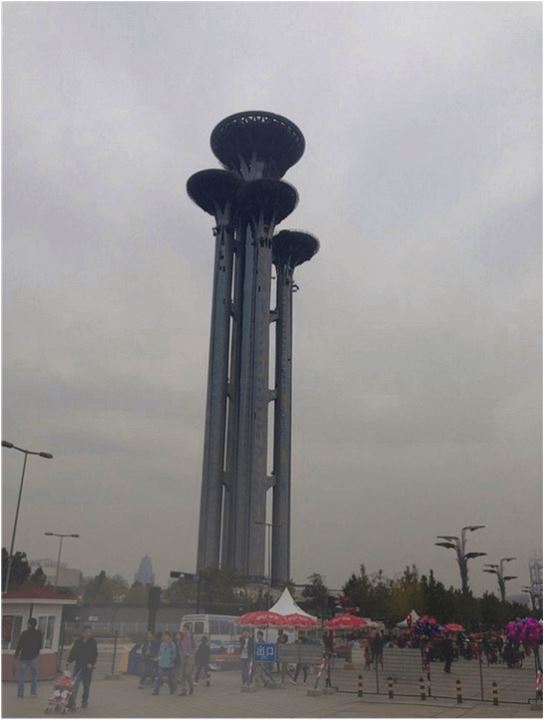} &			
			\includegraphics[width=0.107\linewidth]{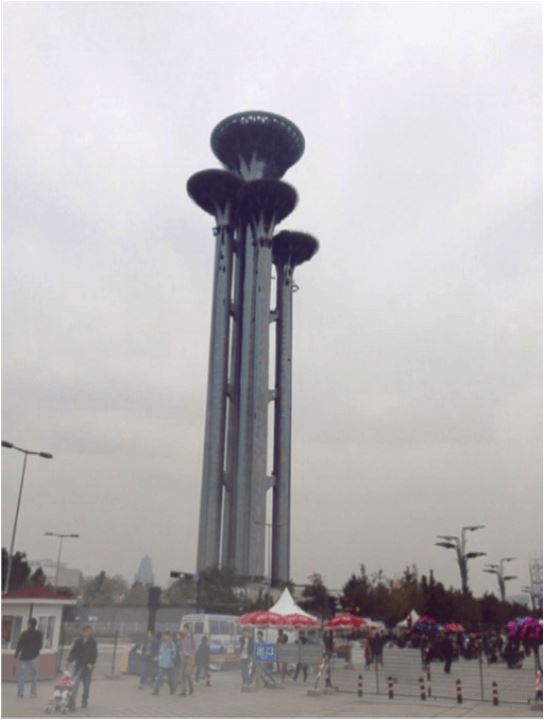} &
			\includegraphics[width=0.107\linewidth]{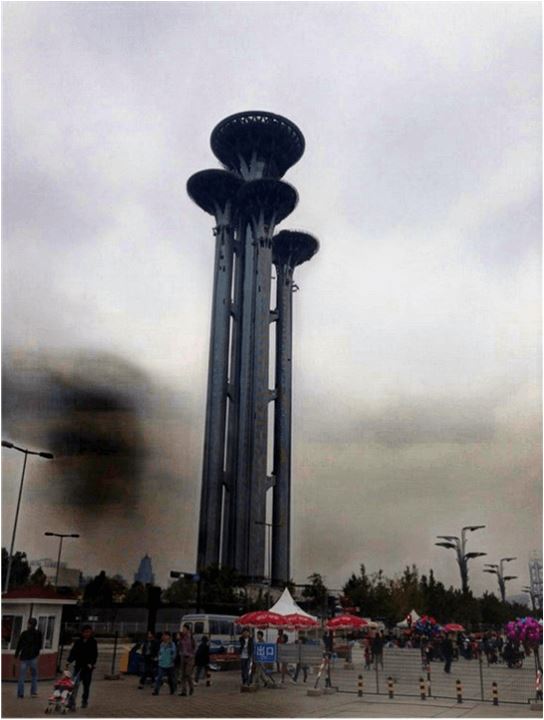} &
			\includegraphics[width=0.107\linewidth]{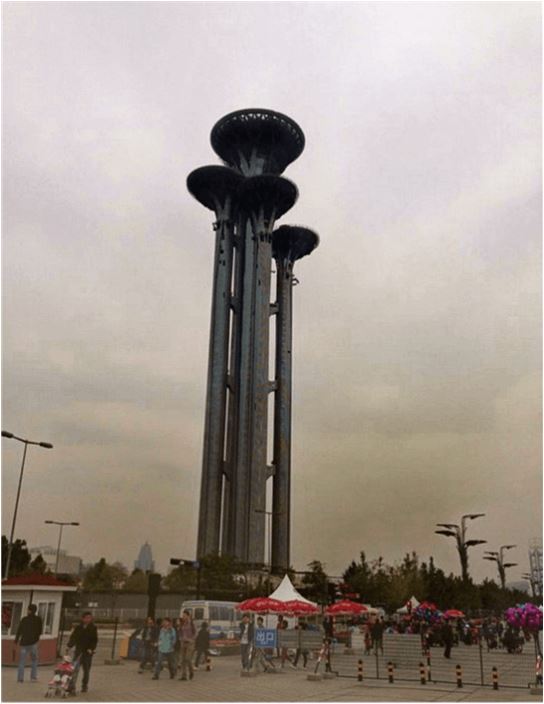} &
			\includegraphics[width=0.107\linewidth]{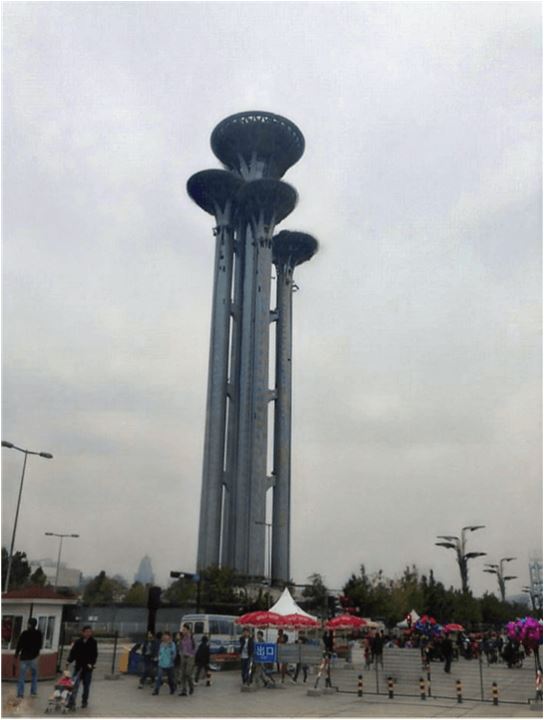} &
			\includegraphics[width=0.107\linewidth]{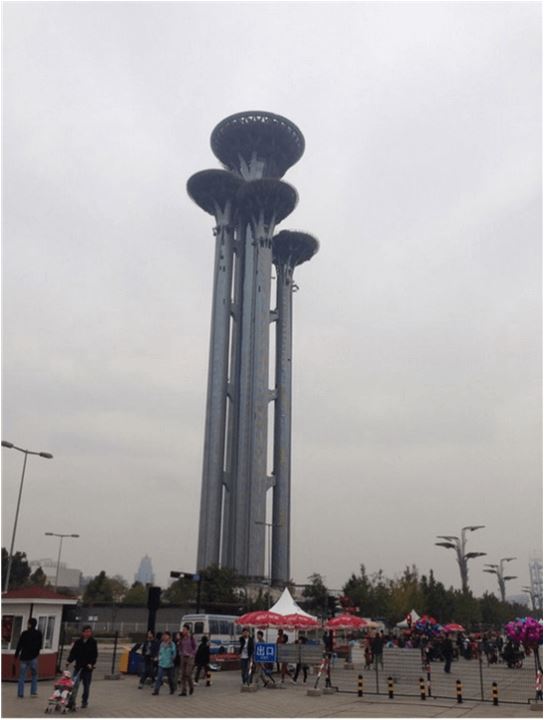} \\
			
			\includegraphics[width=0.107\linewidth]{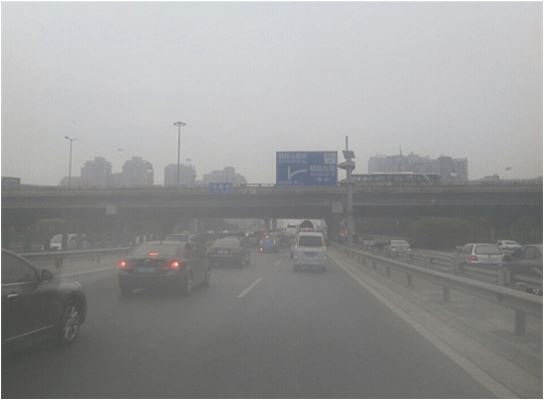} &
			\includegraphics[width=0.107\linewidth]{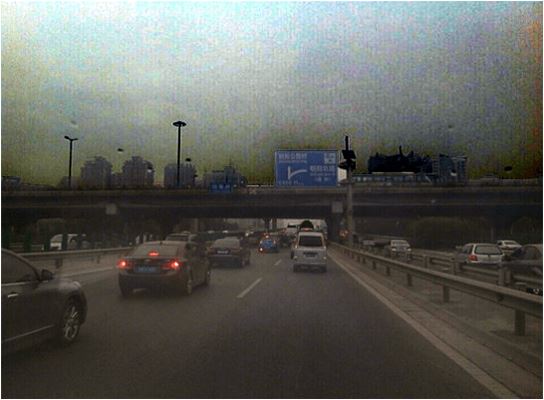} &
			\includegraphics[width=0.107\linewidth]{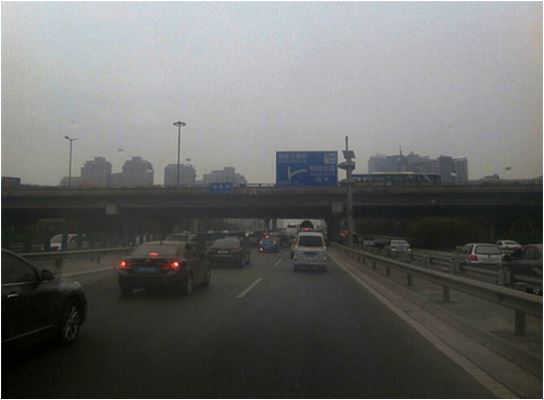} &
			\includegraphics[width=0.107\linewidth]{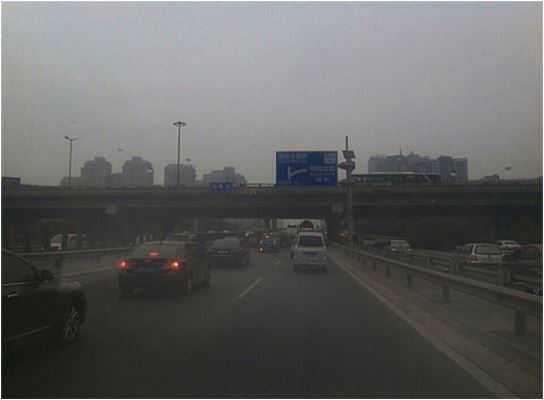} &
			\includegraphics[width=0.107\linewidth]{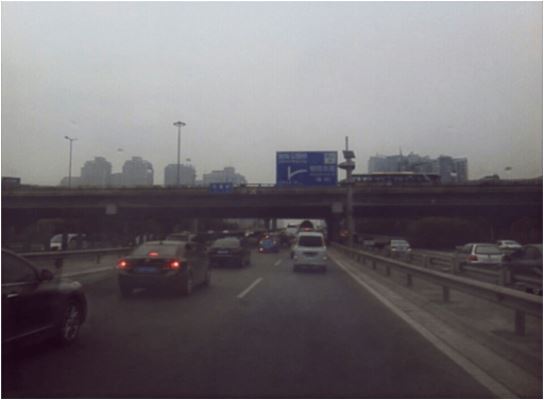} &
			\includegraphics[width=0.107\linewidth]{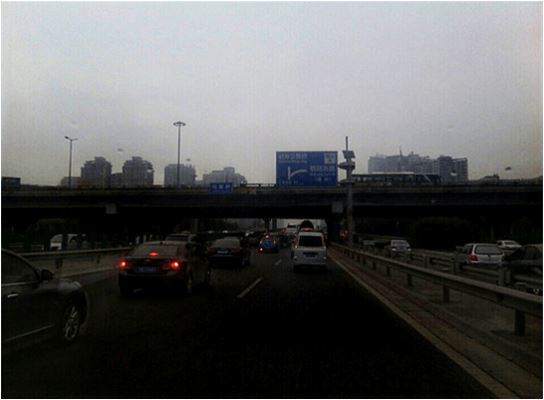} &
			\includegraphics[width=0.107\linewidth]{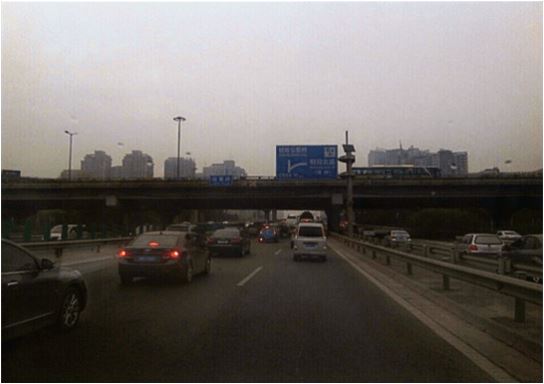} &
			\includegraphics[width=0.107\linewidth]{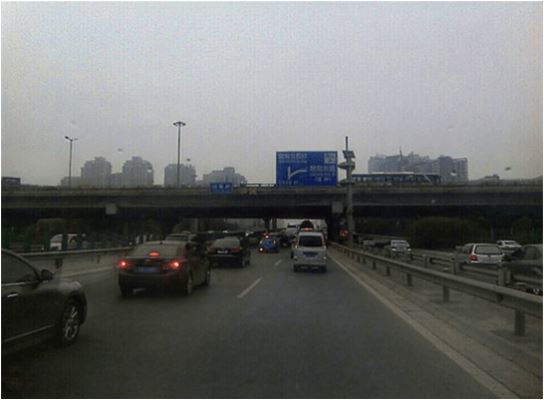} &
			\includegraphics[width=0.107\linewidth]{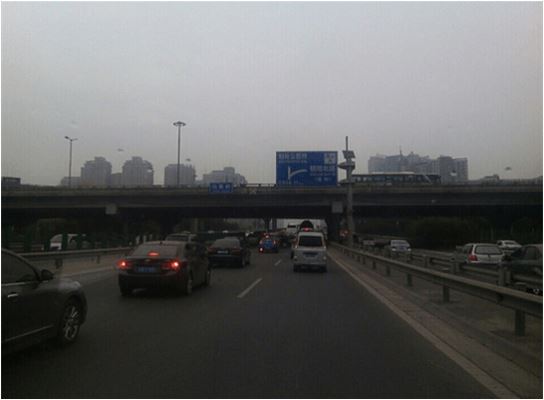} \\

			\includegraphics[width=0.107\linewidth]{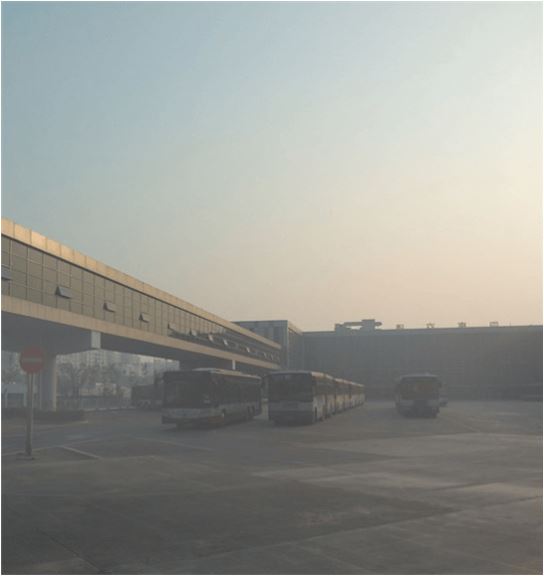} &
			\includegraphics[width=0.107\linewidth]{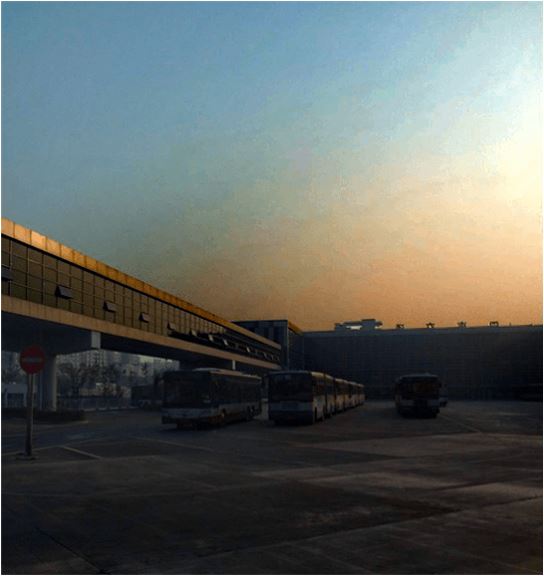} &
			\includegraphics[width=0.107\linewidth]{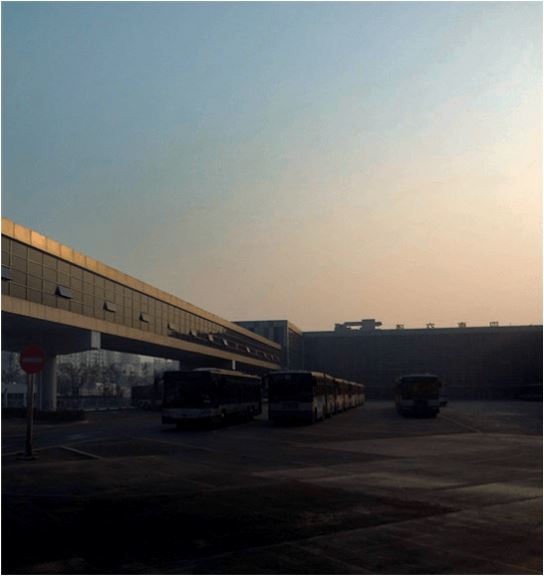} &
			\includegraphics[width=0.107\linewidth]{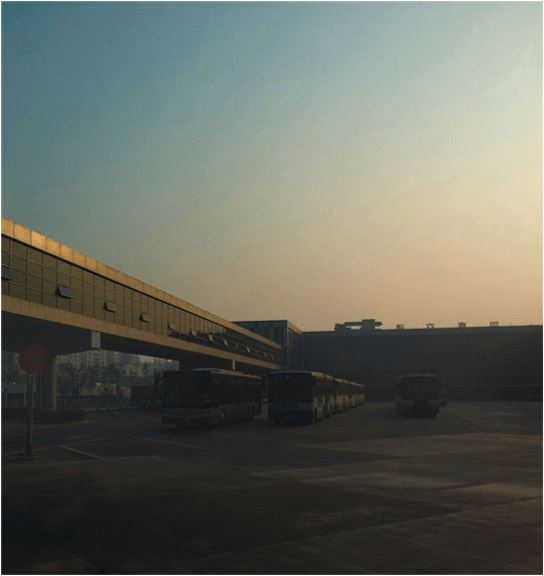} &
			\includegraphics[width=0.107\linewidth]{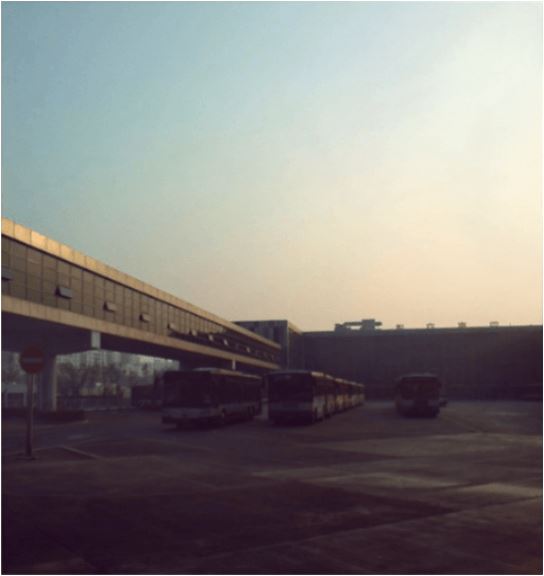} &
			\includegraphics[width=0.107\linewidth]{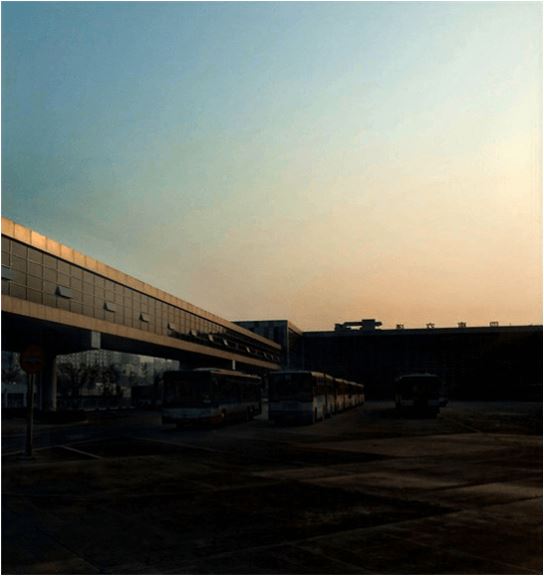} &
			\includegraphics[width=0.107\linewidth]{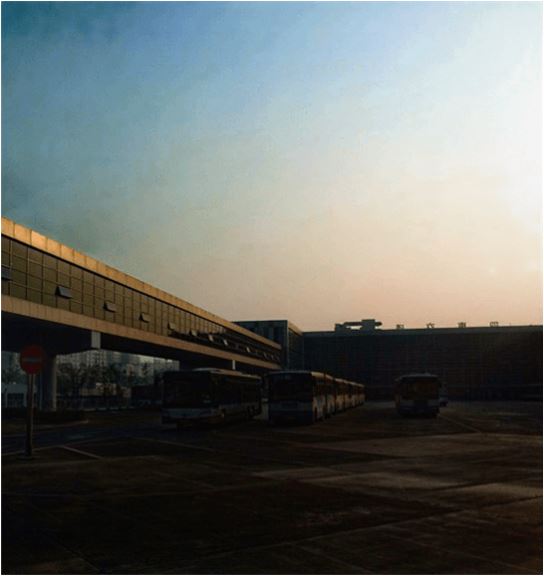} &
			\includegraphics[width=0.107\linewidth]{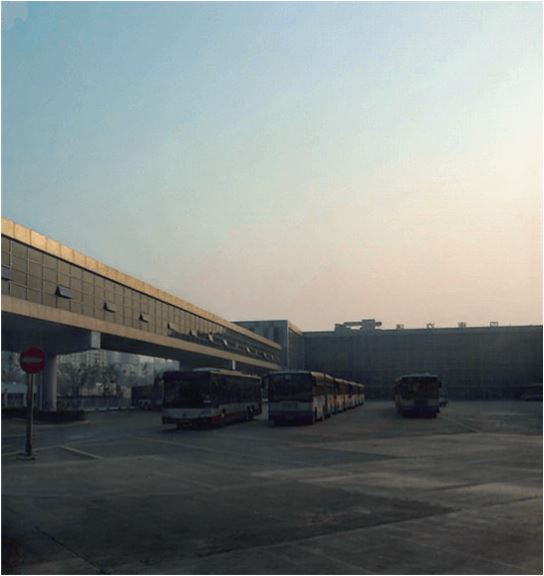} &
			\includegraphics[width=0.107\linewidth]{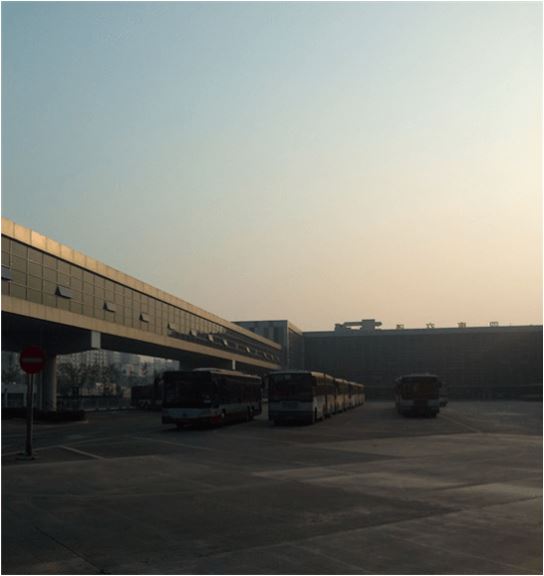} \\
			
			\includegraphics[width=0.107\linewidth]{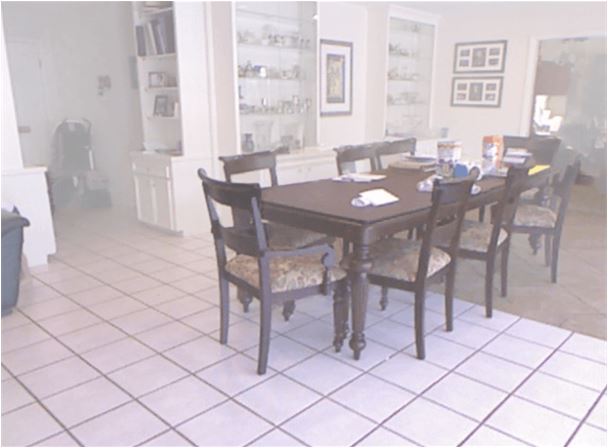} &
			\includegraphics[width=0.107\linewidth]{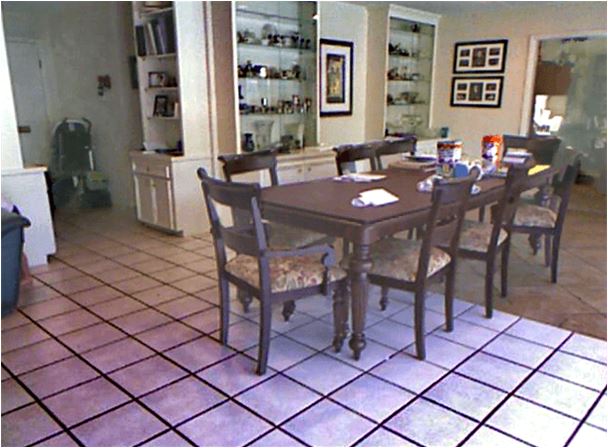} &
			\includegraphics[width=0.107\linewidth]{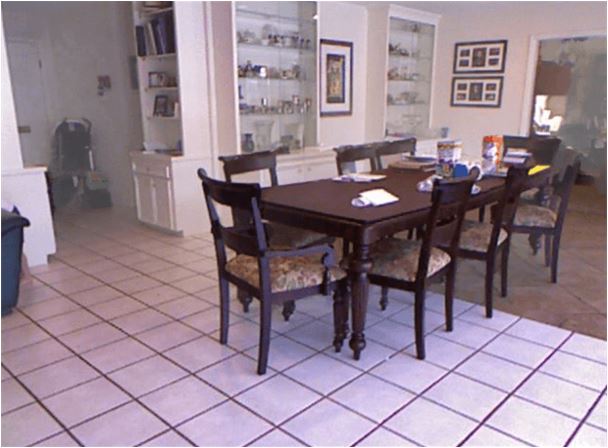} &
			\includegraphics[width=0.107\linewidth]{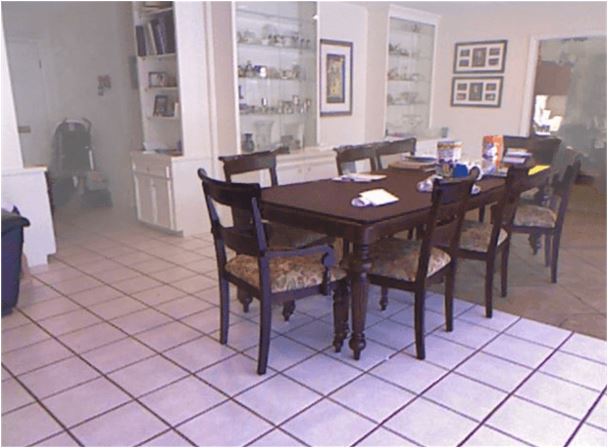} &
			\includegraphics[width=0.107\linewidth]{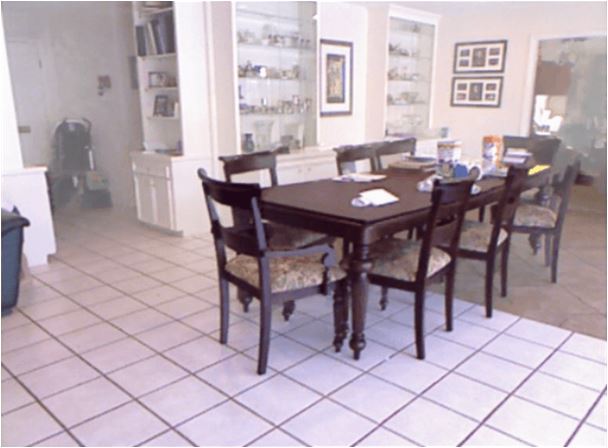} &
			\includegraphics[width=0.107\linewidth]{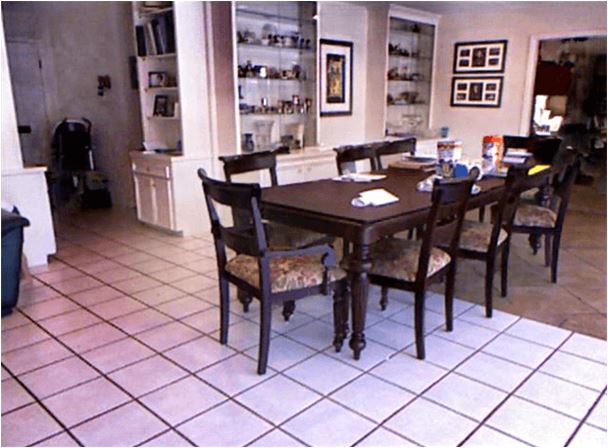} &
			\includegraphics[width=0.107\linewidth]{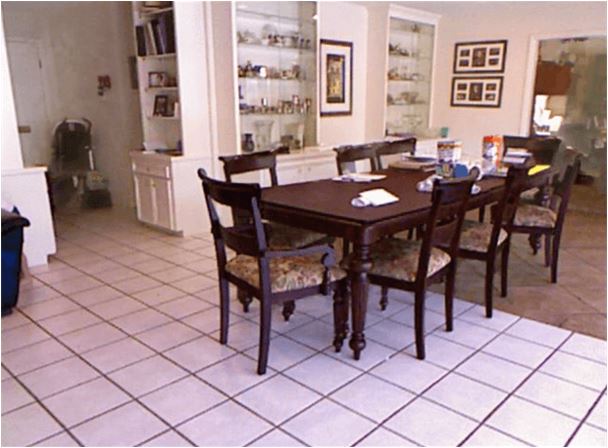} &
			\includegraphics[width=0.107\linewidth]{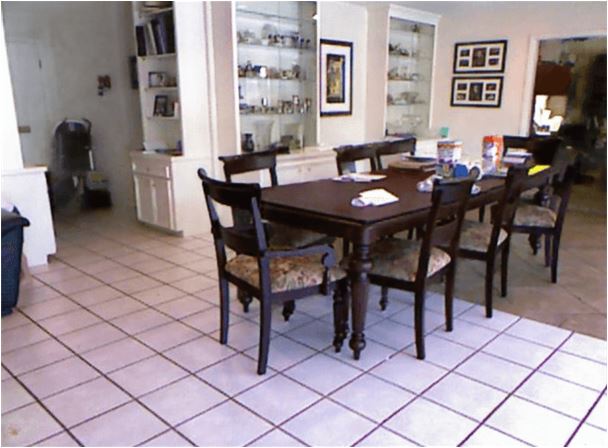} &
			\includegraphics[width=0.107\linewidth]{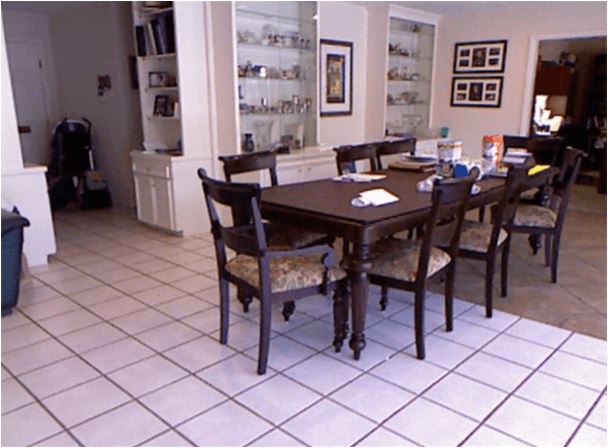} \\
			
			\includegraphics[width=0.107\linewidth]{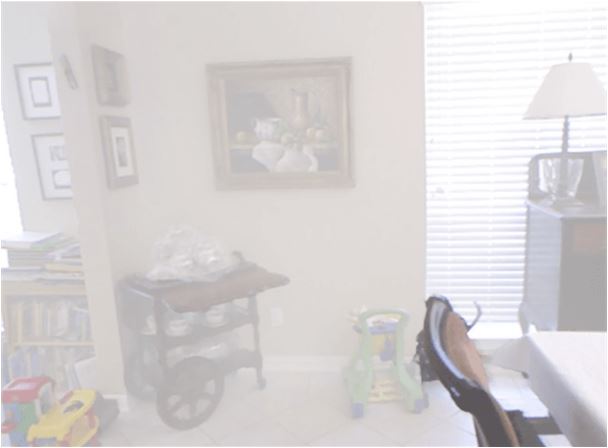} &
			\includegraphics[width=0.107\linewidth]{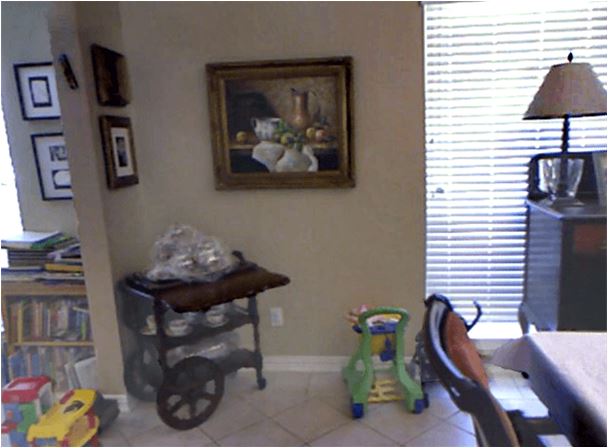} &
			\includegraphics[width=0.107\linewidth]{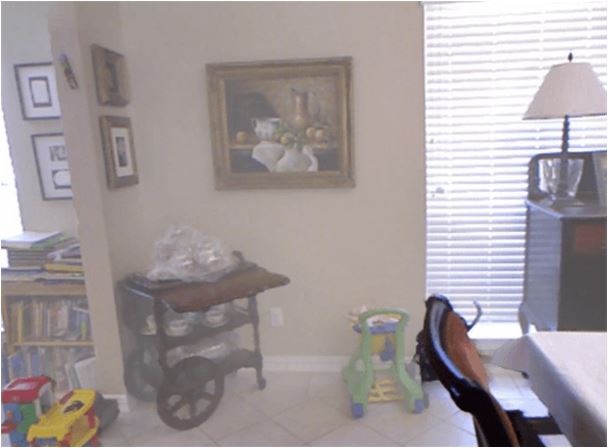} &
			\includegraphics[width=0.107\linewidth]{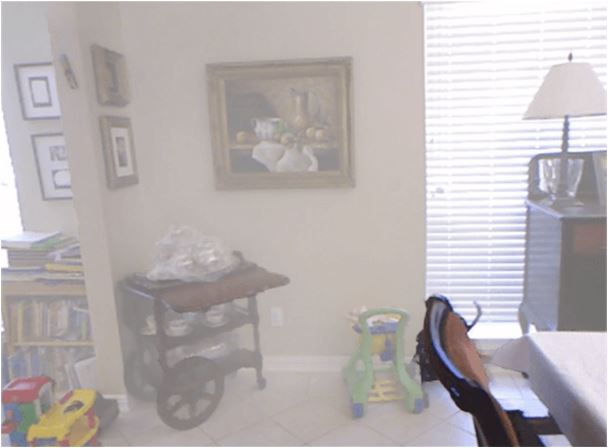} &
			\includegraphics[width=0.107\linewidth]{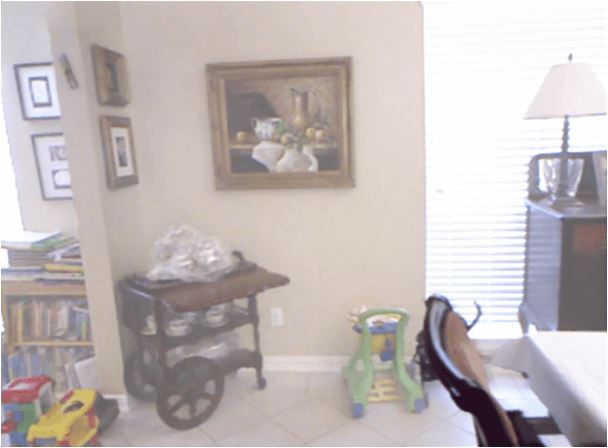} &
			\includegraphics[width=0.107\linewidth]{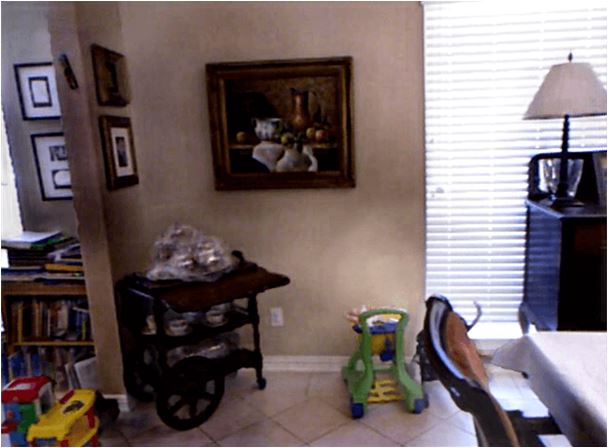} &
			\includegraphics[width=0.107\linewidth]{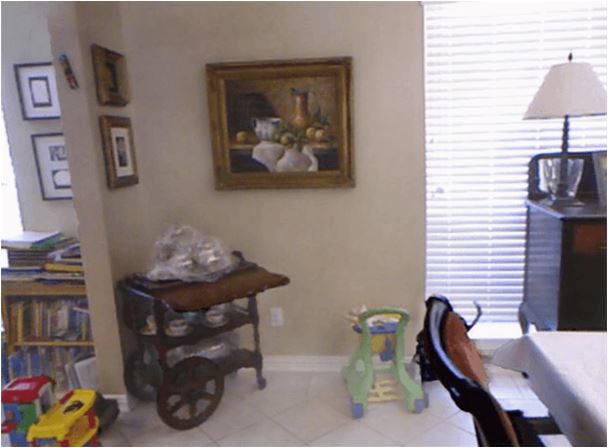} &
			\includegraphics[width=0.107\linewidth]{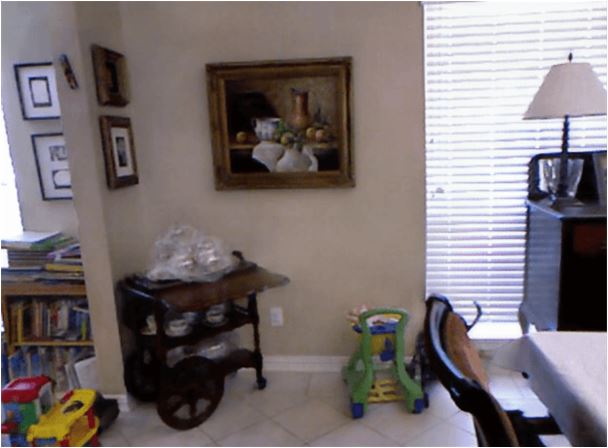} &
			\includegraphics[width=0.107\linewidth]{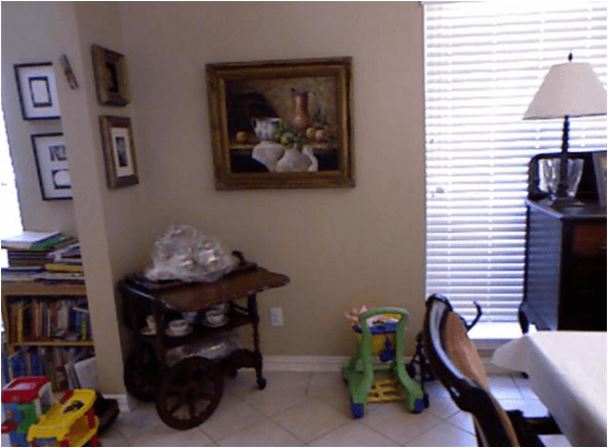} \\
			
			\includegraphics[width=0.107\linewidth]{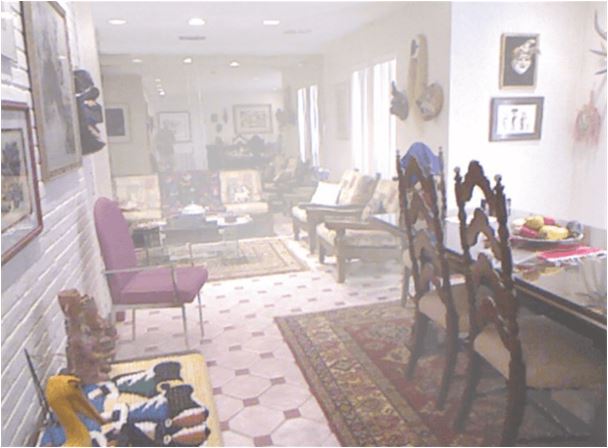} &
			\includegraphics[width=0.107\linewidth]{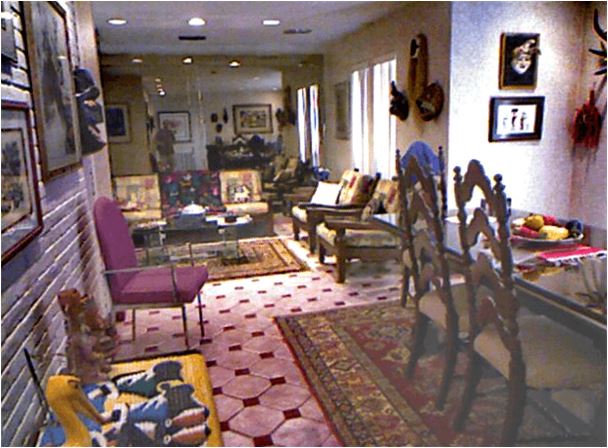} &
			\includegraphics[width=0.107\linewidth]{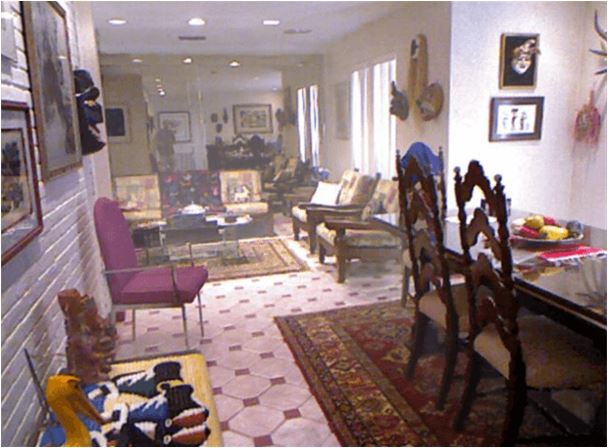} &
			\includegraphics[width=0.107\linewidth]{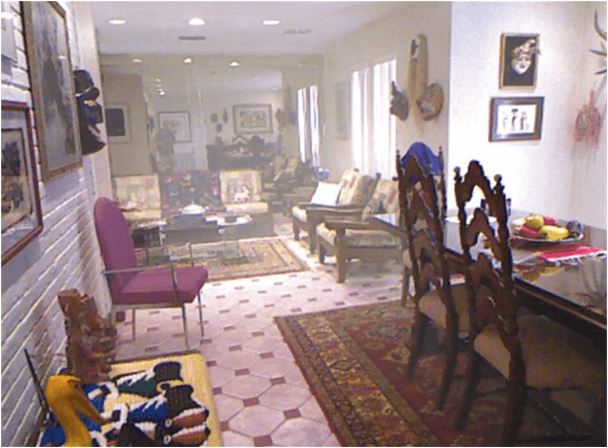} &
			\includegraphics[width=0.107\linewidth]{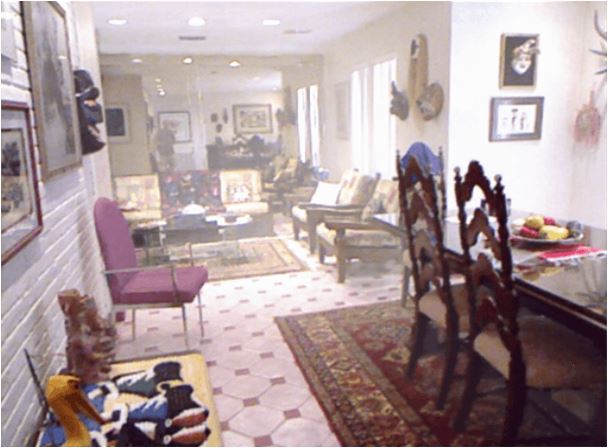} &
			\includegraphics[width=0.107\linewidth]{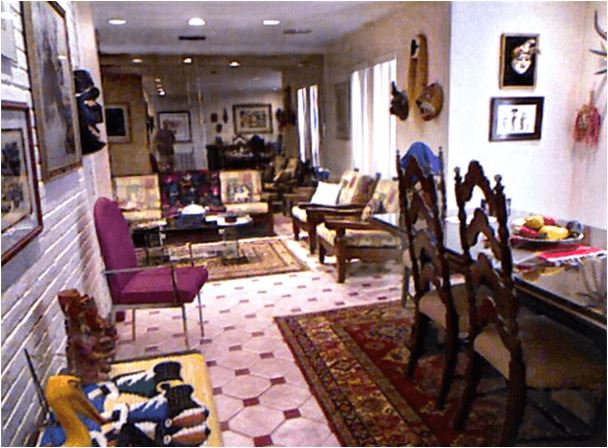} &
			\includegraphics[width=0.107\linewidth]{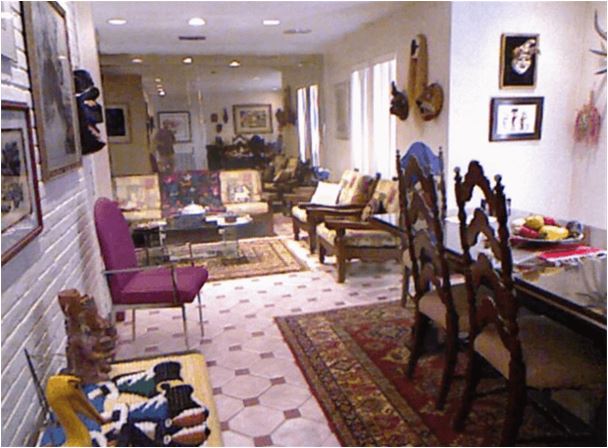} &
			\includegraphics[width=0.107\linewidth]{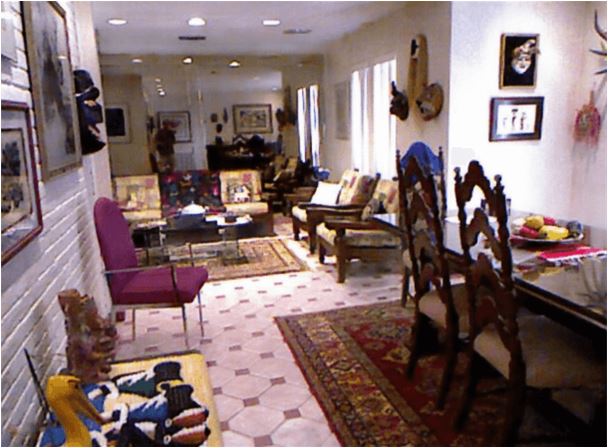} &
			\includegraphics[width=0.107\linewidth]{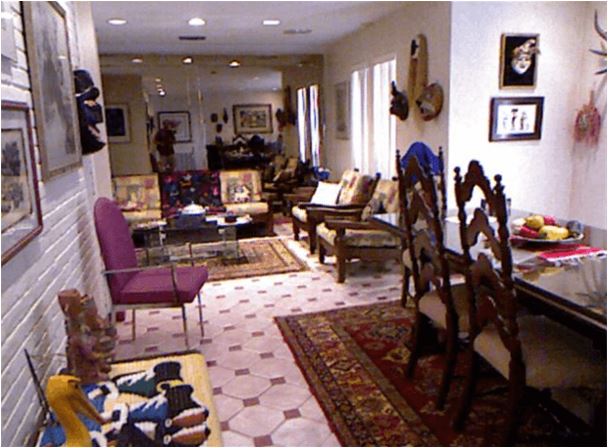} \\
			
			(a) Hazy image&
			(b) NLD~\cite{berman2016non}&
			(c) DehazeNet~\cite{Cai2016DehazeNet} & 
			(d) AOD-Net~\cite{li2017aod} &
			(e) DCPDN~\cite{Zhang_2018_CVPR} &
			(f) GFN~\cite{Ren_2018_CVPR} &
			(g) EPDN~\cite{qu2019enhanced} &
			(h) Ours  &
			(i) Clear image \\
	\end{tabular}
	\end{center}
	\vspace{-2mm}
	\caption{Visual comparisons on the SOTS~\cite{li2019benchmarking} dataset.}
	\label{fig:SOTS}
\end{figure*}
\begin{figure*}[htbp]
	\scriptsize
	\centering
	\renewcommand{\tabcolsep}{1pt} 
	\renewcommand{\arraystretch}{1} 
	\begin{center}
		\begin{tabular}{ccccccccc}
			\includegraphics[width=0.107\linewidth]{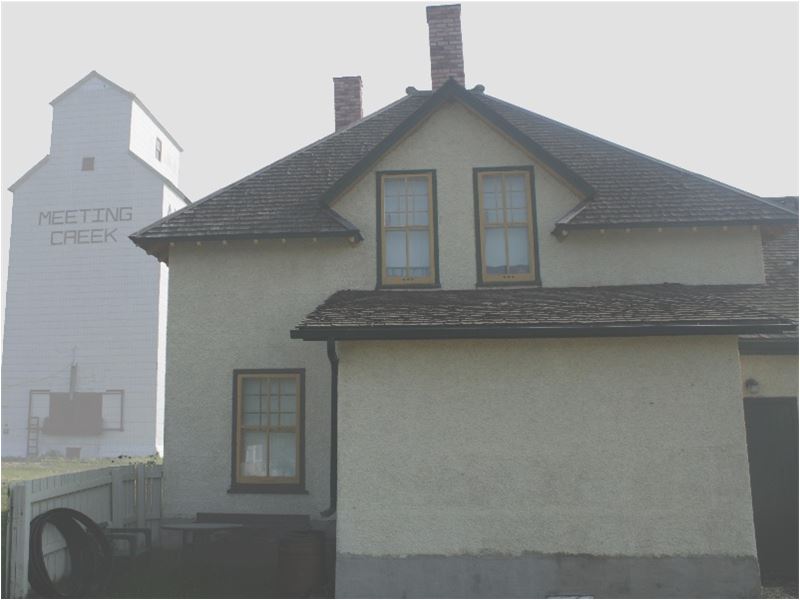} &
			\includegraphics[width=0.107\linewidth]{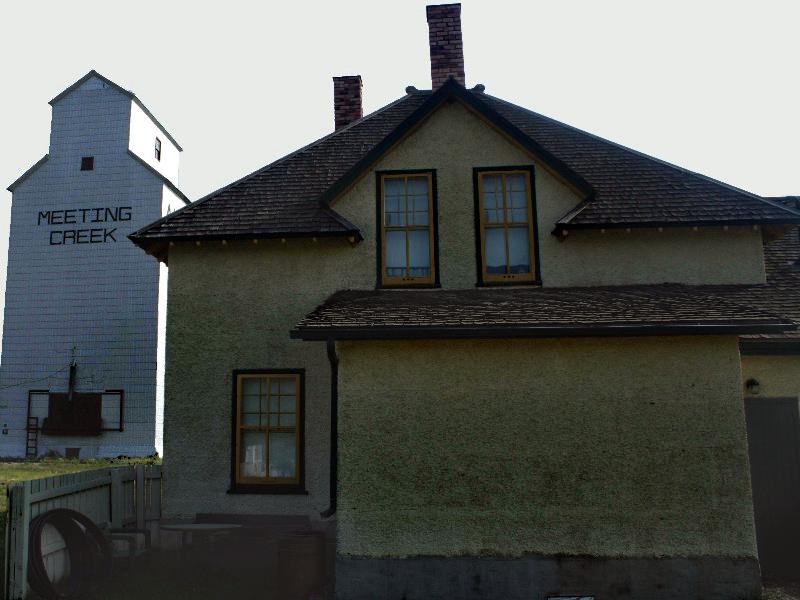} &
			\includegraphics[width=0.107\linewidth]{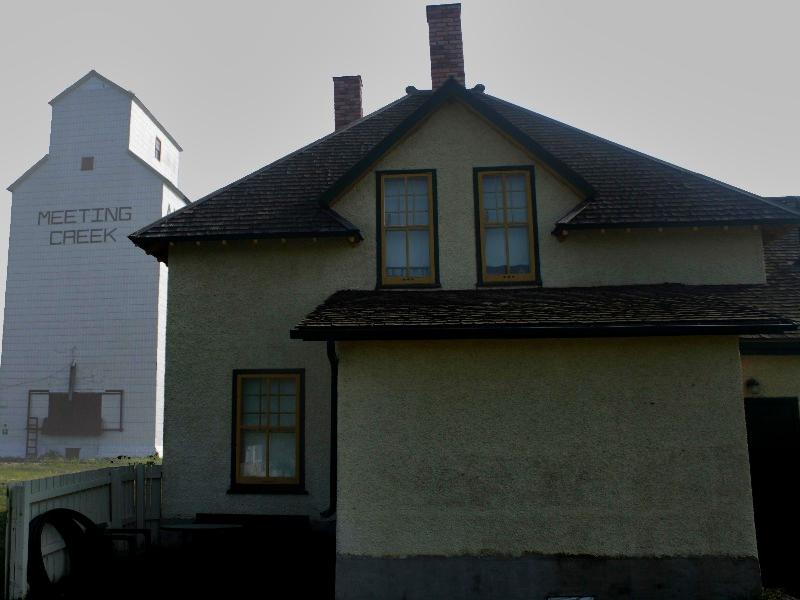} &
			\includegraphics[width=0.107\linewidth]{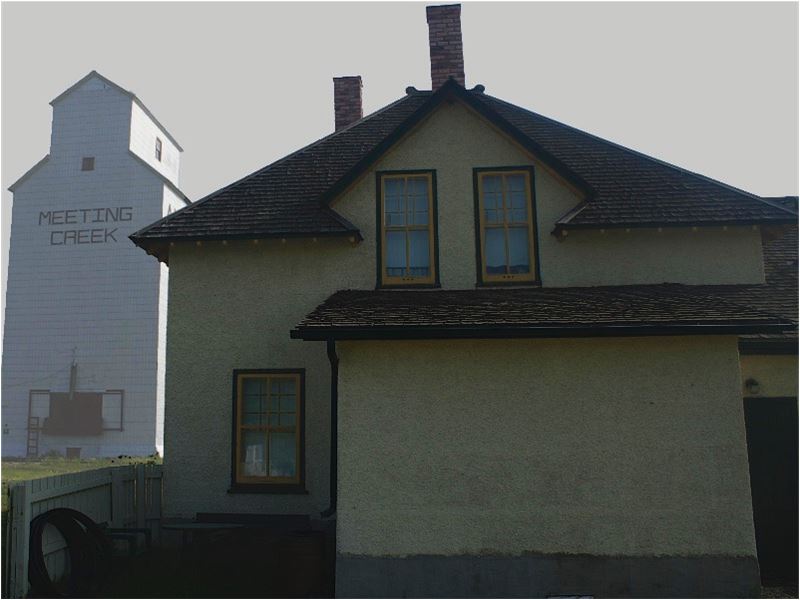} &			
			\includegraphics[width=0.107\linewidth]{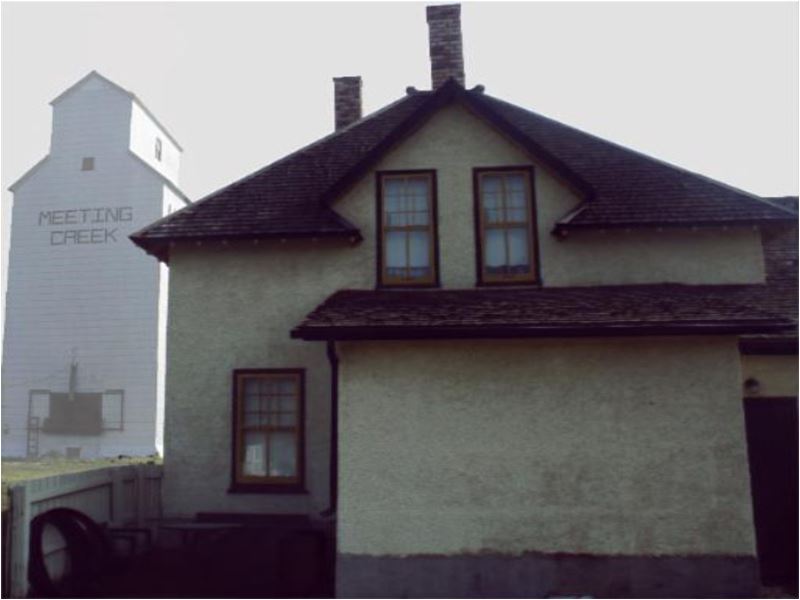} &
			\includegraphics[width=0.107\linewidth]{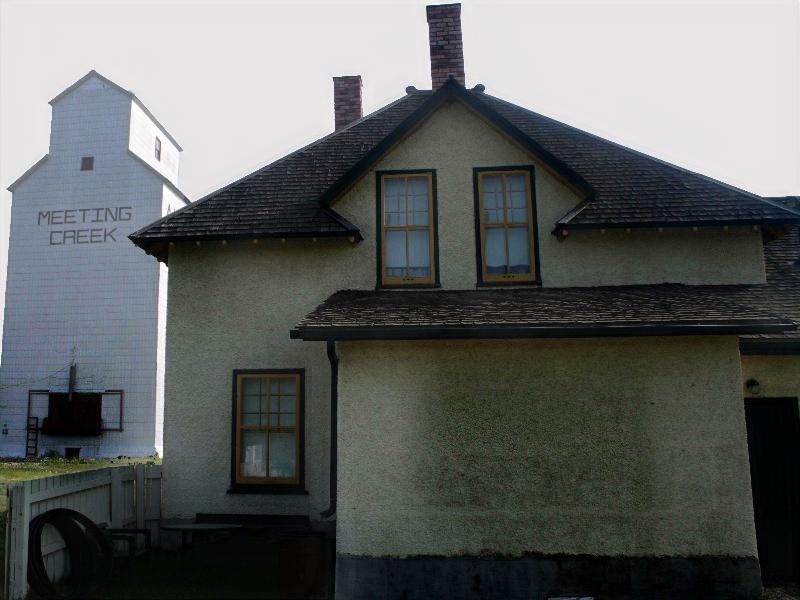} &
			\includegraphics[width=0.107\linewidth]{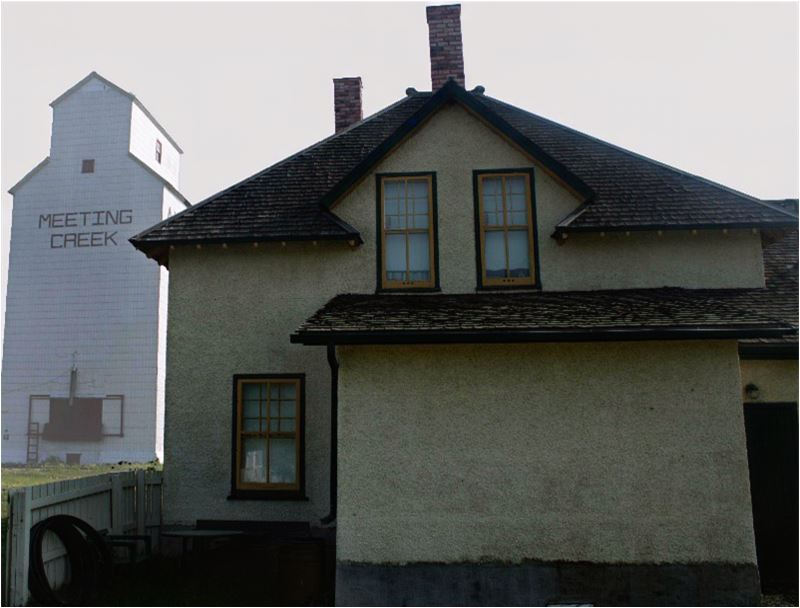} &
			\includegraphics[width=0.107\linewidth]{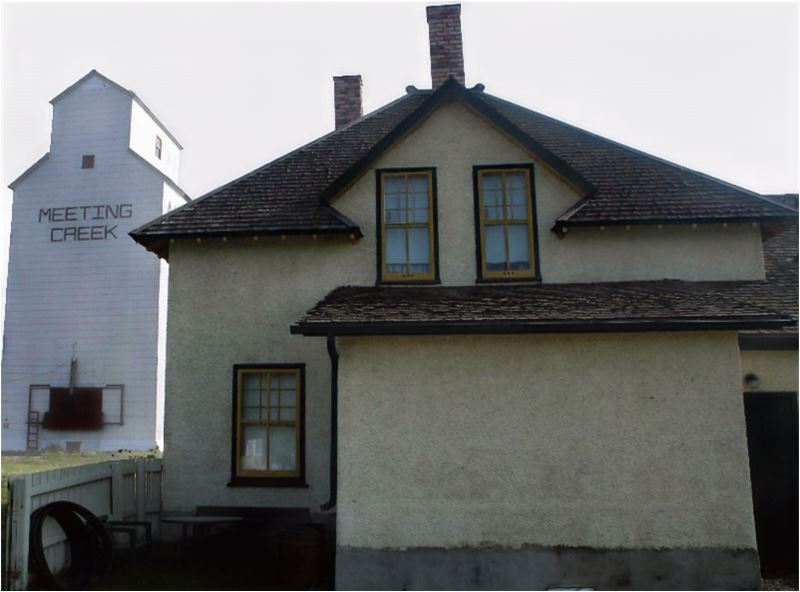} &
			\includegraphics[width=0.107\linewidth]{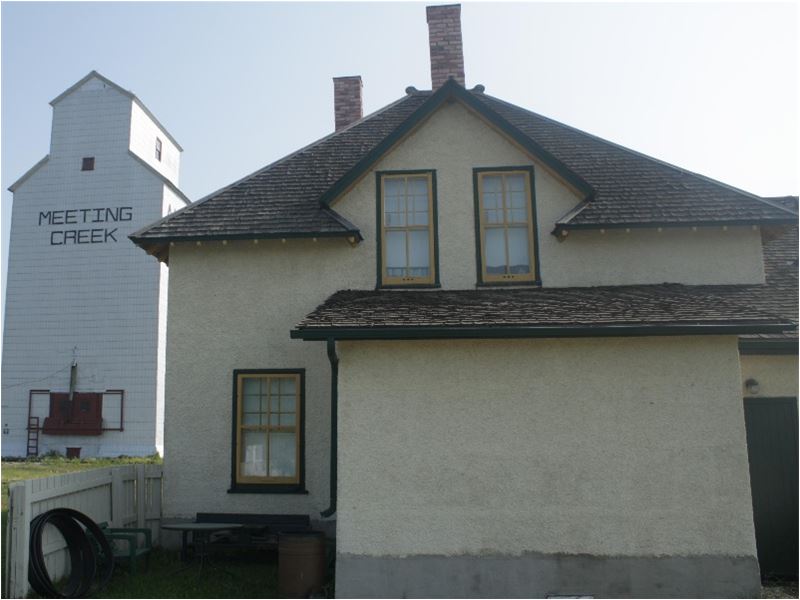} \\
			
			\includegraphics[width=0.107\linewidth]{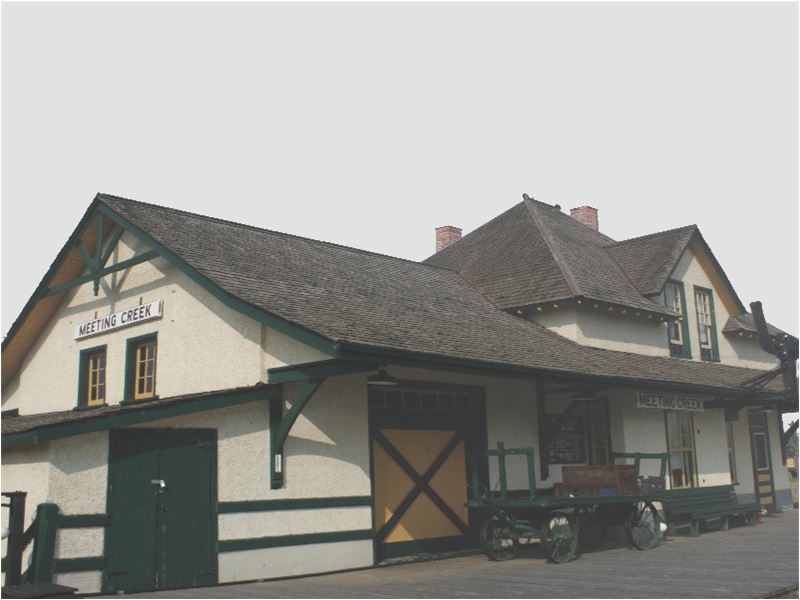} &
			\includegraphics[width=0.107\linewidth]{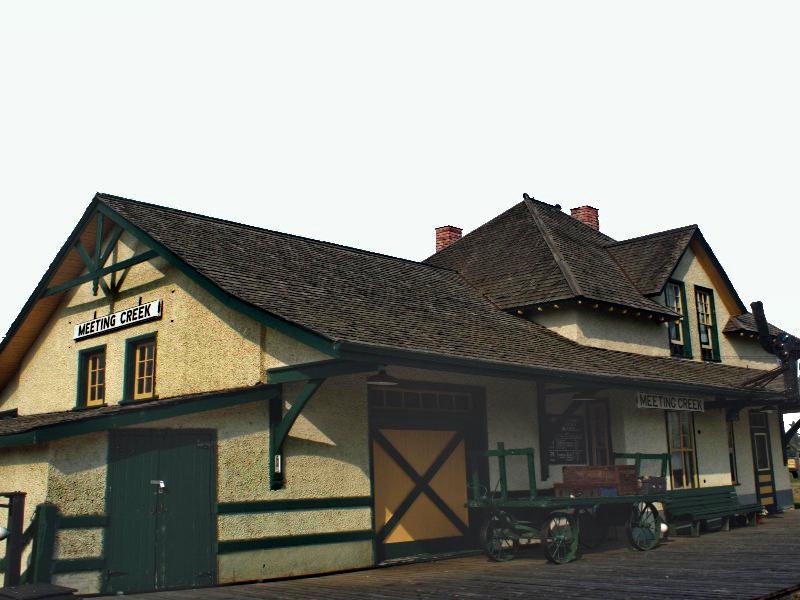} &
			\includegraphics[width=0.107\linewidth]{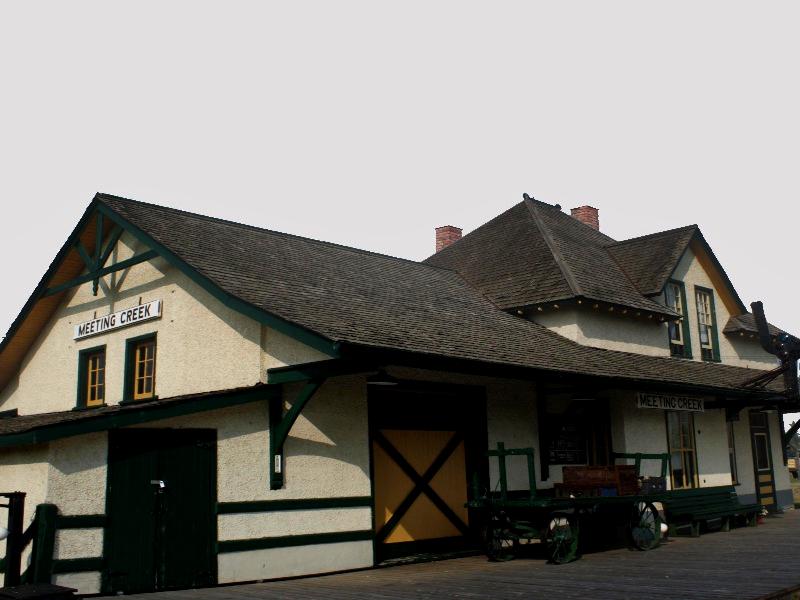} &
			\includegraphics[width=0.107\linewidth]{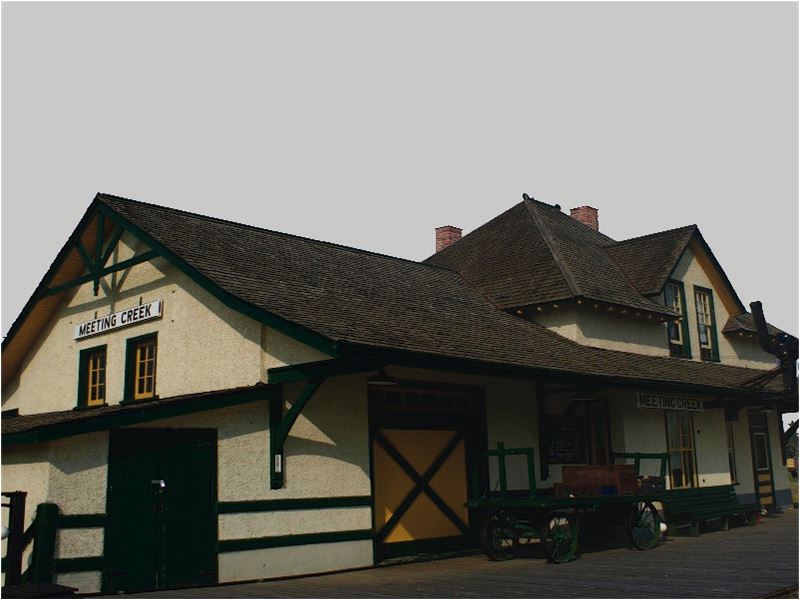} &			
			\includegraphics[width=0.107\linewidth]{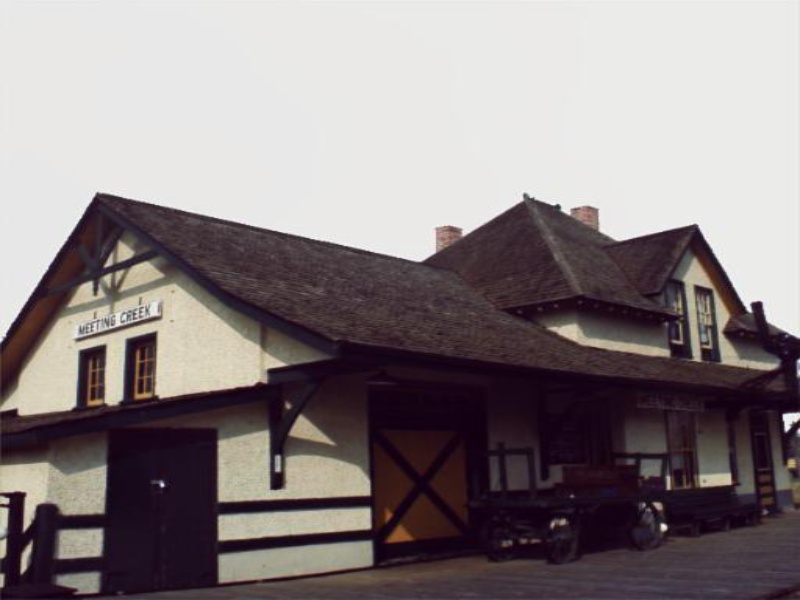} &
			\includegraphics[width=0.107\linewidth]{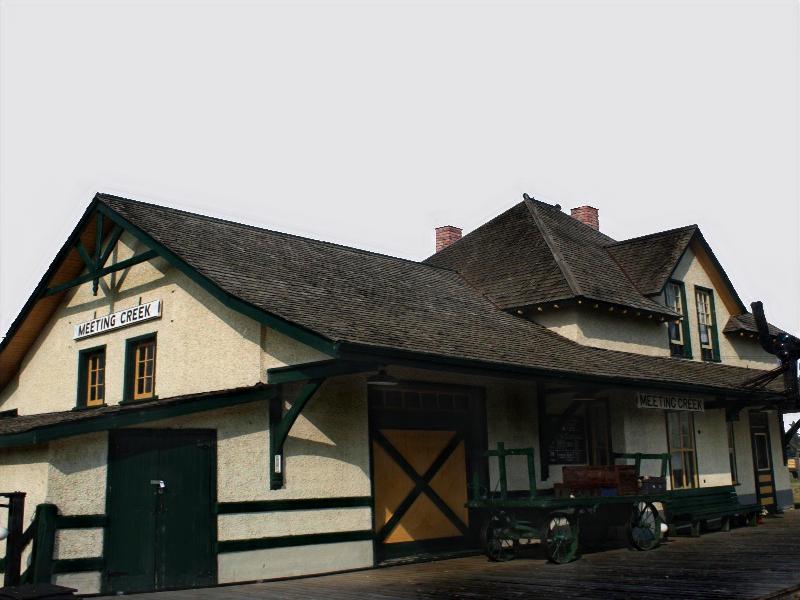} &
			\includegraphics[width=0.107\linewidth]{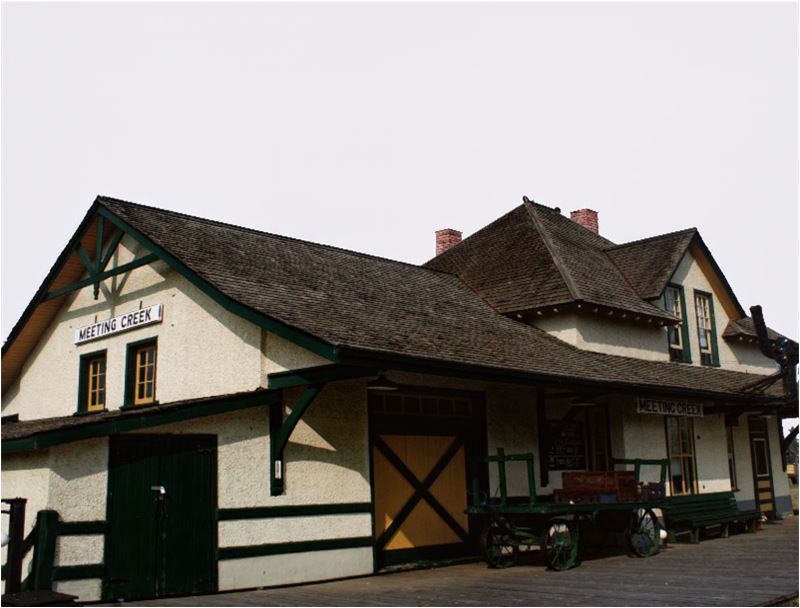} &
			\includegraphics[width=0.107\linewidth]{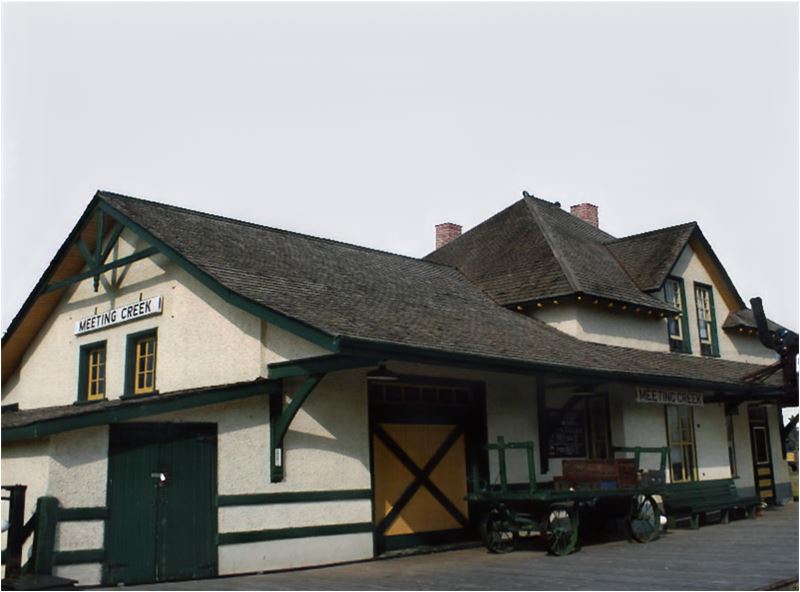} &
			\includegraphics[width=0.107\linewidth]{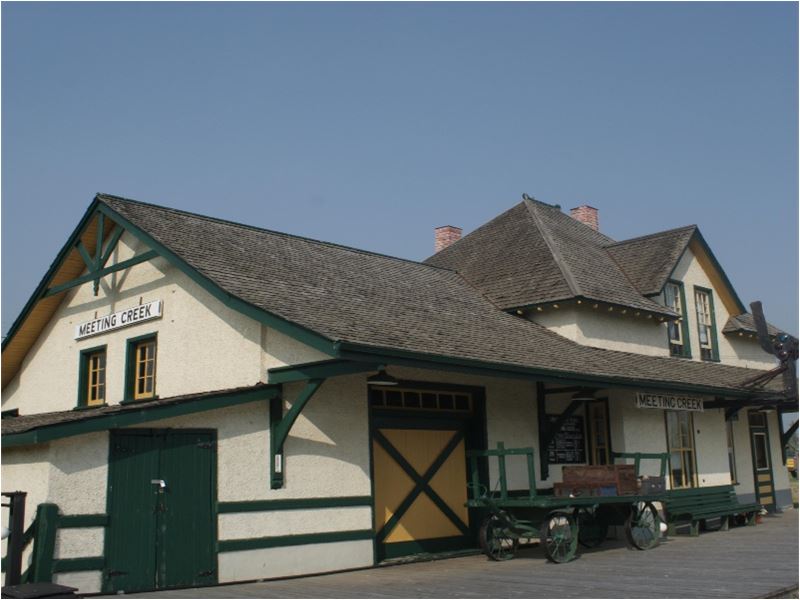} \\
			
			(a) Hazy image&
			(b) NLD~\cite{berman2016non}&
			(c) DehazeNet~\cite{Cai2016DehazeNet} & 
			(d) AOD-Net~\cite{li2017aod} &
			(e) DCPDN~\cite{Zhang_2018_CVPR} &
			(f) GFN~\cite{Ren_2018_CVPR} &
			(g) EPDN~\cite{qu2019enhanced} &
			(h) Ours  &
			(i) Clear image \\
		\end{tabular}
	\end{center}
	\vspace{-2mm}
	\caption{Visual comparisons on the HazeRD~\cite{Zhang:HazeRD:ICIP17b} dataset.}
	\label{fig:HazeRD}
\end{figure*}
\begin{table*}[htbp]
    \footnotesize
	\centering
	\caption{Quantitative comparison (Average PSNR/SSIM) of the dehazing results on two synthetic datasets.}
	\vspace{1mm}
	\label{tab:Syn}
	\resizebox{\linewidth}{!}{
		\begin{tabular}{ccccccccccc}
			\toprule
			&    & DCP~\cite{He2011Single} & NLD~\cite{berman2016non}  & MSCNN~\cite{ren2016single} & DehazeNet~\cite{Cai2016DehazeNet}  & AOD-Net~\cite{li2017aod}    & DCPDN~\cite{Zhang_2018_CVPR}      & GFN~\cite{Ren_2018_CVPR}        &  EPDN~\cite{qu2019enhanced}  & Ours       \\ \midrule
			& SOTS & 15.49/0.64 & 17.27/0.75  & 17.57/0.81 & 21.14/0.85 & 19.06/0.85 & 19.39/0.65 & 22.30/0.88 &  23.82/0.89  & \textbf{27.76}/\textbf{0.93} \\ \midrule
			& HazeRD & 14.01/0.39 & 16.16/0.58  & 15.57/0.42 & 15.54/0.41 & 15.63/0.45 & 16.12/0.34 & 13.98/0.37 &  17.37/0.56  & \textbf{18.07}/\textbf{0.63} \\ \bottomrule
			\vspace{-5mm}
		\end{tabular}
	}
\end{table*}

\section{Experimental Results}
\label{sec:experiments}
In this part, we first present the implementation details of our framework.
Then, we evaluate our domain adaptation method on synthetic datasets and the real images, respectively.
Finally, ablation studies are conducted to analyze the proposed approach.

\subsection{Implementation Details}
\label{sec:details}
\vspace{-1.1mm}
\paragraph{Datasets.}
We randomly choose both synthetic and real-word hazy images from the RESIDE dataset~\cite{li2019benchmarking} for training.
The dataset is divided into five subsets, namely, ITS (Indoor Training Set), OTS (Outdoor Training Set), SOTS (Synthetic Object Testing Set), URHI (Unannotated real Hazy Images), and RTTS (real Task-driven Testing Set).
For synthetic dataset, we choose 6000 synthetic hazy images for training, 3000 from the ITS and 3000 from the OTS.
For real-world hazy images, we train the network by randomly selecting 1000 real hazy images from the URHI.
In the training phase, we randomly crop all the images to $256\times256$ and normalize the pixel values to $\left [ -1, 1 \right ]$.
%
%
\vspace{-4mm}
\paragraph{Training details.}
We implement our framework in PyTorch~\cite{nishino2012bayesian} and utilize ADAM~\cite{kingma2014adam} optimizer with a batch size $2$ to train the network.
First, we train the image translation network ${G_{S \to R}}$ and ${G_{R \to S}}$ for 90 epochs with the momentum ${\beta _1} = 0.5 ,\ {\beta _2} = 0.999$, and the learning rate is set as $5 \times {10^{ - 5}}$.
Then we train ${{\cal G}_R}$ on $\{ {X_R},{G_{S \to R}}({X_{\rm{s}}},{D_{\rm{s}}})\} $, and ${{\cal G}_S}$ on $\{ {X_S},{G_{R \to S}}({X_R})\} $ for 90 epochs using the pre-trained  ${G_{S \to R}}$ and ${G_{R \to S}}$ models.
The momentum and the learning rate are set as:
${\beta _1} = 0.95,\ {\beta _2} = 0.999,\ lr = {10^{ - 4}}$. 
Finally, we fine tune the whole network using the above pre-trained models.
When computing the DC loss, we set the patch as $35\times35$.
The trade-off weights are set as: $\lambda_{tran} = 1,\ \lambda_m  = 10,\ \lambda_d  = {10^{ - 2}},\ \lambda_t  = {10^{ - 3}}$ and $\lambda_c  = {10^{ - 1}}$.
\begin{figure*}[htbp]
	\scriptsize
	\centering
	\renewcommand{\tabcolsep}{1pt} 
	\renewcommand{\arraystretch}{1} 
	\begin{center}
		\begin{tabular}{cccccccc}
			\includegraphics[width=0.12\linewidth]{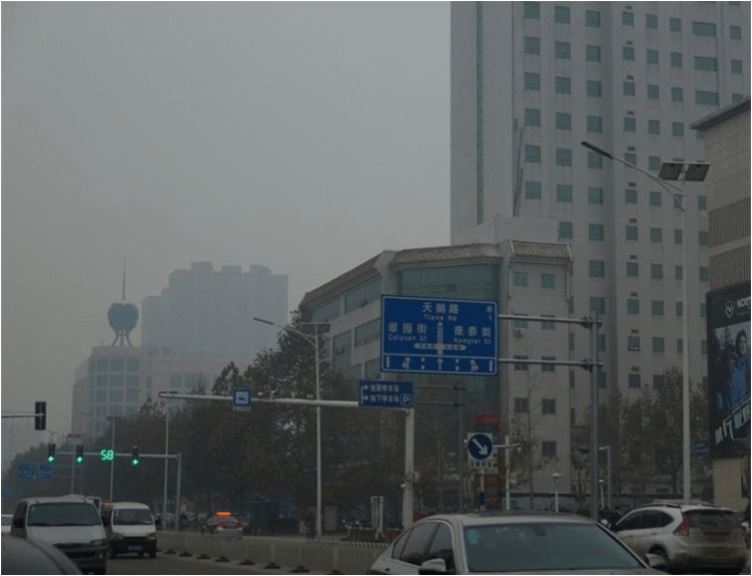} &
			\includegraphics[width=0.12\linewidth]{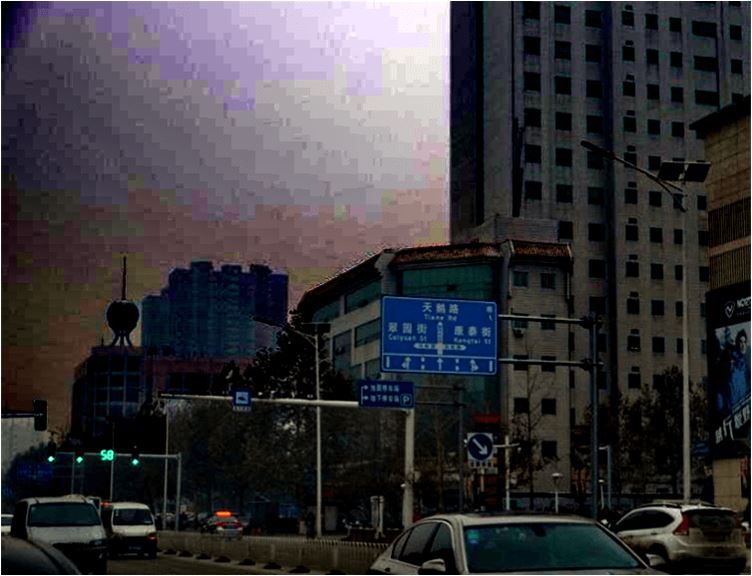} &
			\includegraphics[width=0.12\linewidth]{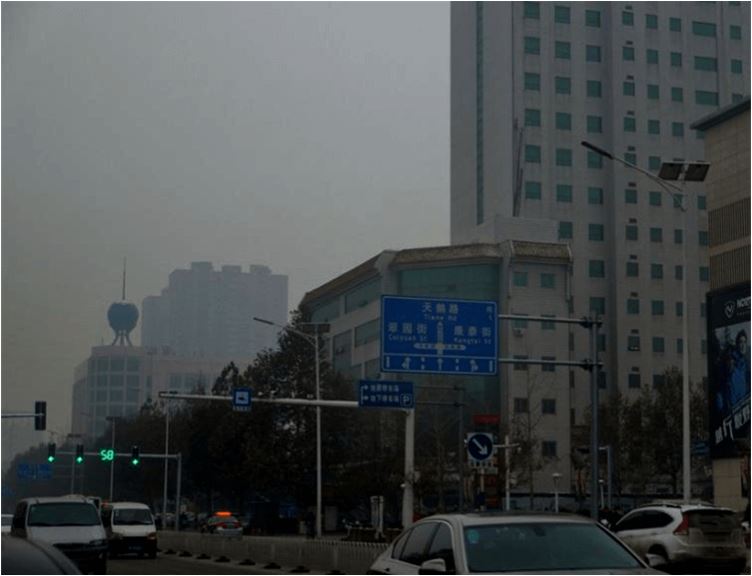} &
			\includegraphics[width=0.12\linewidth]{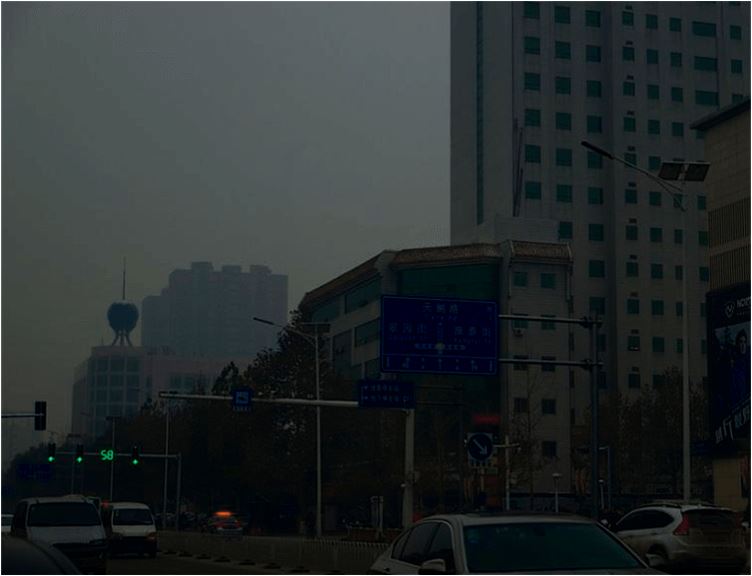} &			
			\includegraphics[width=0.12\linewidth]{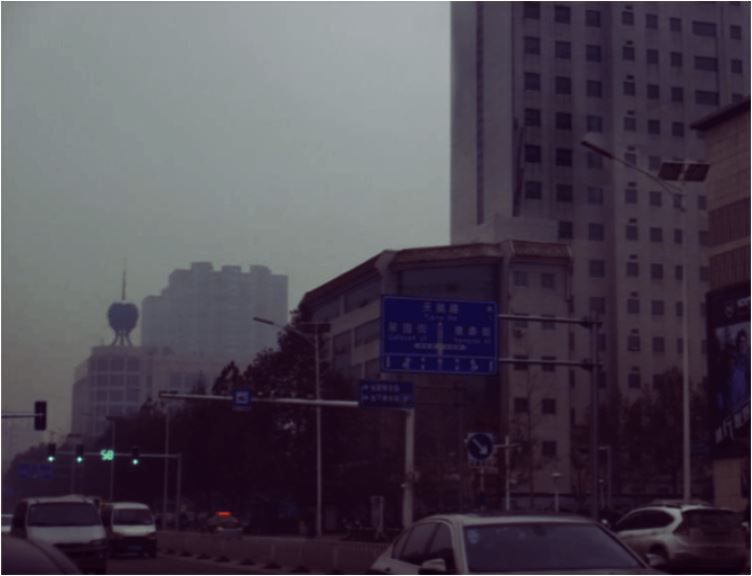} &
			\includegraphics[width=0.12\linewidth]{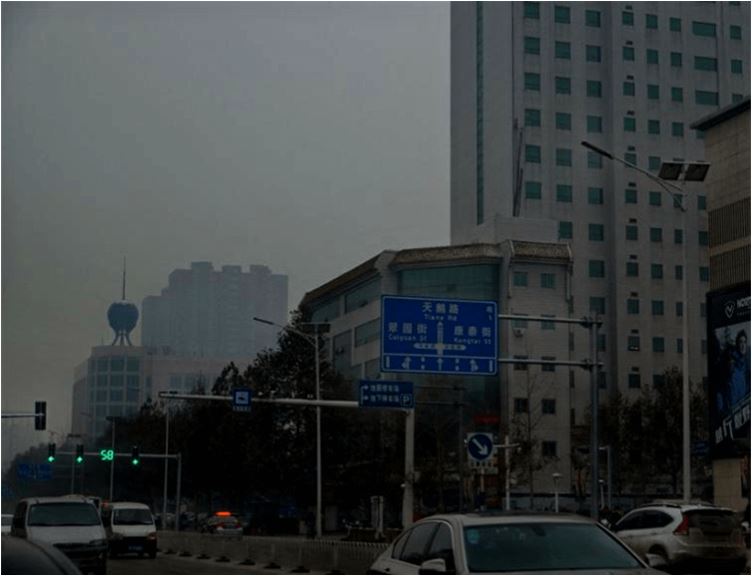} &
			\includegraphics[width=0.12\linewidth]{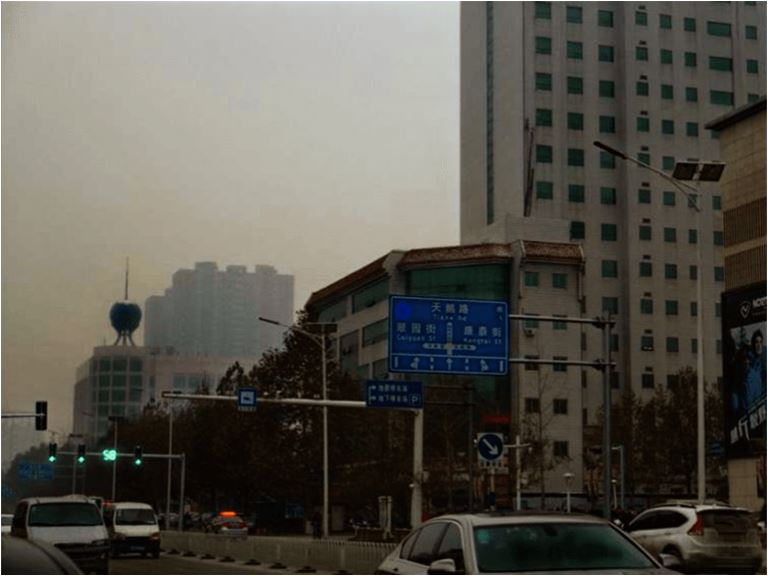} &
			\includegraphics[width=0.12\linewidth]{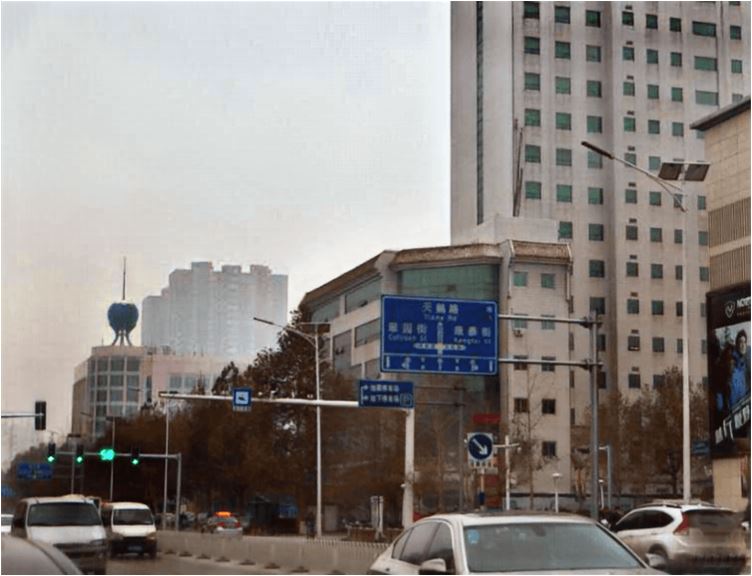} \\
			
			\includegraphics[width=0.12\linewidth]{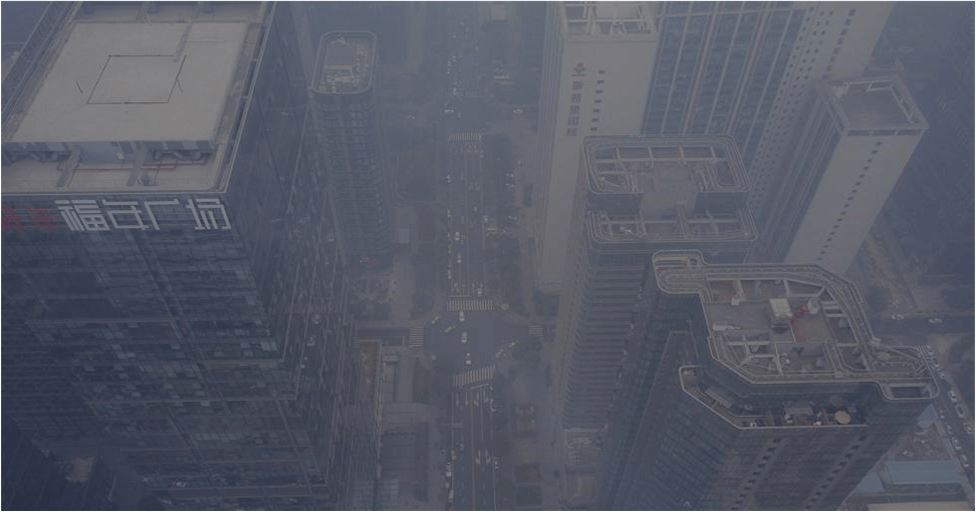} &
			\includegraphics[width=0.12\linewidth]{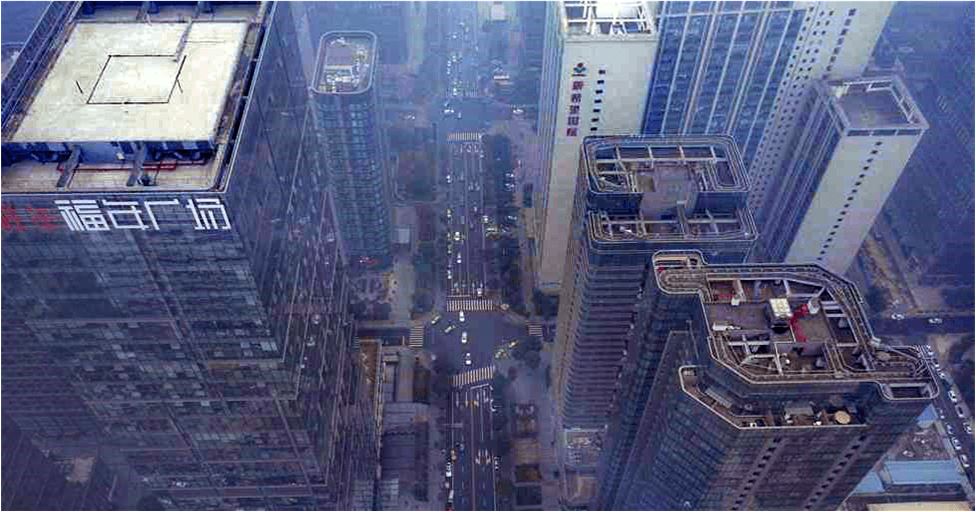} &
			\includegraphics[width=0.12\linewidth]{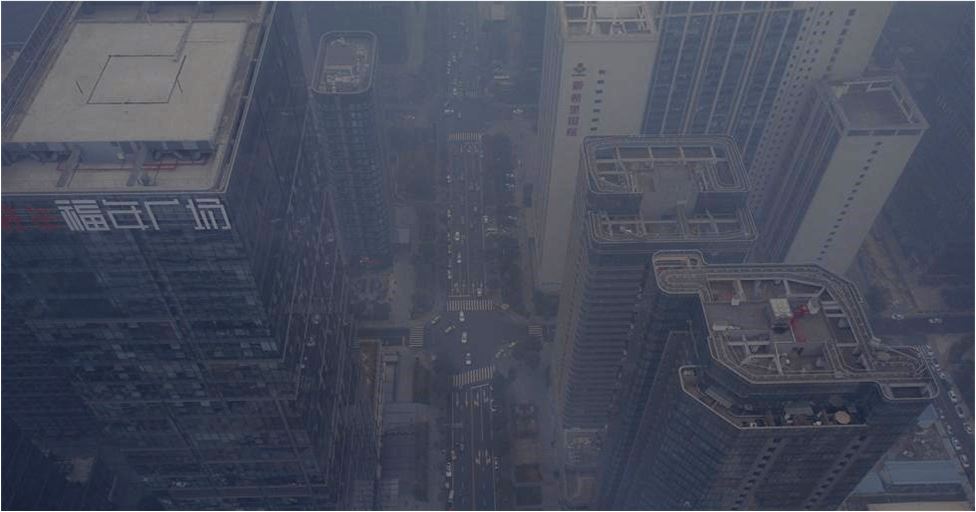} &
			\includegraphics[width=0.12\linewidth]{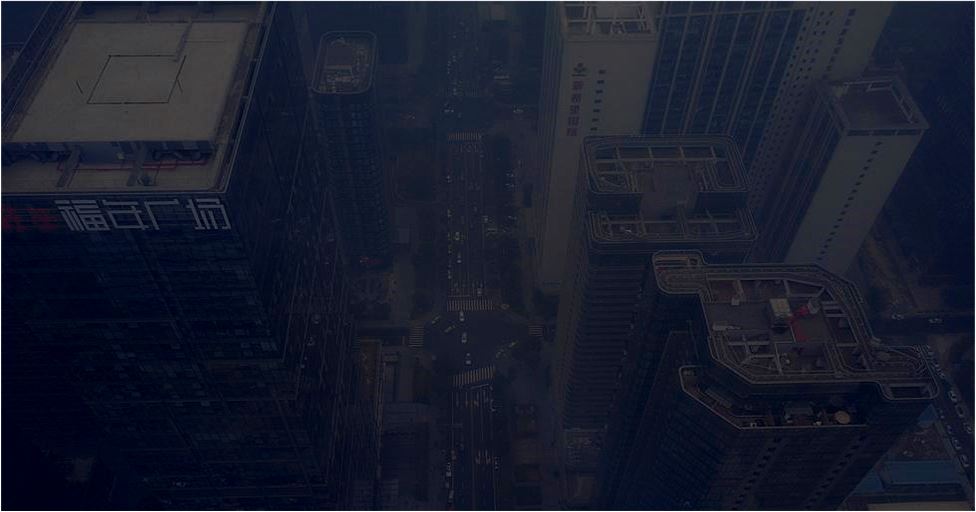} &
			\includegraphics[width=0.12\linewidth]{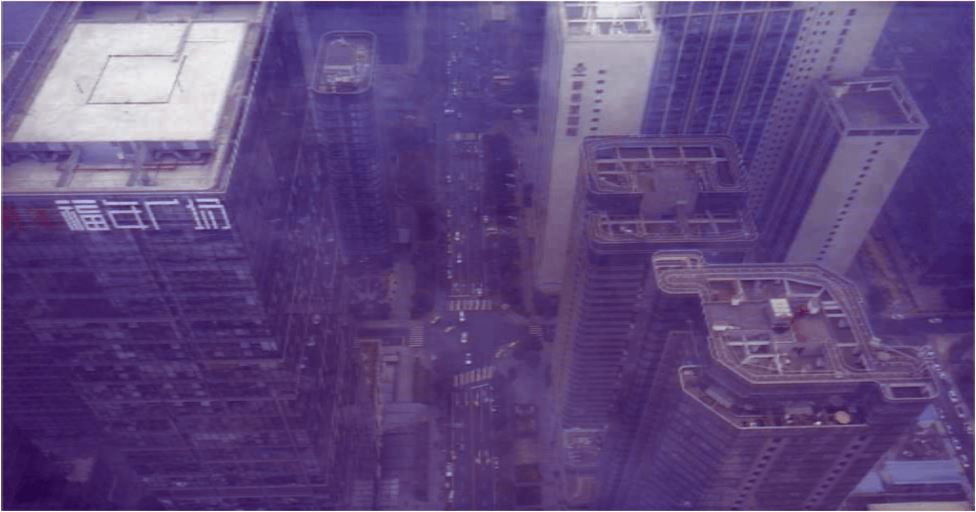} &
			\includegraphics[width=0.12\linewidth]{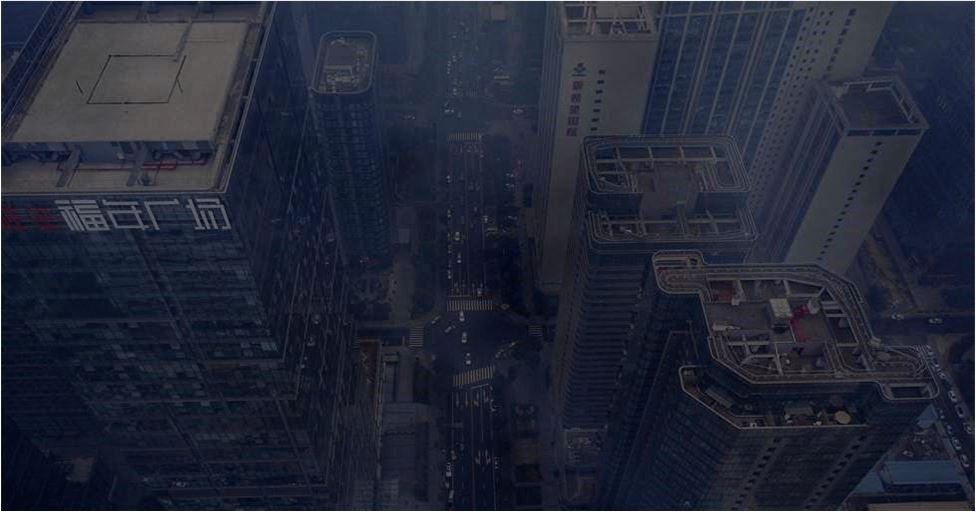} &
			\includegraphics[width=0.12\linewidth]{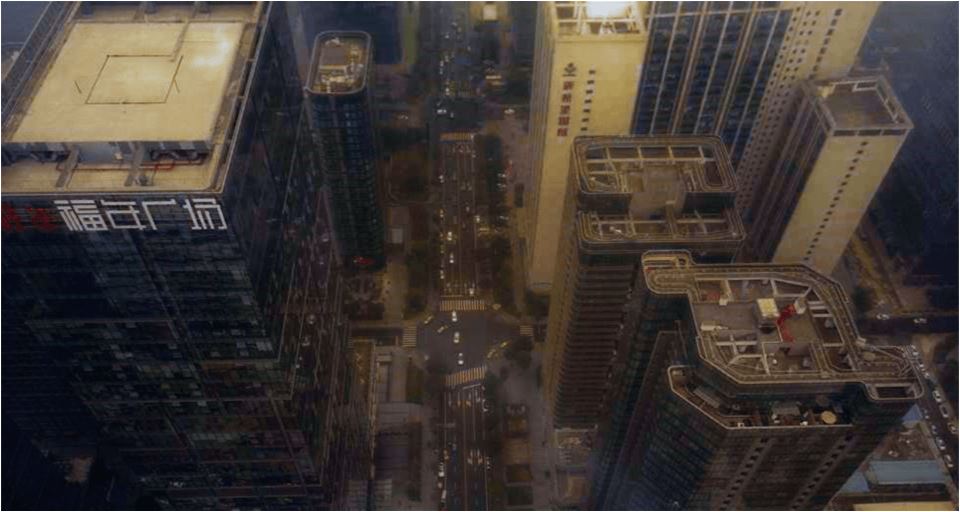} &
			\includegraphics[width=0.12\linewidth]{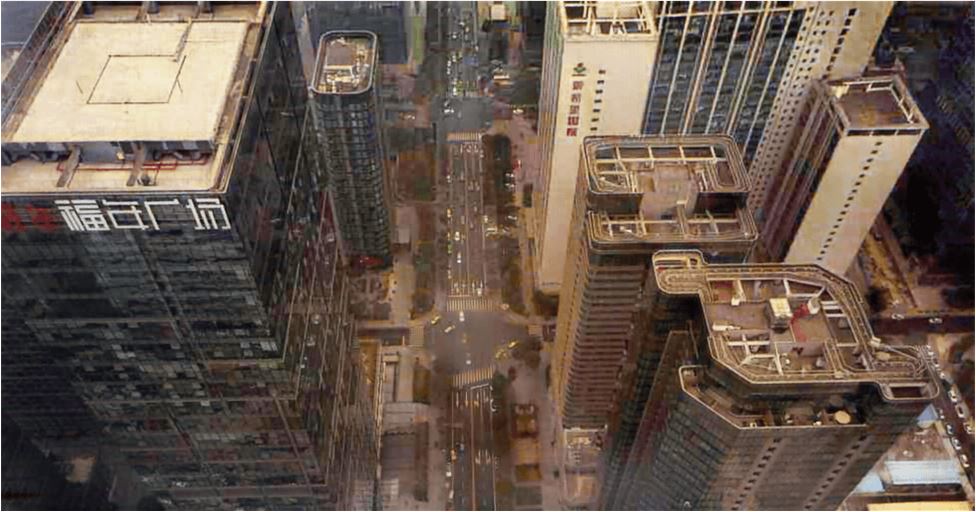} \\
			
			\includegraphics[width=0.12\linewidth]{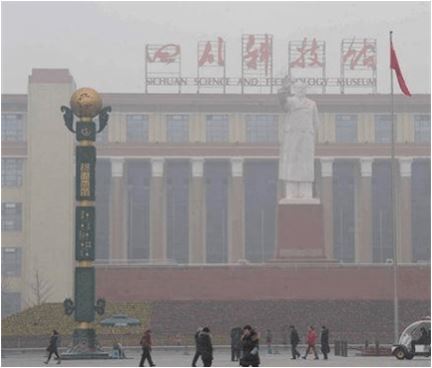} &
			\includegraphics[width=0.12\linewidth]{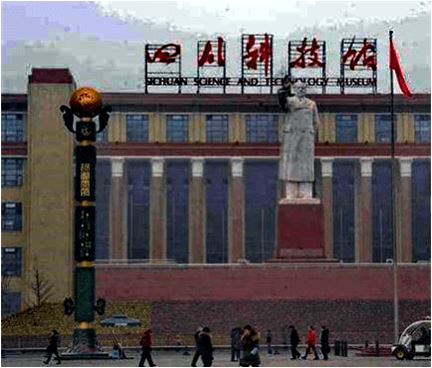} &
			\includegraphics[width=0.12\linewidth]{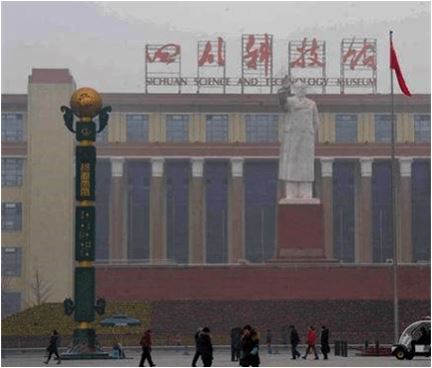} &
			\includegraphics[width=0.12\linewidth]{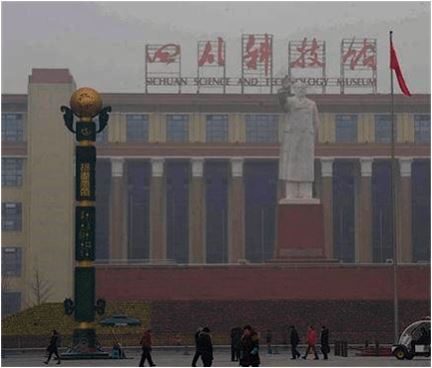} &
			\includegraphics[width=0.12\linewidth]{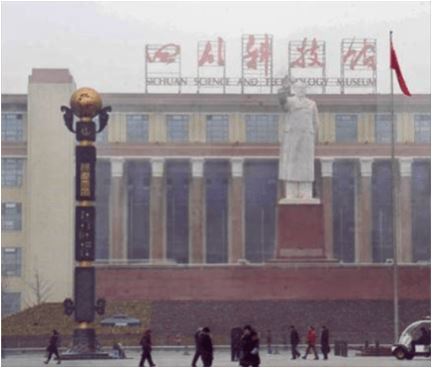} &
			\includegraphics[width=0.12\linewidth]{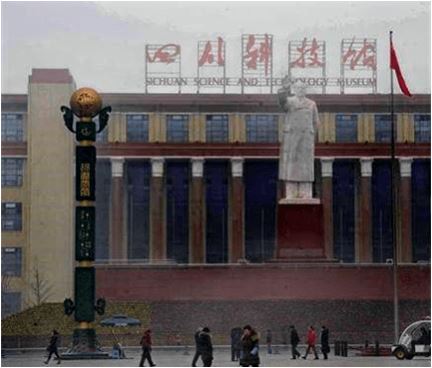} &
			\includegraphics[width=0.12\linewidth]{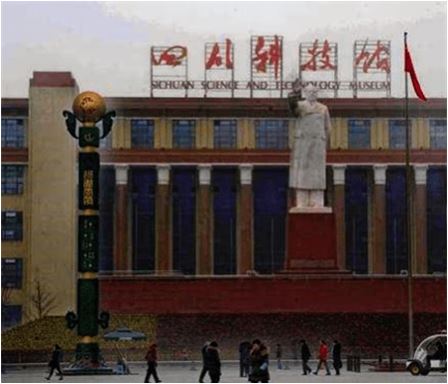} &
			\includegraphics[width=0.12\linewidth]{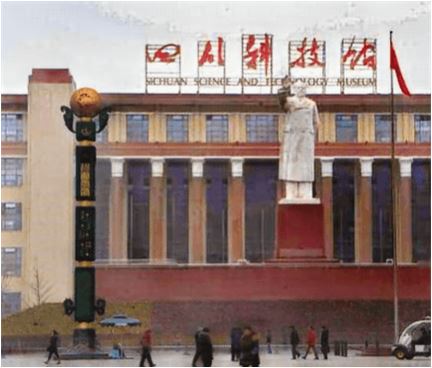} \\
			
			\includegraphics[width=0.12\linewidth]{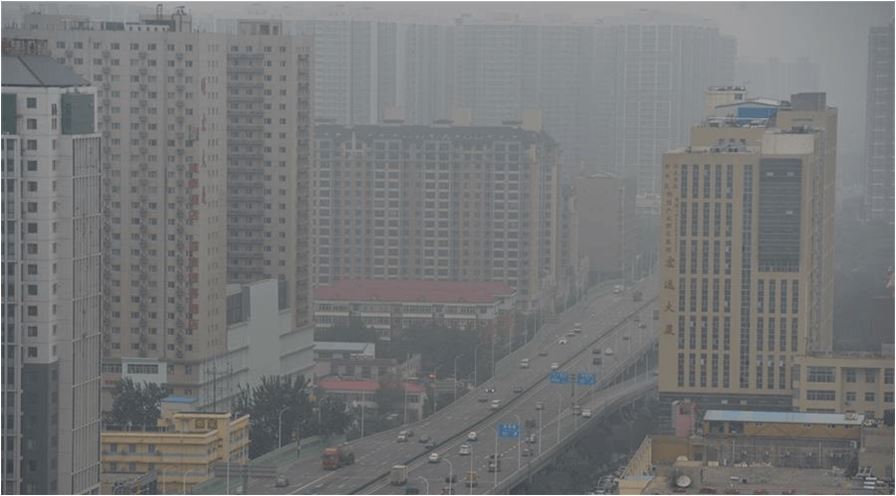} &
			\includegraphics[width=0.12\linewidth]{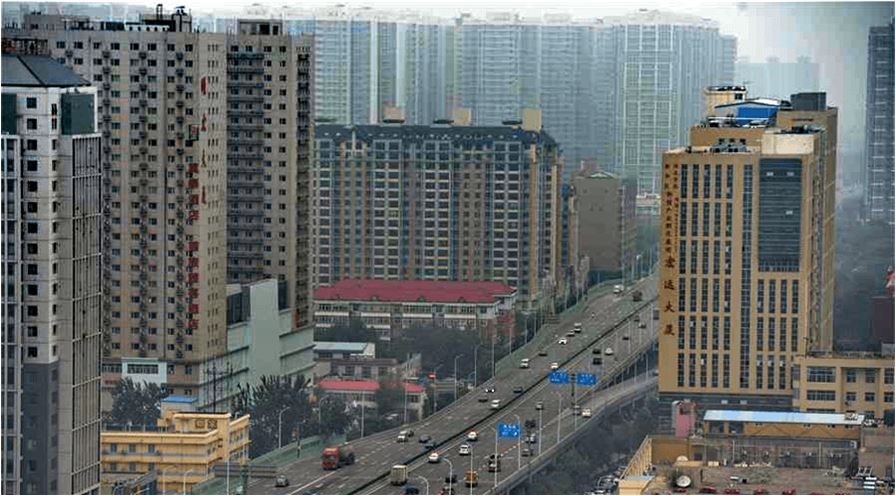} &
			\includegraphics[width=0.12\linewidth]{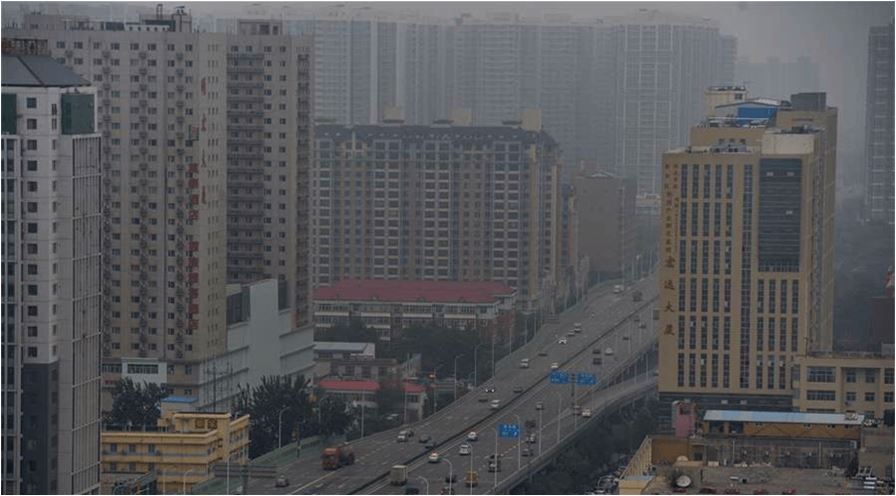} &
			\includegraphics[width=0.12\linewidth]{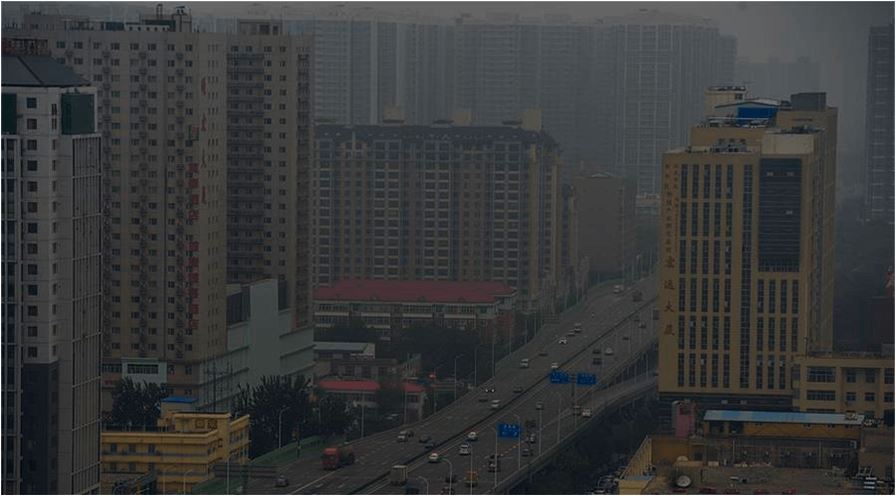} &
			\includegraphics[width=0.12\linewidth]{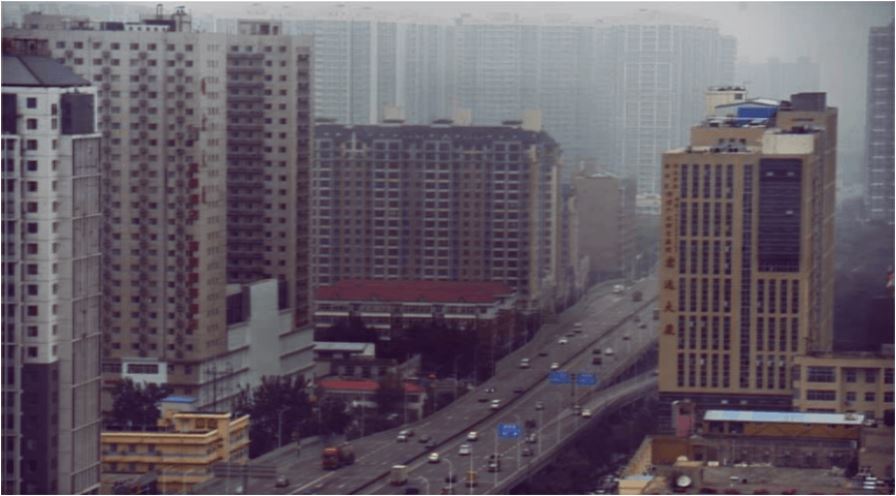} &
			\includegraphics[width=0.12\linewidth]{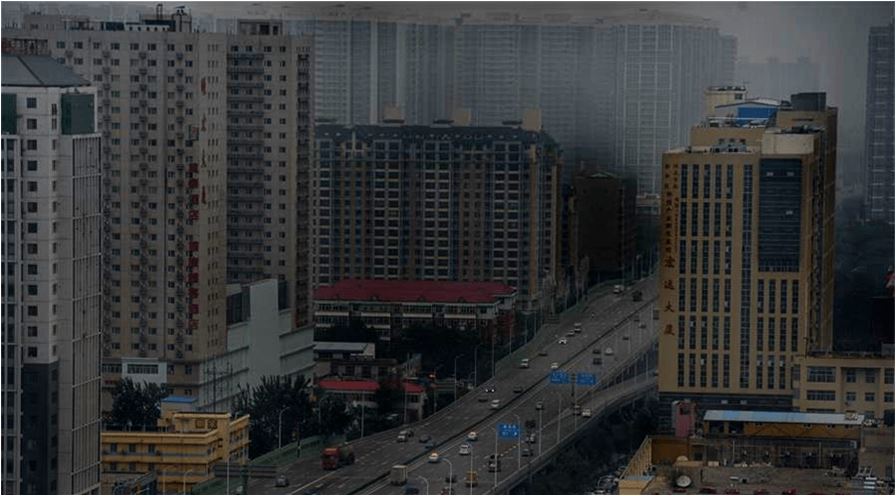} &
			\includegraphics[width=0.12\linewidth]{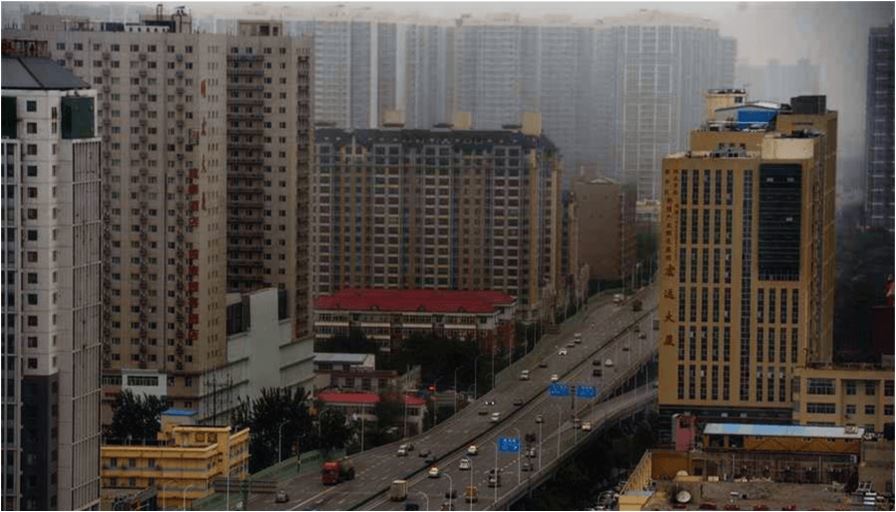} &
			\includegraphics[width=0.12\linewidth]{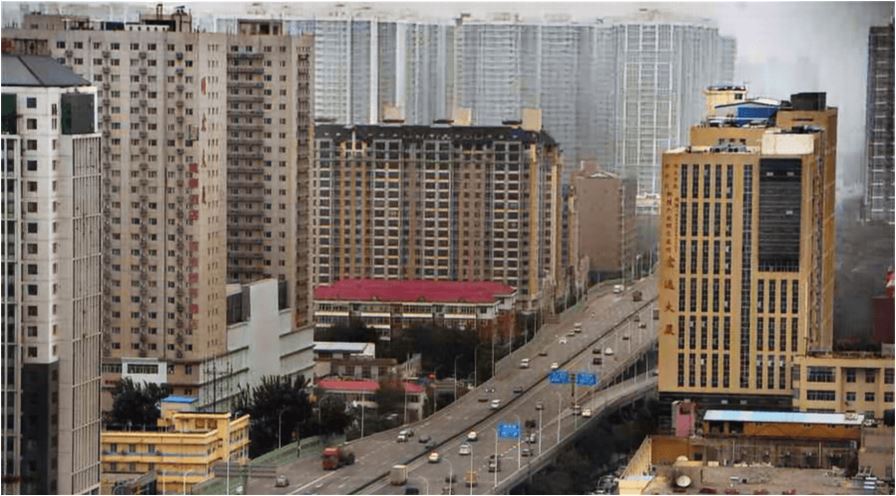} \\
			
			\includegraphics[width=0.12\linewidth]{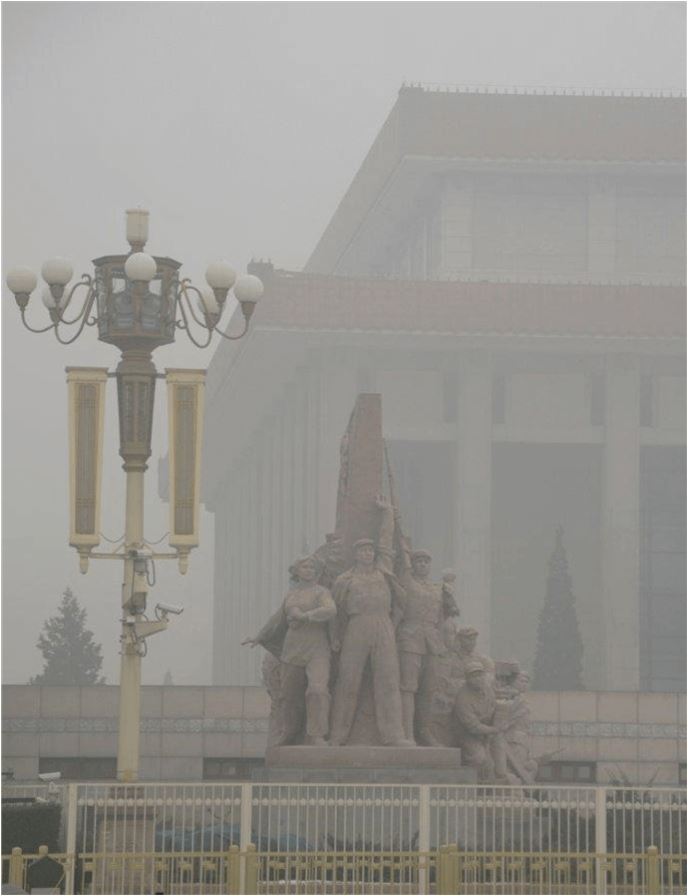} &
			\includegraphics[width=0.12\linewidth]{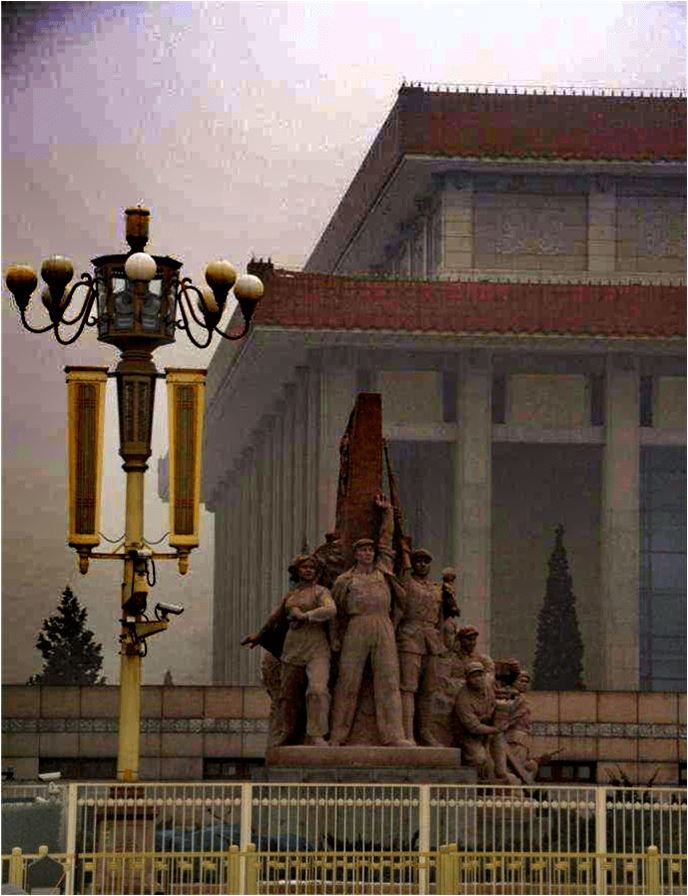} &
			\includegraphics[width=0.12\linewidth]{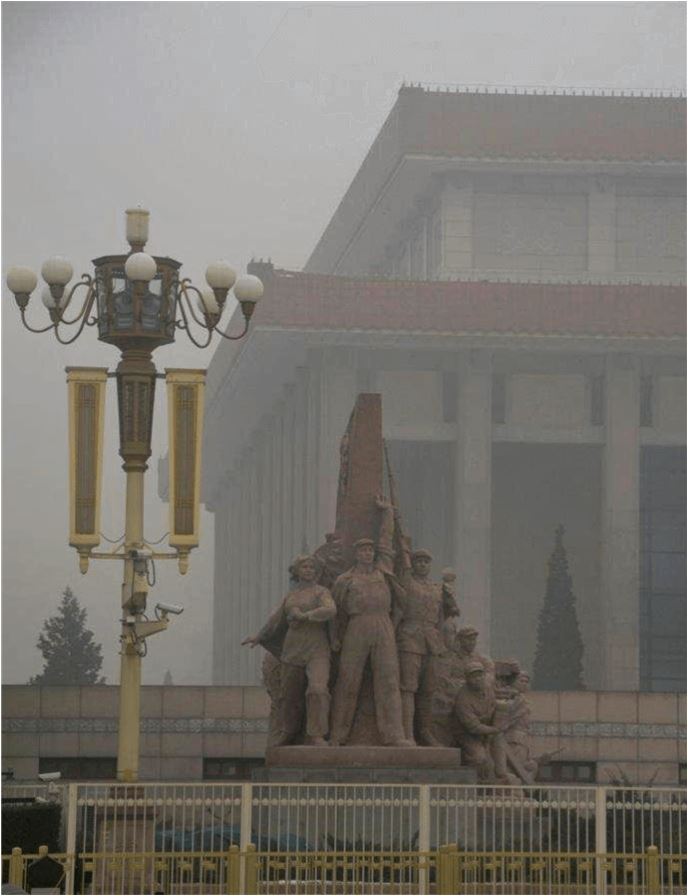} &
			\includegraphics[width=0.12\linewidth]{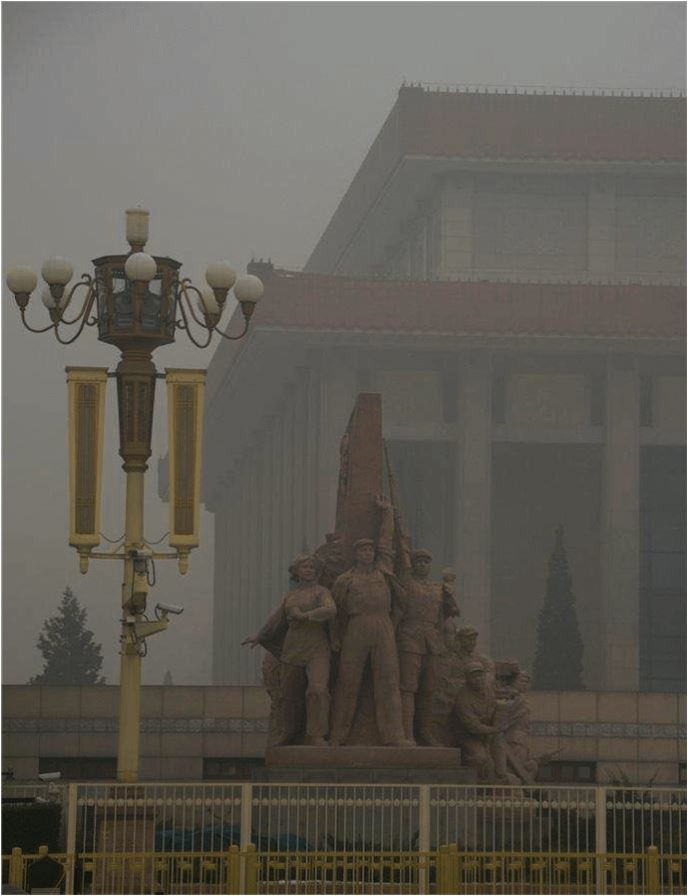} &
			\includegraphics[width=0.12\linewidth]{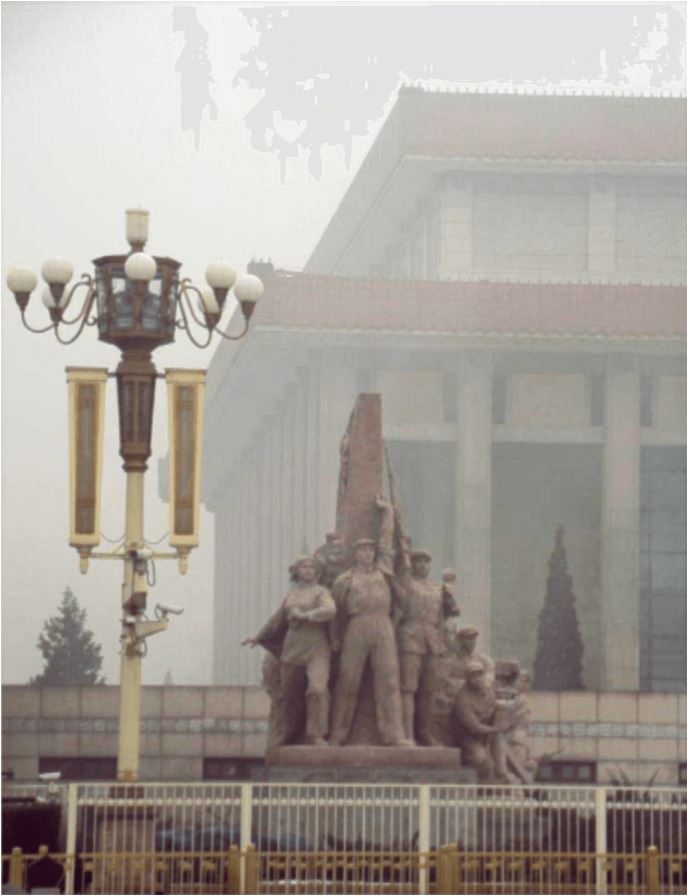} &
			\includegraphics[width=0.12\linewidth]{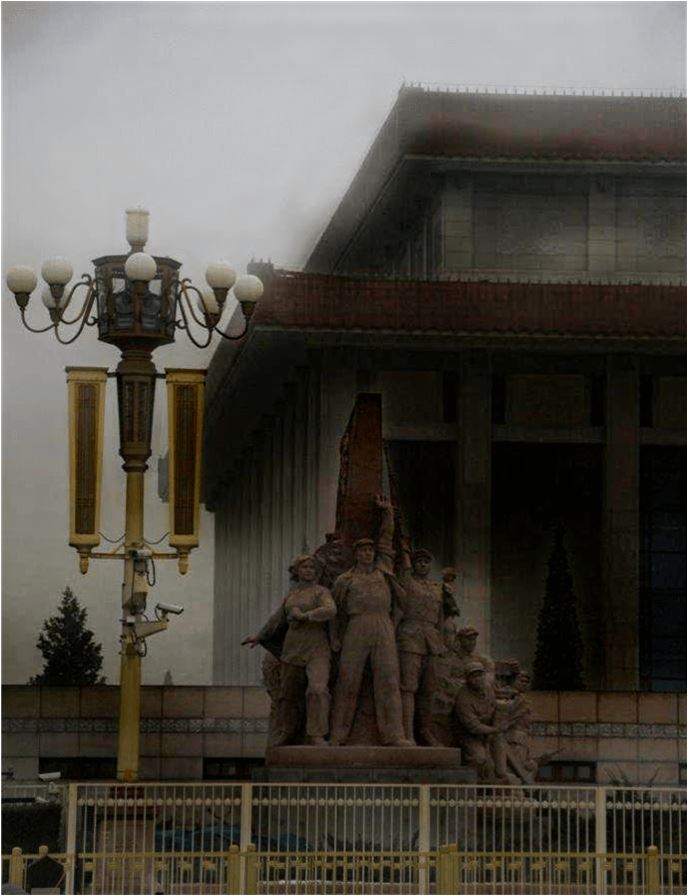} &
			\includegraphics[width=0.12\linewidth]{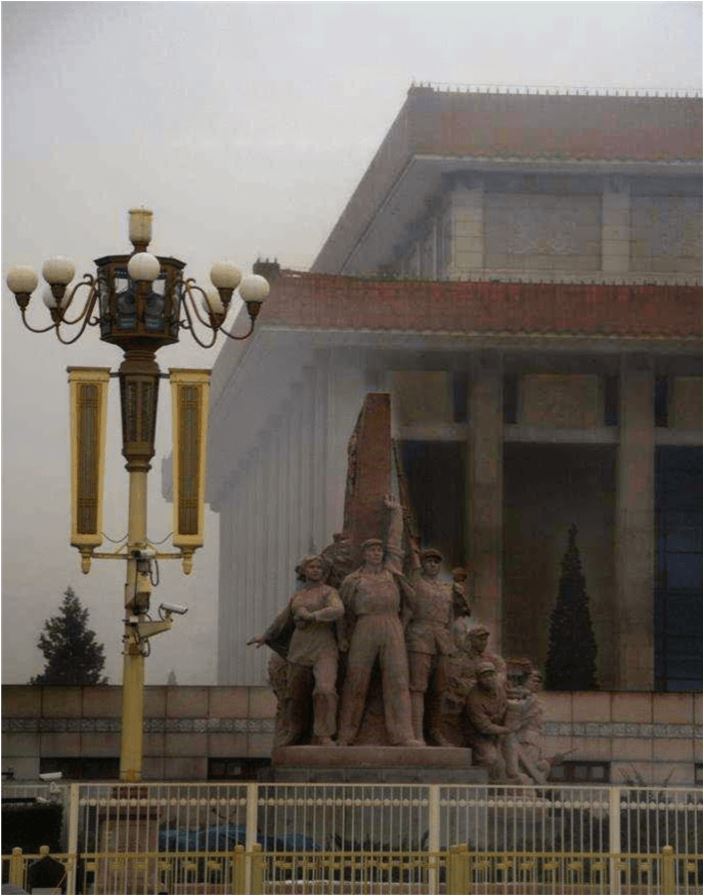} &
			\includegraphics[width=0.12\linewidth]{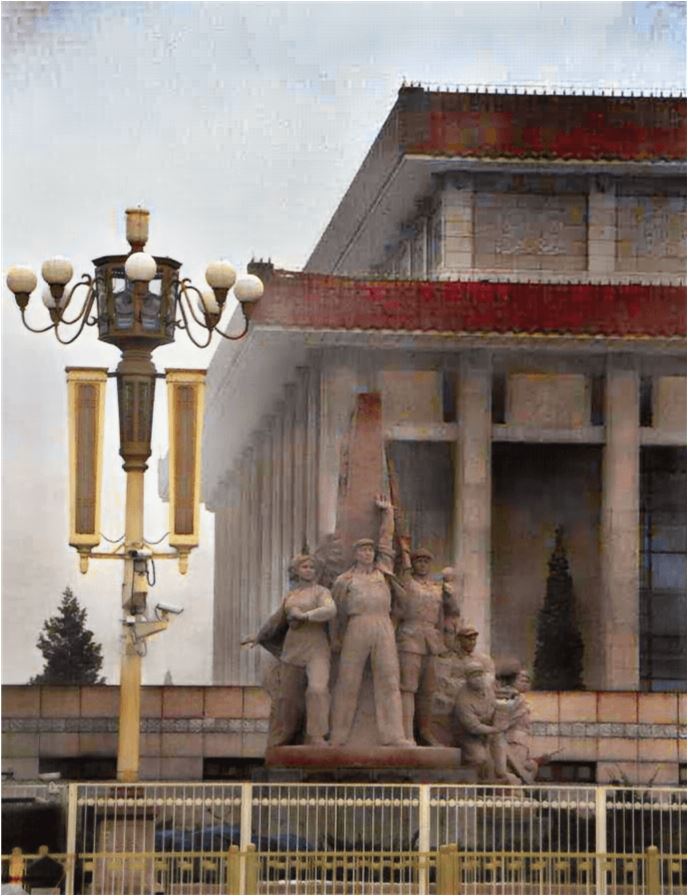} \\
			
			\includegraphics[width=0.12\linewidth]{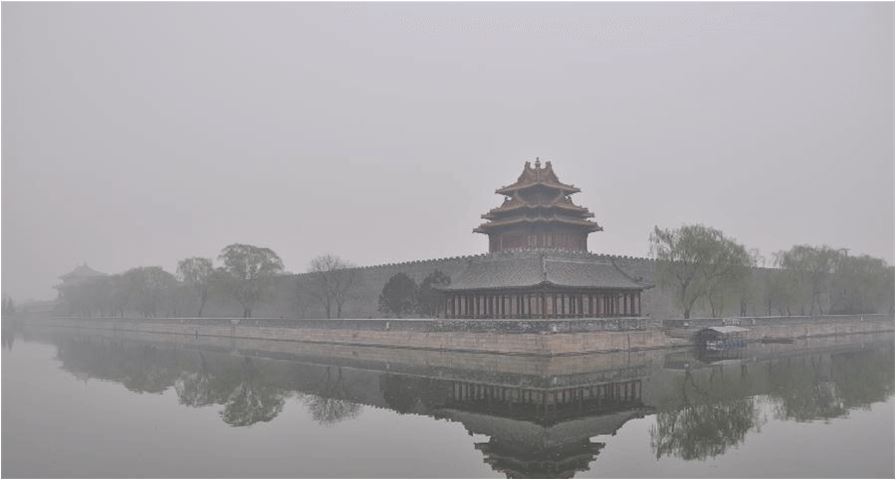} &
			\includegraphics[width=0.12\linewidth]{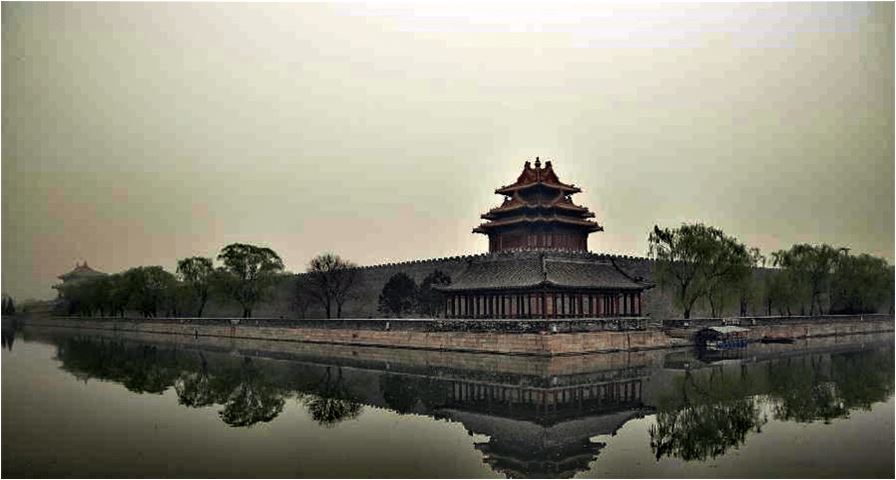} &
			\includegraphics[width=0.12\linewidth]{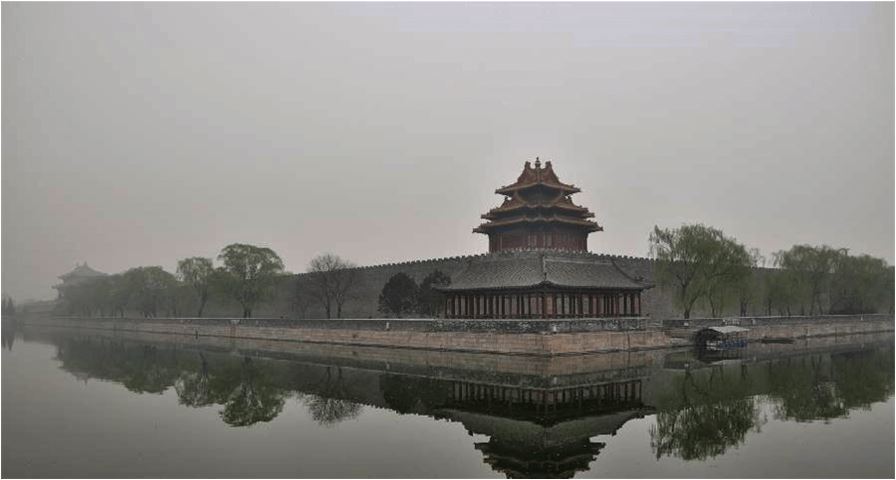} &
			\includegraphics[width=0.12\linewidth]{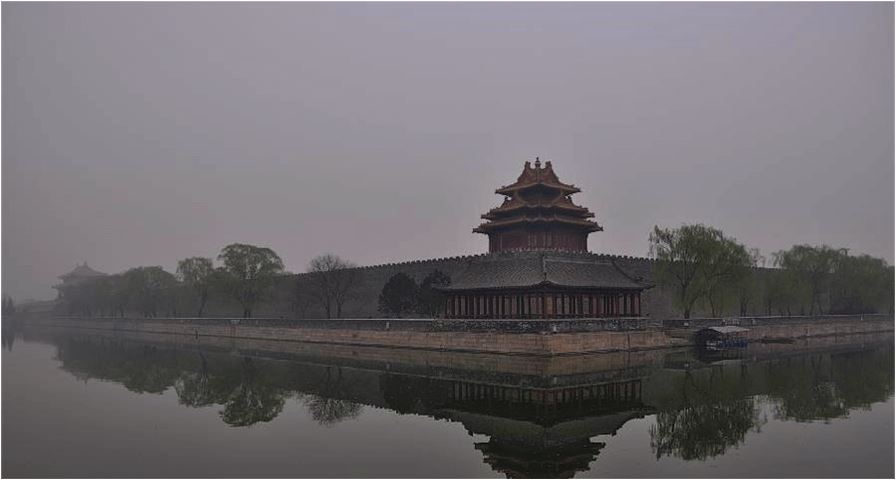} &
			\includegraphics[width=0.12\linewidth]{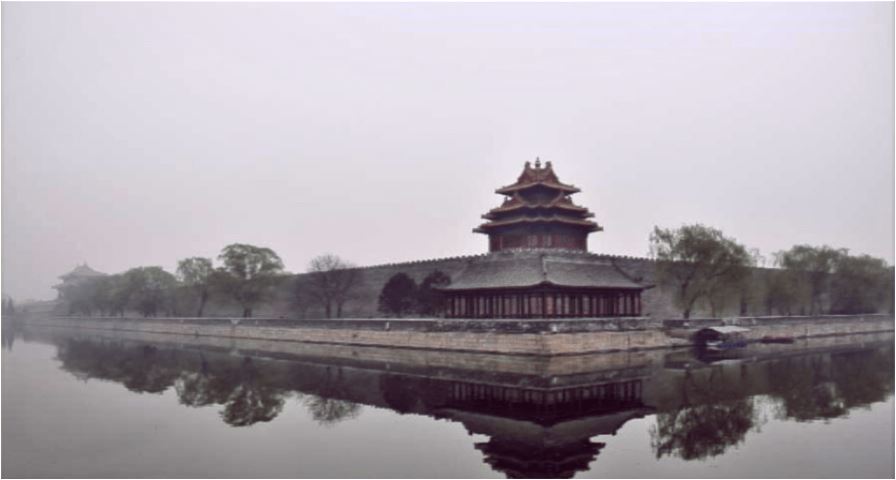} &
			\includegraphics[width=0.12\linewidth]{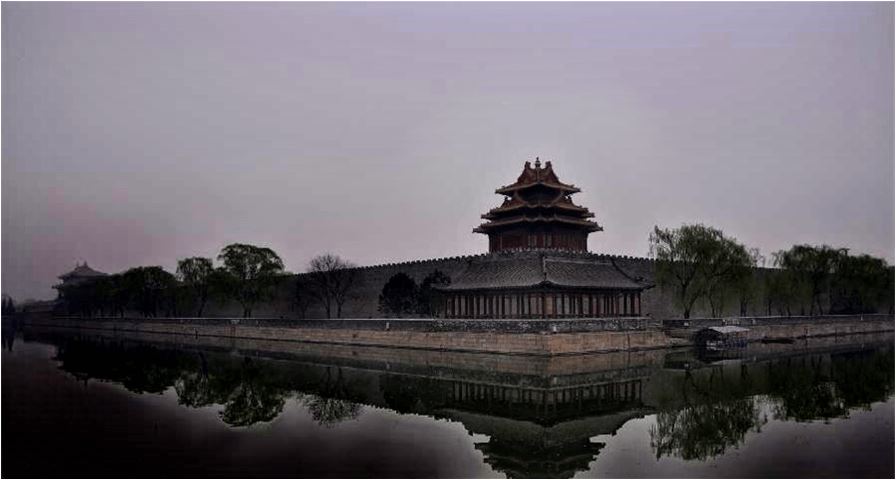} &
			\includegraphics[width=0.12\linewidth]{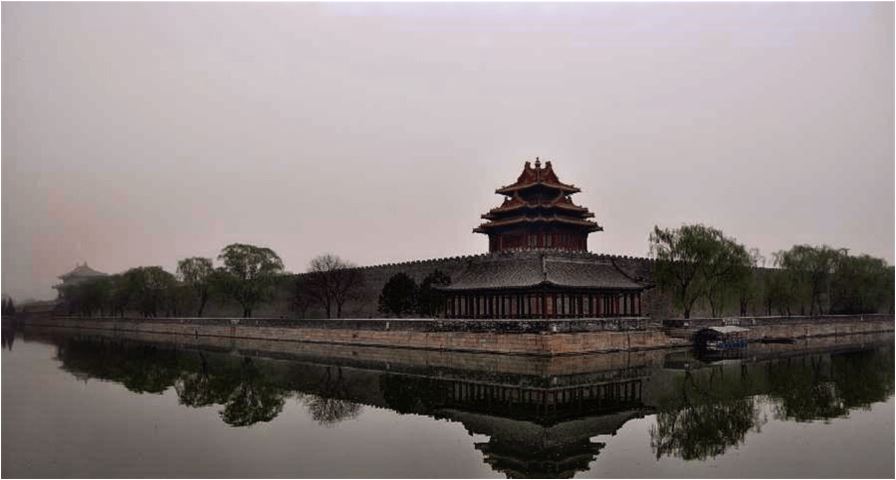} &
			\includegraphics[width=0.12\linewidth]{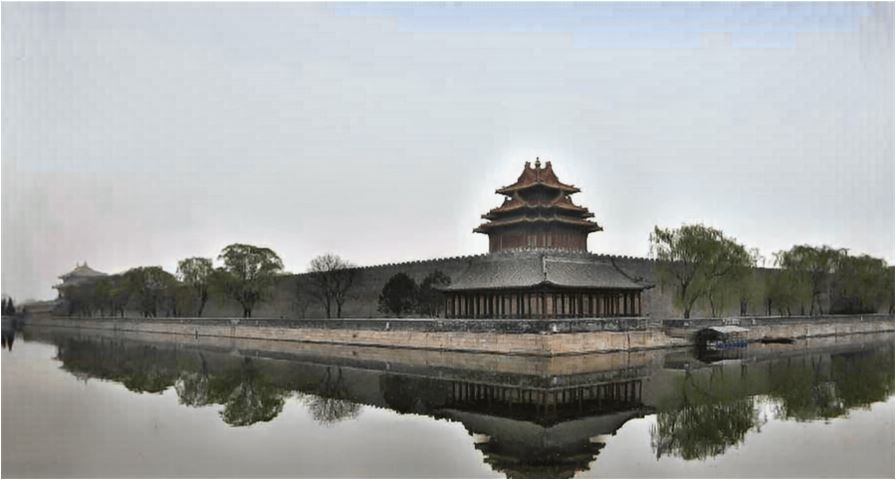} \\
			
			\includegraphics[width=0.12\linewidth]{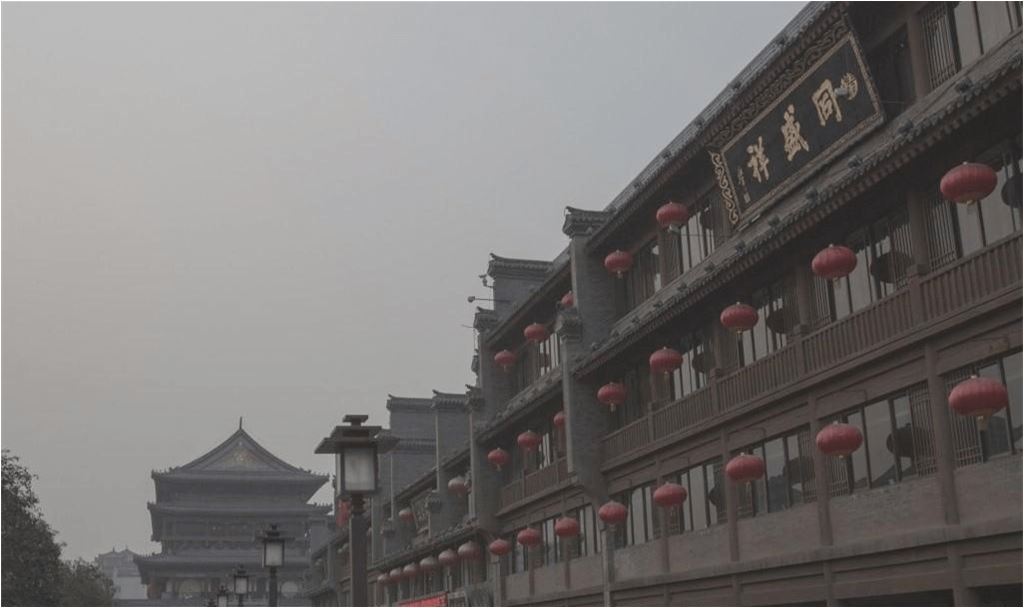} &
			\includegraphics[width=0.12\linewidth]{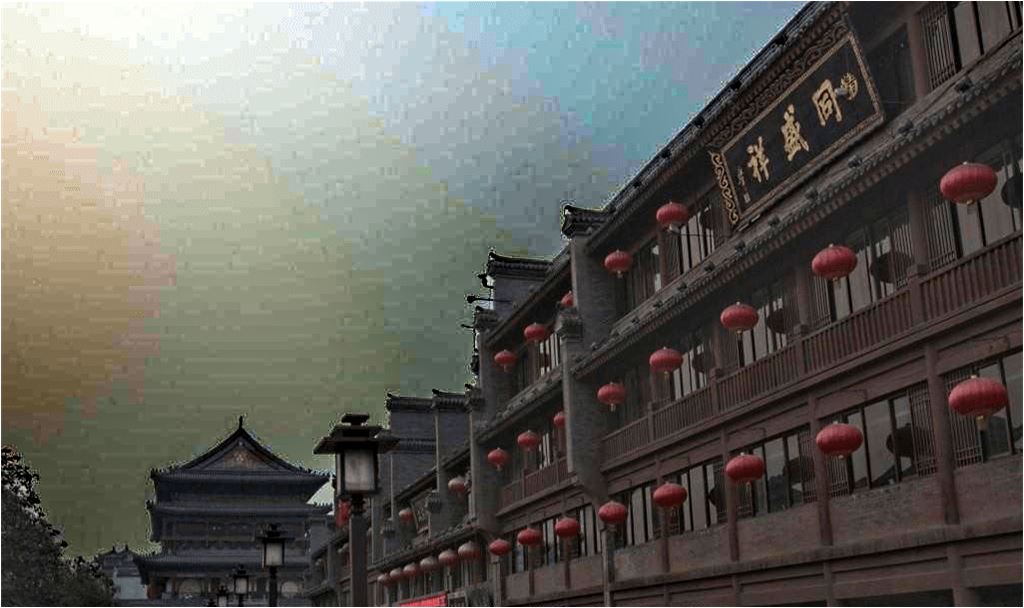} &
			\includegraphics[width=0.12\linewidth]{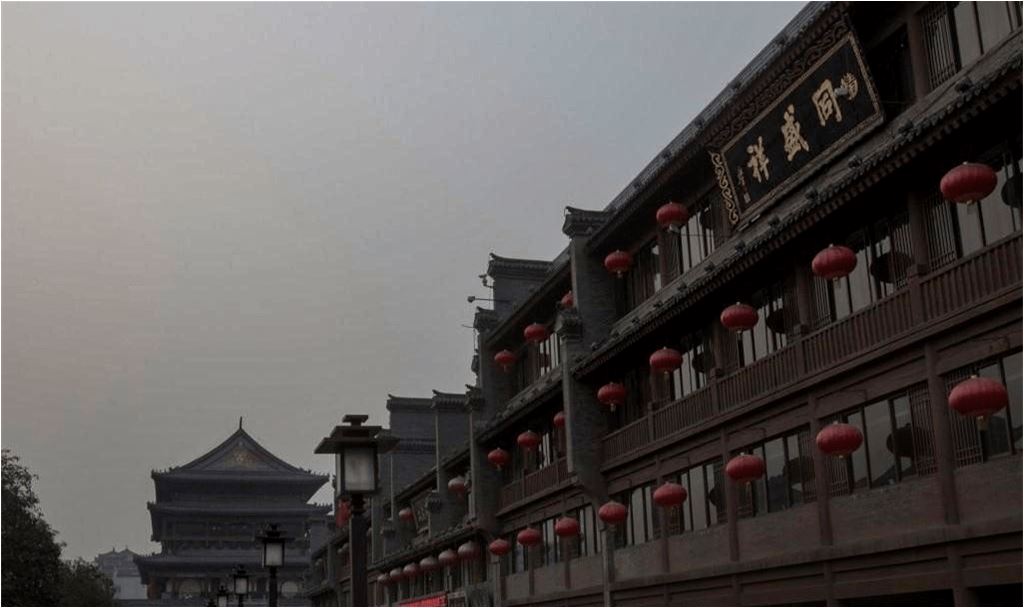} &
			\includegraphics[width=0.12\linewidth]{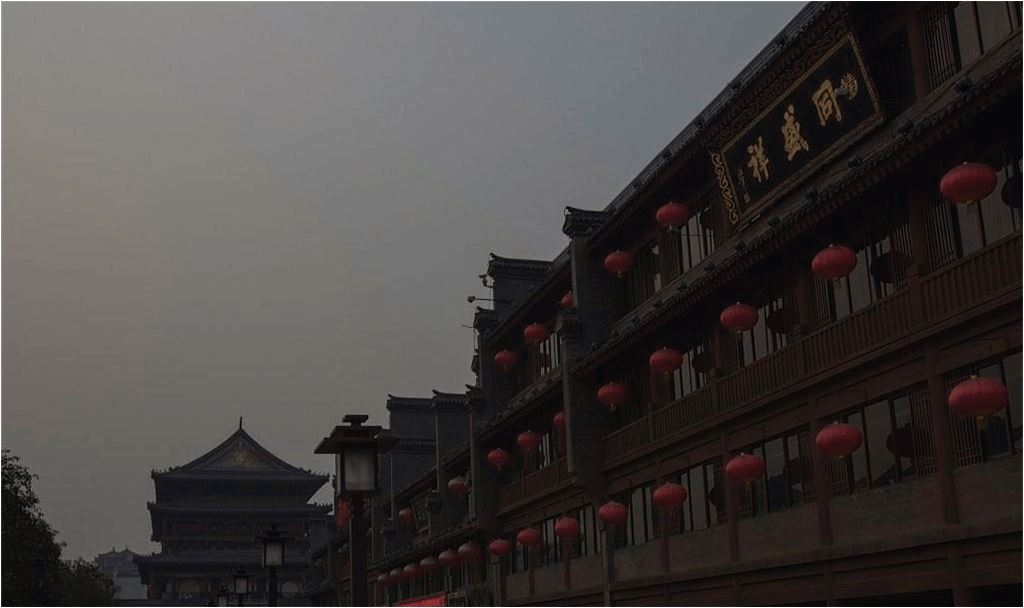} &
			\includegraphics[width=0.12\linewidth]{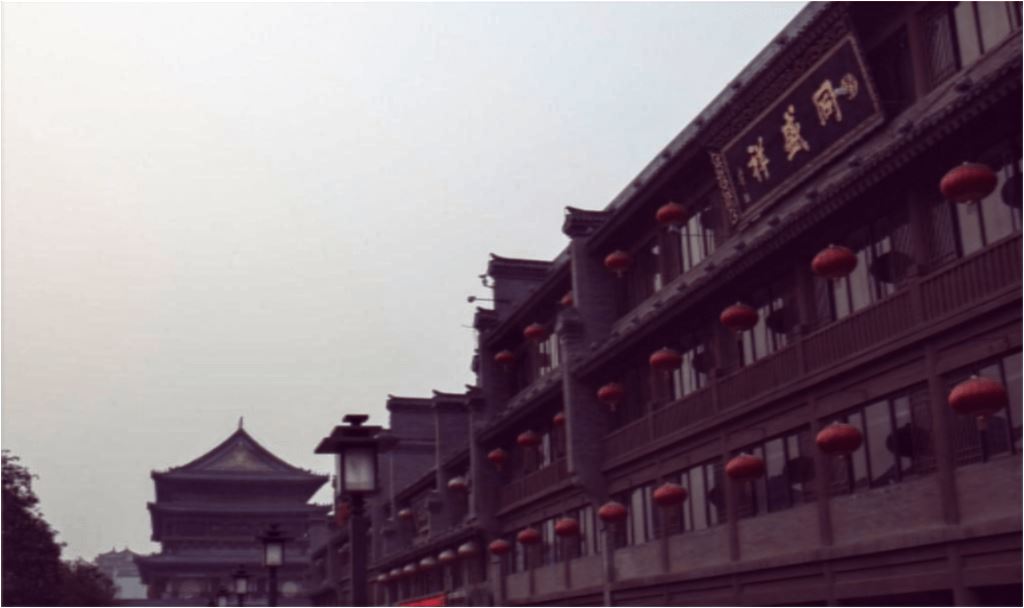} &
			\includegraphics[width=0.12\linewidth]{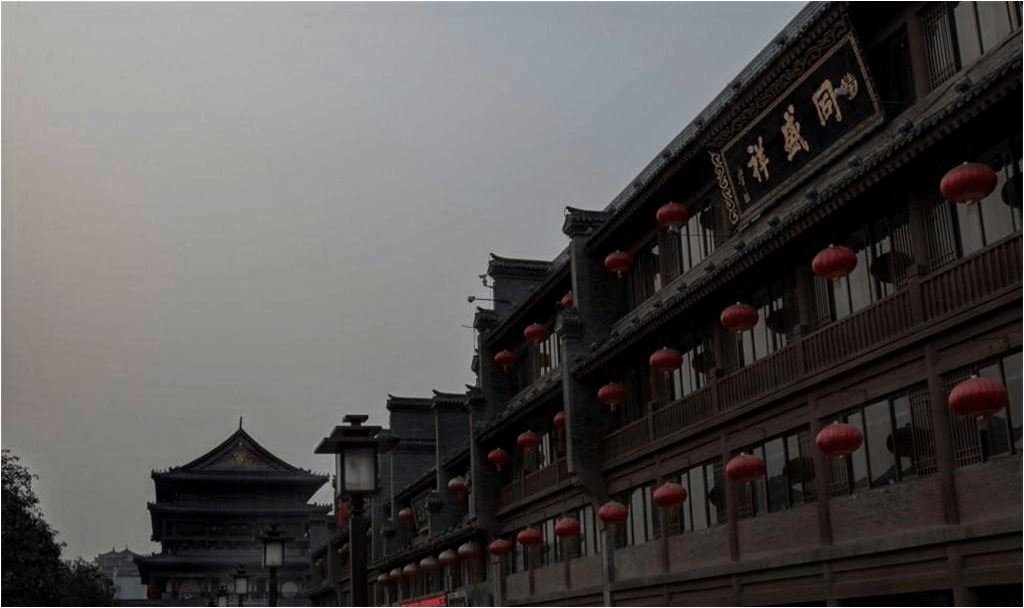} &
			\includegraphics[width=0.12\linewidth]{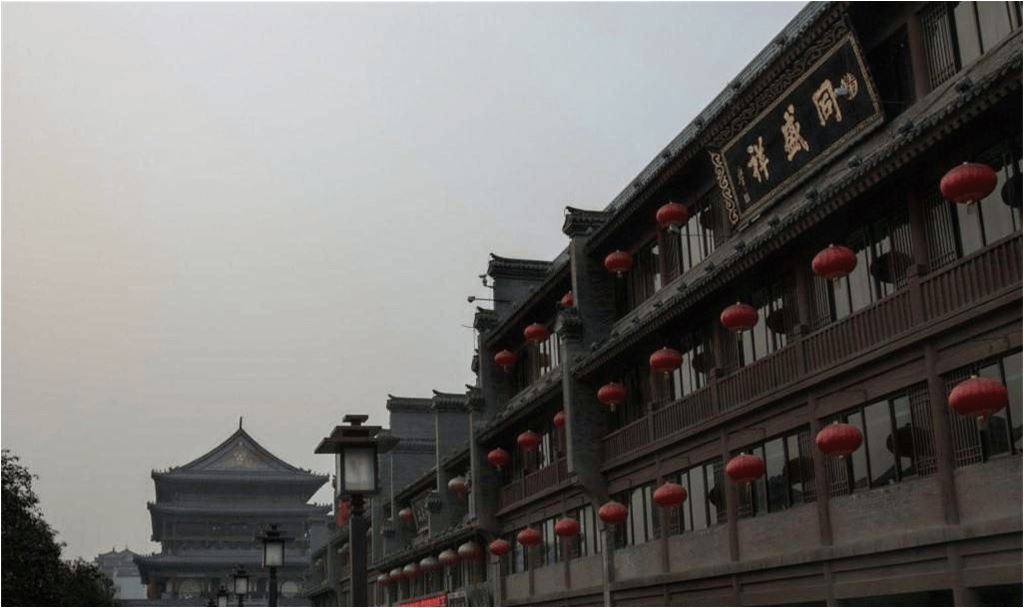} &
			\includegraphics[width=0.12\linewidth]{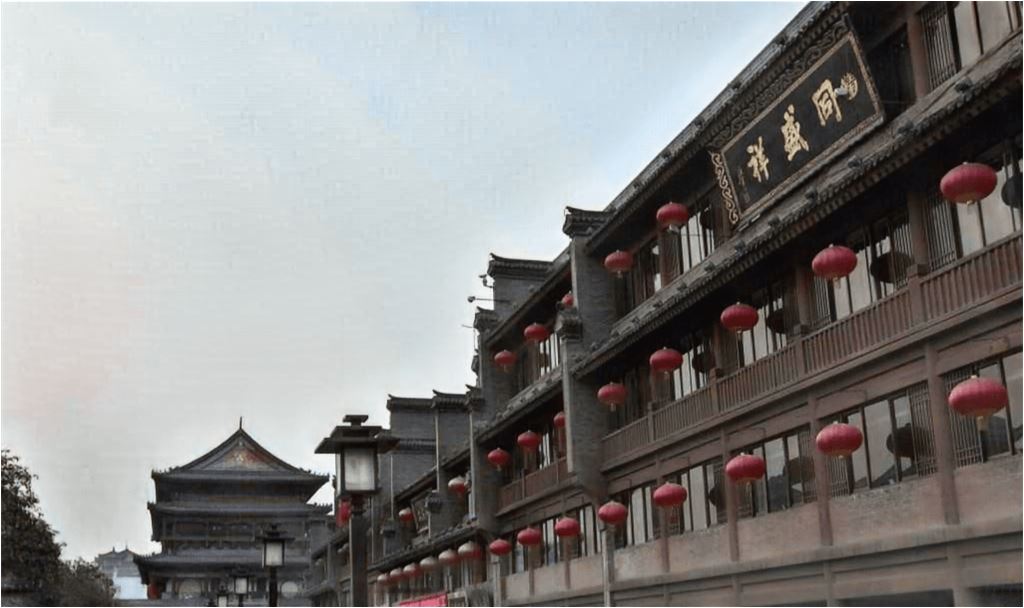} \\
			
			\includegraphics[width=0.12\linewidth]{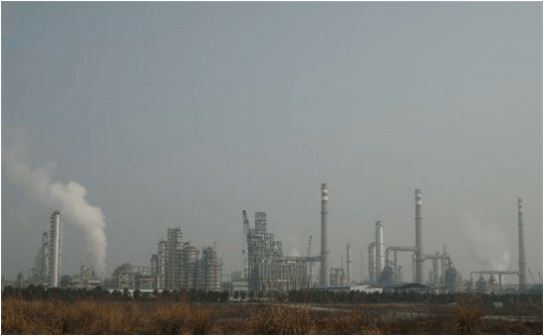} &
			\includegraphics[width=0.12\linewidth]{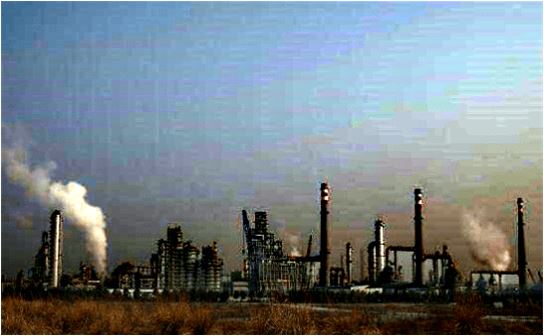} &
			\includegraphics[width=0.12\linewidth]{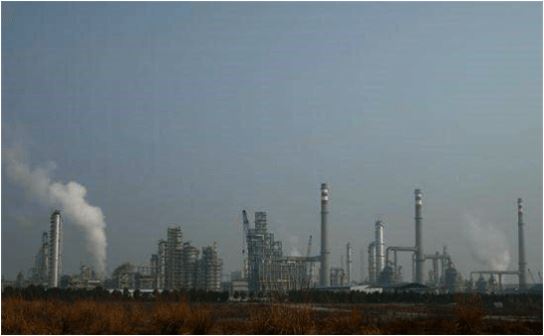} &
			\includegraphics[width=0.12\linewidth]{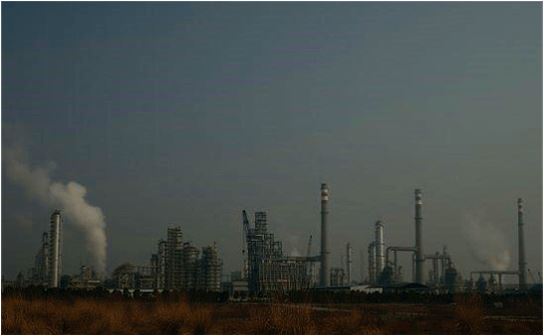} &
			\includegraphics[width=0.12\linewidth]{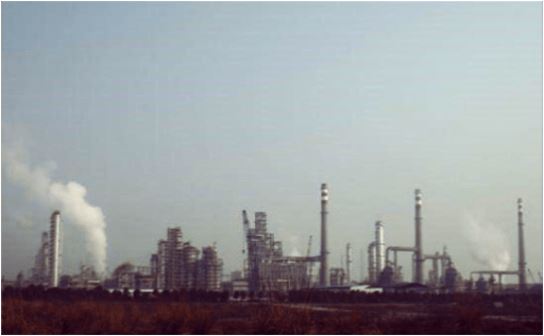} &
			\includegraphics[width=0.12\linewidth]{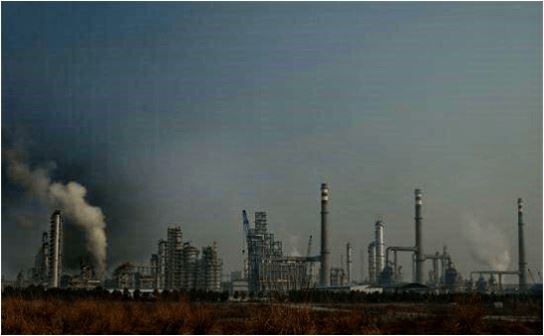} &
			\includegraphics[width=0.12\linewidth]{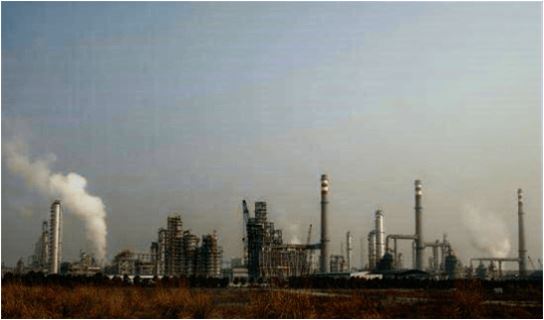} &
			\includegraphics[width=0.12\linewidth]{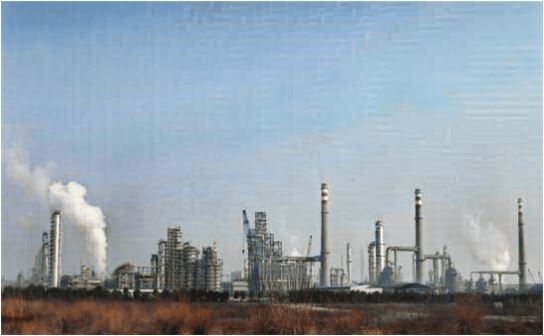} \\
			(a) Hazy image&
			(b) NLD~\cite{berman2016non}&
			(c) DehazeNet~\cite{Cai2016DehazeNet} & 
			(d) AOD-Net~\cite{li2017aod} &
			(e) DCPDN~\cite{Zhang_2018_CVPR} &
			(f) GFN~\cite{Ren_2018_CVPR} &
			(g) EPDN~\cite{qu2019enhanced} &
			(h) Ours  \\
		\end{tabular}
	\end{center}
	\vspace{-2mm}
	\caption{Visual comparisons on the real hazy images.}
	\label{fig:real}
\end{figure*}
\vspace{-4mm}
\paragraph{Comparison methods.}
We evaluate the proposed method against the following state-of-the-art approaches: 
DCP~\cite{He2011Single}, MSCNN~\cite{ren2016single}, DehazeNet~\cite{Cai2016DehazeNet}, NLD~\cite{berman2016non}, AOD-Net~\cite{li2017aod}, GFN~\cite{Ren_2018_CVPR}, DCPDN~\cite{Zhang_2018_CVPR}, and EPDN~\cite{qu2019enhanced}. 
%
%
%
More image dehazing results and comparisons against other dehazing approaches are included in the supplementary materials.

%

%
\begin{figure*}[htbp]
	\scriptsize
	\centering
	\renewcommand{\tabcolsep}{1pt} 
	\renewcommand{\arraystretch}{1} 
	\begin{center}
		\begin{tabular}{ccccc}
			\includegraphics[width=0.17\linewidth]{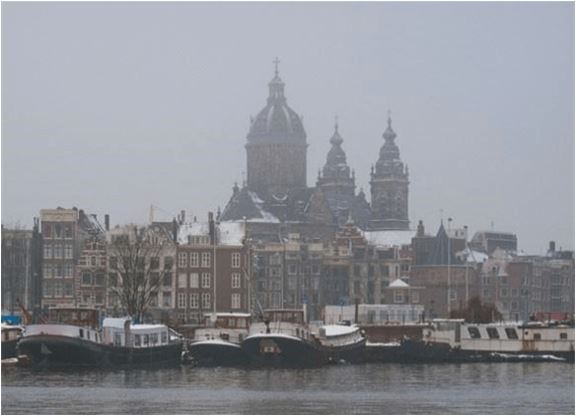} &
			\includegraphics[width=0.17\linewidth]{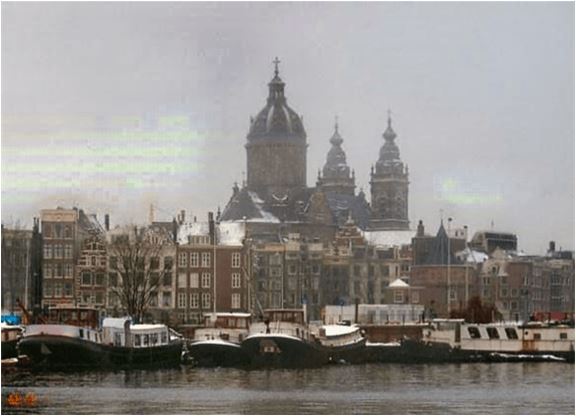} &
			\includegraphics[width=0.17\linewidth]{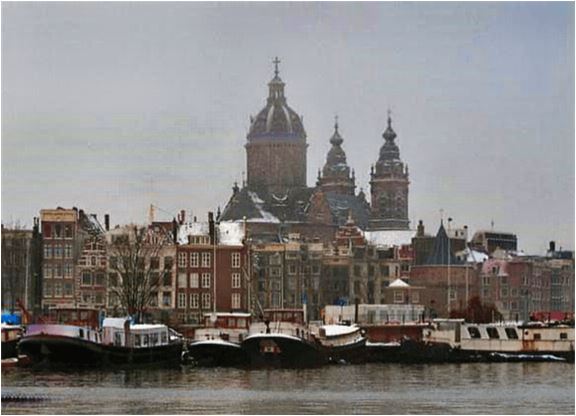} &
			\includegraphics[width=0.17\linewidth]{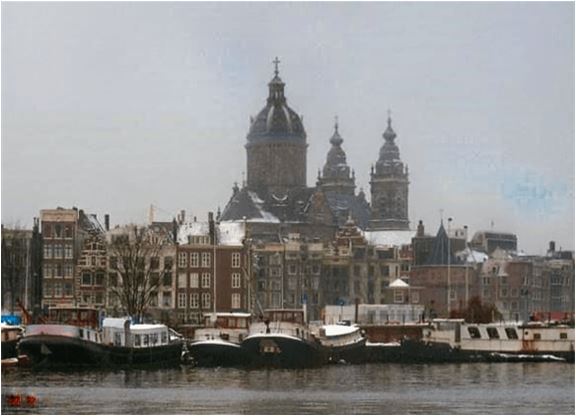} &
			\includegraphics[width=0.17\linewidth]{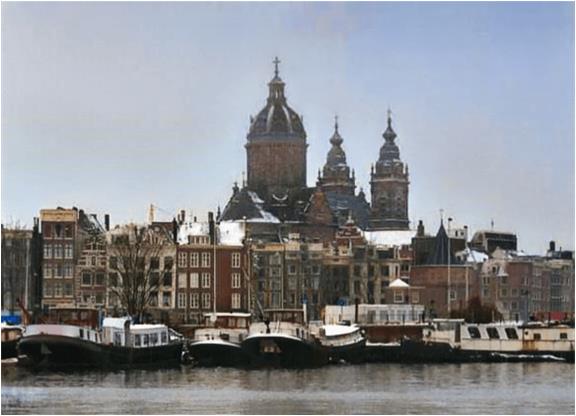} \\
			
			\includegraphics[width=0.17\linewidth]{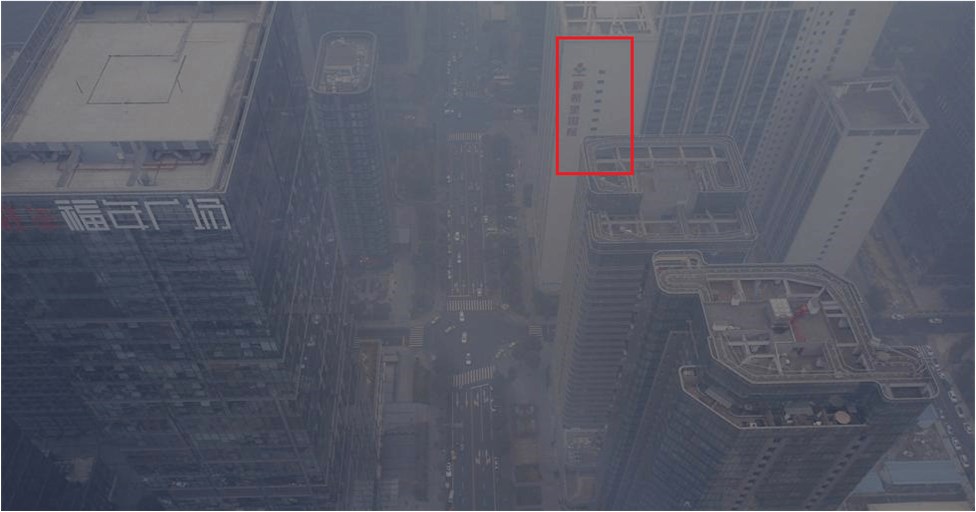} &
			\includegraphics[width=0.17\linewidth]{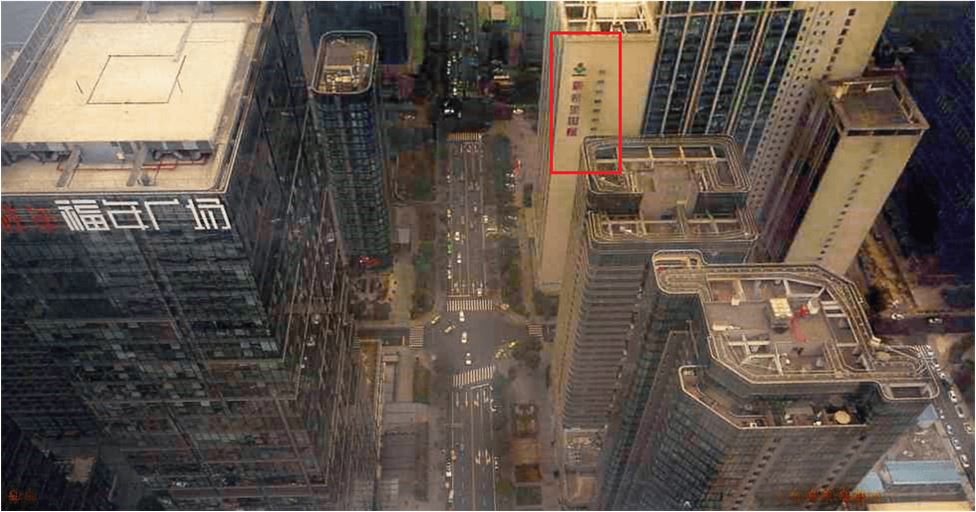} &
			\includegraphics[width=0.17\linewidth]{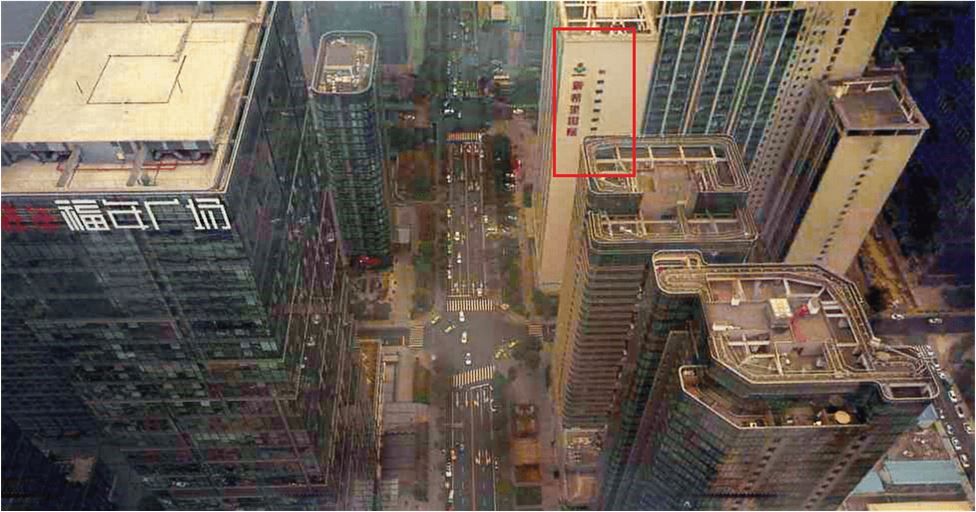} &
			\includegraphics[width=0.17\linewidth]{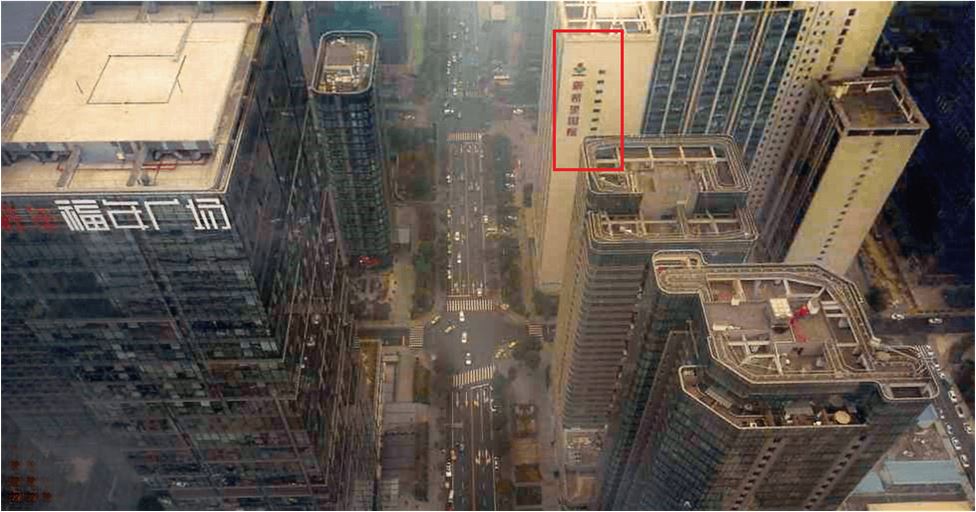} &
			\includegraphics[width=0.17\linewidth]{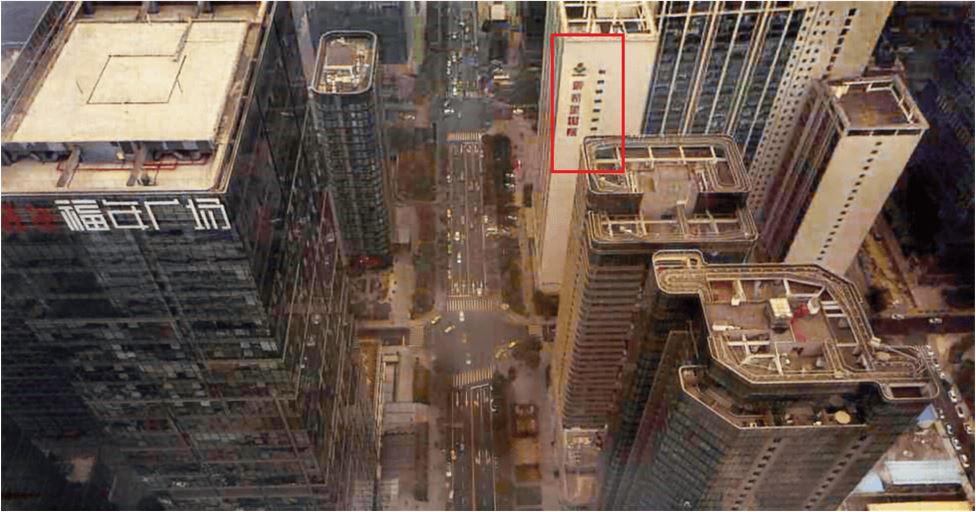} \\
			
			(a) Real hazy image &
			(b) Dehazed result of SYN &
			(c) Dehazed result of S2R   &
			(d) Dehazed result of SYN+U &
			(e) Dehazed result of Ours \\
		\end{tabular}
	\end{center}
	\vspace{-2mm}
	\caption{Comparison of dehazed results of several dehazing models on real hazy images.}
	\label{fig:abl_real}
\end{figure*}
\subsection{Experiments on Synthetic Datasets}
We use two synthetic datasets, namely SOTS~\cite{li2019benchmarking} and HazeRD~\cite{Zhang:HazeRD:ICIP17b}, to evaluate the performance of our proposed method. 
%
%
%
The dehazed images of different methods on these two datasets are shown in Figure~\ref{fig:SOTS} and~\ref{fig:HazeRD}.
From Figure~\ref{fig:SOTS} (b), we can observe that both NLD~\cite{berman2016non} and GFN~\cite{Ren_2018_CVPR} suffer from some color distortion where the result looks unrealistic.
%
%
The dehazing results by EPDN~\cite{qu2019enhanced} are also darker than the ground-truths in some cases, as shown in Figure~\ref{fig:SOTS} (g).
Besides, there is still some remaining haze in the dehazed images by DehazeNet~\cite{Cai2016DehazeNet}, AOD-Net~\cite{li2017aod} and DCPDN~\cite{Zhang_2018_CVPR}.
Compared with these methods, our algorithm restores images with sharper structures and details, which are closer to the ground-truths.
Similar results can be found in the dehazed results on the HazeRD dataset in Figure~\ref{fig:HazeRD}, our algorithm generates results with better visual effect.
%

%
We also give the quantitative comparison of dehazed results in Table~\ref{tab:Syn}. As shown, the proposed method obtains the highest PSNR and SSIM values on both datasets.
Compared with the state-of-the-art EPDN~\cite{qu2019enhanced}, our approach achieves the gain with 3.94 dB and 0.04 in terms of PSNR and SSIM on the STOS dataset, respectively.
For the HazeRD dataset, the PSNR and SSIM produced by our method are higher than EPDN~\cite{qu2019enhanced} by up to 0.7dB and 0.07, respectively.

\subsection{Experiments on Real Images}
To evaluate the generalization of our method on real images, we compare the visual results of different approaches on real hazy images from the URHI dataset.
As shown in Figure~\ref{fig:real}, NLD~\cite{berman2016non} suffers from serious color distortions (see, e.g., the sky in Figure~\ref{fig:real} (b)). 
From Fig~\ref{fig:real} (f), we can figure out that the GFN~\cite{Ren_2018_CVPR} also suffers from color distortion in some cases, and the dehazed results look darker than our method. 
In addition, the dehazed results by DehazeNet~\cite{Cai2016DehazeNet}, AOD-Net~\cite{li2017aod}, and DCPDN~\cite{Zhang_2018_CVPR} have some remaining haze artifacts as shown in the fifth row in Figure~\ref{fig:real} (c-e)). 
Though EPDN~\cite{qu2019enhanced} has achieved a better visual effect than the above methods, the brightness of the dehazing results is lower than our method in general. 
%
Overall, our proposed method restores more details and obtains visually pleasing images.
%


\subsection{Ablation Study}
To verify the effectiveness of the image translation network and the unsupervised loss, we conduct a series of ablations to analyze our method.
We construct the following dehazing models for comparison: 1) \textbf{SYN}: ${{\cal G}_S}$ is only trained on ${X_S}$; 2) \textbf{SYN+U}: ${{\cal G}_S}$ is trained on both ${X_S}$ and ${X_R}$; 3) \textbf{R2S+U}: ${{\cal G}_S}$ is only trained on ${X_S}$ and ${G_{R \to S}}({X_R})$; 4) \textbf{S2R}: ${{\cal G}_R}$ is trained on ${G_{S \to R}}({X_{\rm{s}}},{D_{\rm{s}}})$.
%

We compare the proposed domain adaptation method against these four dehazing models on both synthetic and real hazy images. 
The visual and quantitative results are shown in Table~\ref{tab:abl_syn} and Figure~\ref{fig:abl_real}, which demonstrate that our approach achieves the best performance of image dehazing in terms of PSNR and SSIM as well as visual effects.
As shown in Figure~\ref{fig:abl_real} (b), due to the domain shift, the \textbf{SYN} method causes color distortion or darker artifacts (see, e.g., the sky part and the red rectangle). 
In contrast, 
the dehazing model ${{\cal G}_R}$ trained on the translated images (\textbf{S2R}) achieves better quality image as shown in Figure~\ref{fig:abl_real} (c), which demonstrates that the translators effectively reduces the discrepancy between synthetic data and real images.
Moreover, Figure~\ref{fig:abl_real} (b) and (d) show that the dehazing model with unsupervised loss (\textbf{SYN+U}) can produce better results than \textbf{SYN}, which demonstrate the effectiveness of the unsupervised loss.
Finally, we can observe that the proposed method with both translators and unsupervised loss generates cleaner and visually more pleasing results (e.g., the sky is brighter) in Figure~\ref{fig:abl_real} (e).
%
The quantitative results in Table~\ref{tab:abl_syn} by applying image translation and unsupervised loss also agree with the qualitative results in Figure~\ref{fig:abl_real}. 

As a conclusion, these ablations demonstrate that the image translation model and unsupervised loss are useful to reduce the domain gap between synthetic data and real-world images and improve the performance of image dehazing on both synthetic and real domains.

\begin{table}[t]
\footnotesize
\centering
\caption{Quantitative results of different dehazing models on synthetic domain.}
\vspace{1mm}
\label{tab:abl_syn}
\begin{tabular}{ccc}
\toprule
Method		             &  PSNR      & SSIM 	    \\ \midrule
SYN			 &	25.67	  & 0.8801	    \\
SYN+U			 &  25.75	  & 0.8699	    \\
R2S+U	     &  25.91     & 0.8822       \\
Ours	     &  \textbf{27.76}     & \textbf{0.9284} \\\bottomrule
\end{tabular}
\end{table}

\section{Conclusions}
In this work, we propose a novel domain adaptation framework for single image dehazing, which contains an image translation module and two image dehazing modules.
we first use the image translation network to translate images from one domain to another to reduce the domain discrepancy. 
And then, the image dehazing networks take the translated images and their original images as inputs to perform image dehazing.
To further improve the generalization, we incorporate the real hazy image into the dehazing training by exploiting the properties of clean images. 
Extensive experimental results on both synthetic datasets and real-world images demonstrate that our algorithm performs favorably against state-of-the-arts.
\vspace{-4mm}
\paragraph{Acknowledgements.} 
This work is supported by the Project of the National Natural Science Foundation of China No.61433007 and No.61901184. W. Ren is supported in part by the Zhejiang Lab’s International Talent Fund for Young Professionals and the CCF-Tencent Open Fund.

{\small
\bibliographystyle{ieee_fullname}
\bibliography{reference}

\begin{thebibliography}{10}\itemsep=-1pt

\bibitem{atapour2018real}
Amir Atapour-Abarghouei and Toby~P Breckon.
\newblock Real-time monocular depth estimation using synthetic data with domain
  adaptation via image style transfer.
\newblock In {\em IEEE Conference on Computer Vision and Pattern Recognition},
  2018.

\bibitem{berman2016non}
Dana Berman, Shai Avidan, et~al.
\newblock Non-local image dehazing.
\newblock In {\em IEEE Conference on Computer Vision and Pattern Recognition},
  2016.

\bibitem{bousmalis2017unsupervised}
Konstantinos Bousmalis, Nathan Silberman, David Dohan, Dumitru Erhan, and Dilip
  Krishnan.
\newblock Unsupervised pixel-level domain adaptation with generative
  adversarial networks.
\newblock In {\em IEEE Conference on Computer Vision and Pattern Recognition},
  2017.

\bibitem{Cai2016DehazeNet}
Bolun Cai, Xiangmin Xu, Kui Jia, Qing Chunmei, and Dacheng Tao.
\newblock Dehazenet: An end-to-end system for single image haze removal.
\newblock {\em IEEE Transactions on Image Processing}, 2016.

\bibitem{chen2019learning}
Yuhua Chen, Wen Li, Xiaoran Chen, and Luc~Van Gool.
\newblock Learning semantic segmentation from synthetic data: A geometrically
  guided input-output adaptation approach.
\newblock In {\em IEEE Conference on Computer Vision and Pattern Recognition},
  2019.

\bibitem{chen2018domain}
Yuhua Chen, Wen Li, Christos Sakaridis, Dengxin Dai, and Luc Van~Gool.
\newblock Domain adaptive faster r-cnn for object detection in the wild.
\newblock In {\em IEEE Conference on Computer Vision and Pattern Recognition},
  2018.

\bibitem{dundar2018domain}
Aysegul Dundar, Ming-Yu Liu, Ting-Chun Wang, John Zedlewski, and Jan Kautz.
\newblock Domain stylization: A strong, simple baseline for synthetic to real
  image domain adaptation.
\newblock {\em arXiv preprint arXiv:1807.09384}, 2018.

\bibitem{fattal2014dehazing}
Raanan Fattal.
\newblock Dehazing using color-lines.
\newblock {\em ACM Transactions on Graphics}, 2014.

\bibitem{He2011Single}
Kaiming He, Jian Sun, and Xiaoou Tang.
\newblock Single image haze removal using dark channel prior.
\newblock {\em IEEE Transactions on Pattern Analysis and Machine Intelligence},
  2011.

\bibitem{hoffman2017cycada}
Judy Hoffman, Eric Tzeng, Taesung Park, Jun-Yan Zhu, Phillip Isola, Kate
  Saenko, Alexei~A Efros, and Trevor Darrell.
\newblock Cycada: Cycle-consistent adversarial domain adaptation.
\newblock In {\em International Conference on Machine Learning}, 2018.

\bibitem{kingma2014adam}
Diederik~P Kingma and Jimmy Ba.
\newblock Adam: A method for stochastic optimization.
\newblock {\em arXiv}, 2014.

\bibitem{li2017aod}
Boyi Li, Xiulian Peng, Zhangyang Wang, Jizheng Xu, and Dan Feng.
\newblock Aod-net: All-in-one dehazing network.
\newblock In {\em IEEE International Conference on Computer Vision}, 2017.

\bibitem{li2019benchmarking}
Boyi Li, Wenqi Ren, Dengpan Fu, Dacheng Tao, Dan Feng, Wenjun Zeng, and
  Zhangyang Wang.
\newblock Benchmarking single-image dehazing and beyond.
\newblock {\em IEEE Transactions on Image Processing}, 2019.

\bibitem{li2019semi}
Lerenhan Li, Yunlong Dong, Wenqi Ren, Jinshan Pan, Changxin Gao, Nong Sang, and
  Ming-Hsuan Yang.
\newblock Semi-supervised image dehazing.
\newblock {\em IEEE Transactions on Image Processing}, 29:2766--2779, 2019.

\bibitem{li2020dyna}
Lerenhan Li, Jinshan Pan, Wei-Sheng Lai, Changxin Gao, Nong Sang, and
  Ming-Hsuan Yang.
\newblock Dynamic scene deblurring by depth guided model.
\newblock {\em IEEE Transactions on Image Processing}, 2020.

\bibitem{Li_2018_CVPR}
Runde Li, Jinshan Pan, Zechao Li, and Jinhui Tang.
\newblock Single image dehazing via conditional generative adversarial network.
\newblock In {\em IEEE Conference on Computer Vision and Pattern Recognition},
  2018.

\bibitem{li2015nighttime}
Yu Li, Robby~T Tan, and Michael~S Brown.
\newblock Nighttime haze removal with glow and multiple light colors.
\newblock In {\em IEEE International Conference on Computer Vision}, 2015.

\bibitem{li2017haze}
Yu Li, Shaodi You, Michael~S Brown, and Robby~T Tan.
\newblock Haze visibility enhancement: A survey and quantitative benchmarking.
\newblock {\em Computer Vision and Image Understanding}, 165:1--16, 2017.

\bibitem{long2015learning}
Mingsheng Long, Yue Cao, Jianmin Wang, and Michael~I Jordan.
\newblock Learning transferable features with deep adaptation networks.
\newblock In {\em International Conference on Machine Learning}, 2015.

\bibitem{long2013transfer}
Mingsheng Long, Guiguang Ding, Jianmin Wang, Jiaguang Sun, Yuchen Guo, and
  Philip~S Yu.
\newblock Transfer sparse coding for robust image representation.
\newblock In {\em IEEE Conference on Computer Vision and Pattern Recognition},
  2013.

\bibitem{mccartney1976optics}
Earl~J McCartney.
\newblock Optics of the atmosphere: scattering by molecules and particles.
\newblock {\em New York, John Wiley and Sons, Inc., 1976. 421 p.}, 1976.

\bibitem{meng2013efficient}
Gaofeng Meng, Ying Wang, Jiangyong Duan, Shiming Xiang, and Chunhong Pan.
\newblock Efficient image dehazing with boundary constraint and contextual
  regularization.
\newblock In {\em IEEE International Conference on Computer Vision}, 2013.

\bibitem{narasimhan2002vision}
Srinivasa~G Narasimhan and Shree~K Nayar.
\newblock Vision and the atmosphere.
\newblock {\em International Journal of Computer Vision}, 48(3):233--254, 2002.

\bibitem{nishino2012bayesian}
Ko Nishino, Louis Kratz, and Stephen Lombardi.
\newblock Bayesian defogging.
\newblock {\em International Journal of Computer Vision}, 2012.

\bibitem{qu2019enhanced}
Yanyun Qu, Yizi Chen, Jingying Huang, and Yuan Xie.
\newblock Enhanced pix2pix dehazing network.
\newblock In {\em IEEE Conference on Computer Vision and Pattern Recognition},
  2019.

\bibitem{ren2016single}
Wenqi Ren, Si Liu, Hua Zhang, Jinshan Pan, Xiaochun Cao, and Ming-Hsuan Yang.
\newblock Single image dehazing via multi-scale convolutional neural networks.
\newblock In {\em European Conference on Computer Vision}, 2016.

\bibitem{Ren_2018_CVPR}
Wenqi Ren, Lin Ma, Jiawei Zhang, Jinshan Pan, Xiaochun Cao, Wei Liu, and
  Ming-Hsuan Yang.
\newblock Gated fusion network for single image dehazing.
\newblock In {\em IEEE Conference on Computer Vision and Pattern Recognition},
  2018.

\bibitem{shrivastava2017learning}
Ashish Shrivastava, Tomas Pfister, Oncel Tuzel, Joshua Susskind, Wenda Wang,
  and Russell Webb.
\newblock Learning from simulated and unsupervised images through adversarial
  training.
\newblock In {\em IEEE Conference on Computer Vision and Pattern Recognition},
  2017.

\bibitem{tan2008visibility}
Robby~T. Tan.
\newblock Visibility in bad weather from a single image.
\newblock In {\em IEEE Conference on Computer Vision and Pattern Recognition},
  2008.

\bibitem{tarel2009fast}
Jean-Philippe Tarel and Nicolas Hautiere.
\newblock Fast visibility restoration from a single color or gray level image.
\newblock In {\em IEEE International Conference on Computer Vision}.

\bibitem{tsai2018learning}
Yi-Hsuan Tsai, Wei-Chih Hung, Samuel Schulter, Kihyuk Sohn, Ming-Hsuan Yang,
  and Manmohan Chandraker.
\newblock Learning to adapt structured output space for semantic segmentation.
\newblock In {\em IEEE Conference on Computer Vision and Pattern Recognition},
  2018.

\bibitem{tzeng2017adversarial}
Eric Tzeng, Judy Hoffman, Kate Saenko, and Trevor Darrell.
\newblock Adversarial discriminative domain adaptation.
\newblock In {\em IEEE Conference on Computer Vision and Pattern Recognition},
  2017.

\bibitem{wang2018recovering}
Xintao Wang, Ke Yu, Chao Dong, and Chen Change~Loy.
\newblock Recovering realistic texture in image super-resolution by deep
  spatial feature transform.
\newblock In {\em IEEE Conference on Computer Vision and Pattern Recognition},
  2018.

\bibitem{Yang_2018_ECCV}
Dong Yang and Jian Sun.
\newblock Proximal dehaze-net: A prior learning-based deep network for single
  image dehazing.
\newblock In {\em European Conference on Computer Vision}, 2018.

\bibitem{Zhang_2018_CVPR}
He Zhang and Vishal~M. Patel.
\newblock Densely connected pyramid dehazing network.
\newblock In {\em IEEE Conference on Computer Vision and Pattern Recognition},
  2018.

\bibitem{Zhang:HazeRD:ICIP17b}
Yanfu Zhang, Li Ding, and Gaurav Sharma.
\newblock Hazerd: an outdoor scene dataset and benchmark for single image
  dehazing.
\newblock In {\em IEEE International Conference on Image Processing}, 2017.

\bibitem{zheng2018t2net}
Chuanxia Zheng, Tat-Jen Cham, and Jianfei Cai.
\newblock T2net: Synthetic-to-realistic translation for solving single-image
  depth estimation tasks.
\newblock In {\em European Conference on Computer Vision}, 2018.

\bibitem{zhu2017unpaired}
Jun-Yan Zhu, Taesung Park, Phillip Isola, and Alexei~A Efros.
\newblock Unpaired image-to-image translation using cycle-consistent
  adversarial networks.
\newblock In {\em IEEE International Conference on Computer Vision}, 2017.

\bibitem{zhu2015fast}
Qingsong Zhu, Jiaming Mai, Ling Shao, et~al.
\newblock A fast single image haze removal algorithm using color attenuation
  prior.
\newblock {\em IEEE Transactions on Image Processing}, 2015.

\end{thebibliography}
}

\end{document}